\documentclass[10pt,journal,compsoc]{IEEEtran}
\usepackage{amsmath,amsfonts}
\usepackage{algorithm}
\usepackage{algorithmicx}
\usepackage{algpseudocode} 
\usepackage{array}
\usepackage{graphicx}
\usepackage{subcaption}
\usepackage{textcomp}
\usepackage{stfloats}
\usepackage{url}
\usepackage{verbatim}
\usepackage{cite}

\usepackage[breaklinks=true,bookmarks=true,pagebackref=false]{hyperref}
\usepackage[numbers,sort]{natbib}
\usepackage{cleveref}
\usepackage{color}
\usepackage{booktabs}
\usepackage{threeparttable}
\usepackage{multirow} 
\usepackage{rotating}

\makeatletter
\def\@IEEEtitleabstractindextext{}
\makeatother

\begin{document}
\title{A Pose-only Geometric Constraint for Multi-Camera Pose Adjustment}
\author{Shunkun Liang, Banglei Guan, Bin Li, Qifeng Yu, and Yang Shang
	\IEEEcompsocitemizethanks{
		\IEEEcompsocthanksitem 
		S. Liang, B. Guan, B. Li, Q. Yu and Y. Shang are with the College of Aerospace Science and Engineering, National University of Defense Technology, Changsha 410000, China. E-mail: \{liangshunkun, guanbanglei12, libin19a, yuqifeng, shangyang1977\}@nudt.edu.cn.
		\IEEEcompsocthanksitem
		Corresponding author: Banglei Guan.
		\IEEEcompsocthanksitem
		This work was supported by the National Natural Science Foundation of China under Grant 12372189 and the Science and Technology Innovation Program of Hunan Province under Grant 2025RC1045
	}%
}

\maketitle

\begin{abstract}
Multi-camera systems offer rich observation capabilities for visual navigation and 3D scene reconstruction; however, the resulting feature redundancy often compromises computational efficiency. This challenge is particularly pronounced during bundle adjustment, where the non-linear optimization of both system poses and scene points incurs substantial computational overhead. To address this challenge, this paper introduces a pose-only geometric constraint for multi-camera systems and proposes a corresponding pose adjustment algorithm. Specifically, we use generalized camera model to establish a unified representation of the multi-camera system. Building upon this model, we formulate the multi-camera pose-only constraint, which implicitly represents a 3D scene point using two base observations and their associated poses, thereby achieving a pose-only representation of the projection geometry. Subsequently, we introduce a multi-camera pose adjustment algorithm that eliminates 3D points from the parameter space, thereby achieving efficient and focused pose optimization. Experimental results on both synthetic and real-world datasets demonstrate that the proposed algorithm outperforms baseline bundle adjustment methods in computational efficiency, while maintaining or even improving pose estimation accuracy.
\end{abstract}


\section{Introduction}
\label{sec:intro}
\IEEEPARstart{R}{ecovering} the visual system poses and the scene structure from a set of images is a classic and fundamental problem in 3D vision. Conventional monocular or stereo systems, limited by their narrow field of view, often suffer from fragility and low accuracy in ill-conditioned scenarios such as dynamic environments, low-texture regions, or occlusions. Multi-camera systems typically offer a broader field of view and richer observations, thus demonstrating greater robustness in these challenging scenarios. Their unique advantages have led to widespread attention in applications such as Visual Odometry (VO) \cite{Liu2018Towards,Forster2017SVO}, 3D reconstruction \cite{Cui2023mcsfm,Schmied2023r3d3}, and autonomous driving \cite{Qin2020AVP,Heng2019AutoVision}. Within these applications, the pose optimization of multi-camera systems is a critical technique that determines localization accuracy and 3D reconstruction quality. 

\begin{figure}[tbp]
	\includegraphics[width=\linewidth]{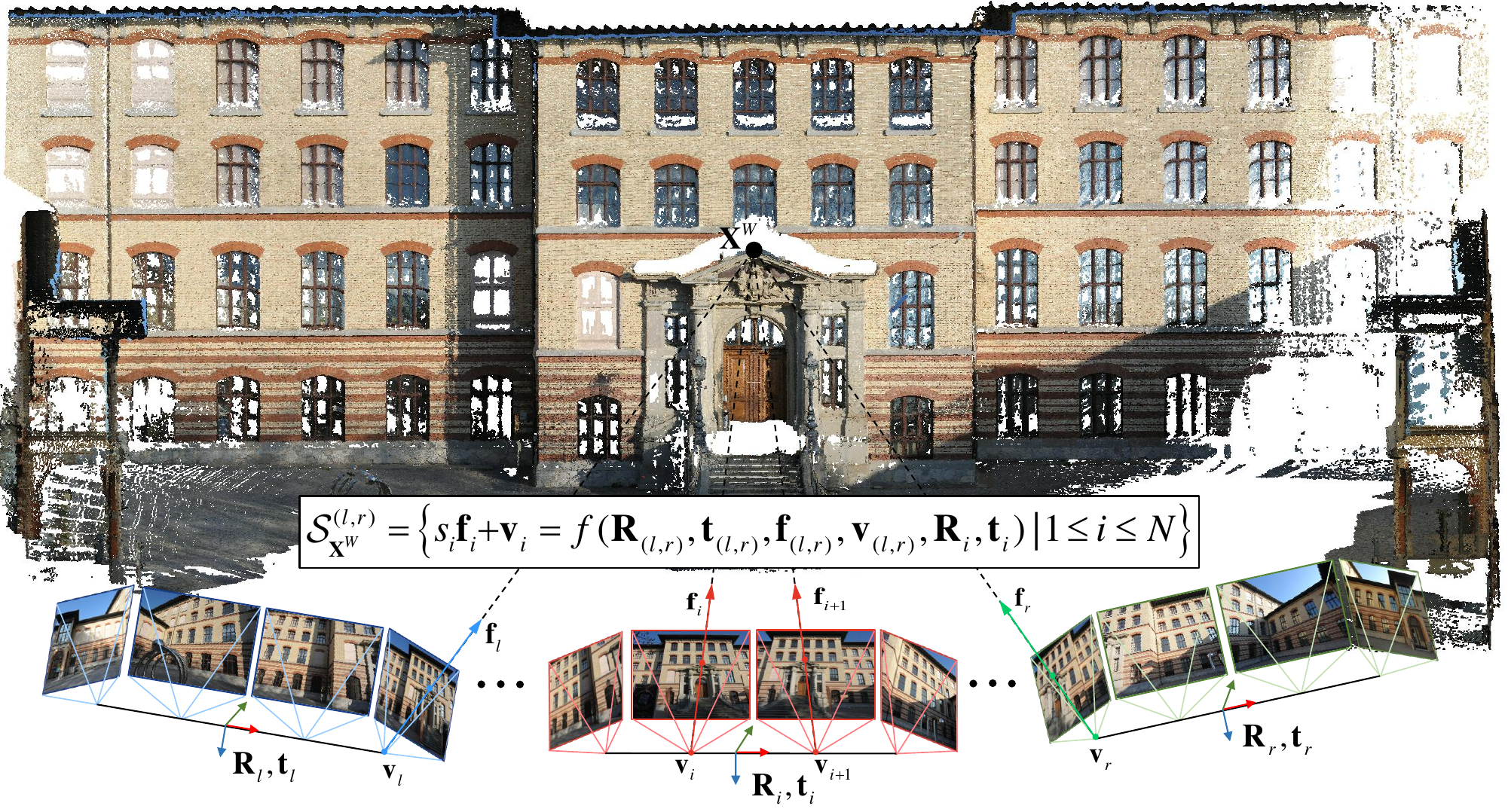} 
	\caption{Schematic diagram of multi-camera pose-only constraint. The multi-camera system is represented by the generalized camera model, where each image observation uniquely corresponds to a 6D vector $(\mathbf{f}, \mathbf{v})$. After determining the left- and right-base observations for 3D point $\mathbf{X}^W$, the remaining observations can be analytically expressed as functions of the base observations and their corresponding poses, together forming the pose-only constraint set $\mathcal{S}_{\mathbf{X}^W}^{(l, r)}$.}
	\label{fig:SXW_first_figure}
\end{figure}

Bundle adjustment optimizes 3D scene points and camera poses by minimizing reprojection errors, serving as the gold standard for achieving final estimation accuracy. \cite{Triggs2000bundle}. Many excellent methods and frameworks have been proposed for improving its computational efficiency and scalability \cite{Agarwal2010bundle,Agarwal_Ceres_Solver_2022,Ye2022coli,Weber2023power}. These methods can be directly extended and applied to multi-camera systems, primarily by introducing additional extrinsic constraints \cite{Schneider2012bundle,Urban2017multicol}. However, the efficiency problem of pose optimization in multi-camera system has received relatively less attention. The abundant observational features also introduce data redundancy, causing the parameter space to grow rapidly and leading to computational bottlenecks. Therefore, how to effectively enhance the efficiency of pose optimization for multi-camera systems while ensuring accuracy has become a core issue that needs to be addressed for their practical application. 

Jointly optimizing camera poses and 3D points results in a high-dimensional parameter space significantly increases the computational cost per iteration. We note that in simultaneous localization and mapping (SLAM) and structure from motion (SfM), 3D scene points are essentially determined by camera poses and observations, and can be reconstructed via triangulation once accurate poses are available. This indicates that 3D points are not essential component in the pose optimization process. Eliminating them from the optimization variables can significantly reduce the parameter space. Recently, Cai et al. proposed a depth-pose-only (DPO) constraint for a single camera, which analytically expresses 3D scene point depths using only poses \cite{Cai2023pose,Cai2019equivalent}. Inspired by this work, we extend the pose-only formulation to multi-camera systems and propose a corresponding pose adjustment algorithm.

Extending directly at the individual camera level requires addressing the issues of coordinate system drift and scale inconsistency among the camera subsystems. To avoid these challenges, we model the multi-camera system as a generalized camera \cite{Pless2003using}. The observation in multi-camera system is represented as a 6D-vector, which uniquely corresponds to an image point through the known extrinsic and each camera's intrinsic. This modeling approach allows our algorithm to be applied to arbitrary multi-camera system configurations. Based on this representation, we propose a multi-camera pose-only constraint where 3D scene points are analytically expressed using two base observations and corresponding poses. The remaining observations can be represented as a function of the base observation pair and poses, without relying on explicit 3D scene points coordinates. \Cref{fig:SXW_first_figure} illustrates the schematic diagram of our multi-camera pose-only constraint.

The multi-camera pose-only constraint allows us to define a cost function that depends solely on poses. Based on this formulation, we propose a multi-camera pose adjustment algorithm, in which 3D point variables are eliminated from the optimization process, thereby effectively avoiding a loss in computational efficiency. Accordingly, we use the term pose adjustment to distinguish it from conventional bundle adjustment and to emphasize our core innovation. Base observations selection is a critical step in the pose-only constraint, as it directly influences the final optimization accuracy. We therefore propose a base observations selection algorithm based on the uncertainty ellipsoid roundness. Following pose optimization, we propose a statistically optimal 3D reconstruction algorithm for the generalized camera model to complete the visual geometric pipeline.

Our contributions are summarized as follows: 

\begin{itemize}
	\item[$\bullet$] We propose a \textbf{multi-camera pose-only constraint} based on generalized camera model. The projection geometry is equivalently expressed as a function of two base observations and the corresponding poses.
	\item[$\bullet$] We develop a \textbf{multi-camera pose adjustment algorithm} that removes 3D points from the optimization variables and focuses on poses, avoiding the high-dimensional parameter space of bundle adjustment.
	\item[$\bullet$] We propose a key \textbf{base observations selection algorithm} for the multi-camera pose-only constraint. This algorithm uses the uncertainty ellipsoid roundness of reconstructed 3D points as the metric.
	\item[$\bullet$] Given the optimized poses, we propose a \textbf{statistically optimal reconstruction algorithm} to complete visual geometric computing. 
\end{itemize}

The paper is organized as follows. \Cref{sec:related_work} reviews related works on multi-camera visual odometry and bundle adjustment. The multi-camera pose-only constraint is established in \Cref{sec:mcig}. In \Cref{sec:mcpa}, we propose a complete multi-camera pose adjustment algorithm along with its auxiliary algorithms. \Cref{sec:experiment} demonstrates the performance of the proposed algorithms on both synthetic and real-world data. The paper is concluded in \Cref{sec:conclusion}.

\section{Related Work}
\label{sec:related_work}
Since the proposed method is primarily aimed at addressing the pose optimization problem of multi-camera system, we review works on multi-camera visual odometry and bundle adjustment.

\textbf{Multi-Camera Visual Odometry}. Currently, there are numerous studies and open-source systems on visual odometry. These studies mainly focus on monocular \cite{Davison2007MonoSLAM,Forster2014svo}, stereo \cite{Mur2017orbslam2,Wang2017stereo}, and fused visual-inertial odometry\cite{Qin2018vins,Leutenegger2015keyframe}. Recently, multi-camera visual odometry is gradually gaining more attention due to its wider field of view and stronger robustness. MCPTAM \cite{Tribou2015multi} extends PTAM \cite{Klein2007parallel} to achieve real-time pose tracking and mapping for multiple non-overlapping field-of-view cameras. MultiCol-SLAM \cite{Urban2016multicol} expands on ORB-SLAM2 \cite{Mur2017orbslam2} to support rigidly connected multi-fisheye camera systems and introduces the multi-keyframes for pose optimization. SVO \cite{Forster2017SVO} directly extends its monocular work \cite{Forster2014svo} to multi-camera systems. In addition, recent works extend monocular visual-inertial odometry to multi-camera systems to further enhance the robustness and accuracy of pose estimation \cite{He2022towards,Zhang2023bamf,Wang2024mavis}. These extended methods achieve significant performance improvements compared to monocular systems. 

In recent years, some literature has focused on addressing the computational burden caused by redundant features in multi-camera systems. Zhang et al. \cite{Zhang2021balancing} explore feature tracking methods for multi-camera visual odometry front-end and propose a selection mechanism for the optimal feature subset, effectively reducing the back-end optimization time. Kuo et al. \cite{Kuo2020redesigning} redesign the front-end of multi-camera SLAM and introduce an entropy-based adaptive keyframe selection algorithm to enhance system efficiency. Liu et al. \cite{Liu2018Towards} propose a pose tracker and local mapper suitable for arbitrary multi-camera systems. The pose tracker optimizes the pose by minimizing the photometric error between the keyframe from multiple cameras and the current frame. 
Kaveti et al. \cite{Kaveti2023Design} calculate and utilize the overlapping regions of multiple camera fields of view for cross-matching, thereby avoiding duplicate observations of features. 
These methods primarily focus on optimizing front-end performance. However, the pose optimization stage in the back-end still faces the issue of redundant observed features.

\textbf{Bundle Adjustment.}
Bundle adjustment (BA) represents one of the core steps in 3D vision applications, primarily focusing on optimizing camera system parameters and 3D scene points by minimizing reprojection errors \cite{Hartley2003multiple,Triggs2000bundle} or photometric errors \cite{Delaunoy2014photometric,Engel2017direct}. 
This technique ensures the final accuracy of most vision , such as VO \cite{Mur2017orbslam2,guan2025affine}, SfM \cite{Johannes2016colmap,Cui2023mcsfm}, and camera calibration \cite{Liang2024camera,tan2026optimal}.
A detailed study on bundle adjustment can be found in the classical literature \cite{Triggs2000bundle}. For several decades, researchers have been working to improve the computational efficiency and scalability of bundle adjustment. Recent work has focused on achieving large-scale bundle adjustment through efficient solutions to the normal equations \cite{Agarwal2010bundle,Demmel2021square,Weber2023power,zhou2020stochastic}. Agarwal et al. \cite{Agarwal2010bundle} propose an efficient solver using the preconditioned conjugate gradient. Square root BA \cite{Demmel2021square} marginalizes 3D points via QR decomposition to avoid explicitly forming the Hessian matrix, thereby reducing both memory footprint and computational cost. Power BA \cite{Weber2023power} approximates the inverse of the Schur complement using a power series expansion to avoid the high cost of explicitly inverting the Hessian matrix. In addition, some studies have significantly improved the runtime through efficient parallelization and distributed computing techniques \cite{wu2011multicore,Ren2022megba}.

In addition to accelerating the solving process of bundle adjustment itself, another approach is to eliminate the optimization of 3D scene points that consume significant computational resources. The pose graph optimization method \cite{Lu1997globally} significantly improves computational speed by omitting the optimization of 3D scene points; however, it completely ignores image observation, resulting in accuracy that is entirely dependent on the quality of relative pose estimates. Lee et al. \cite{Lee2021rotation} extend the pure rotation constraints of the two-view geometry \cite{Kneip2013direct} and proposed a multi-view global rotation optimization method. This approach considers image observations as input, allowing for independent rotation optimization. Cai et al. \cite{Cai2019equivalent} introduce a pair of pose-only constraints in the two-view geometry, analytically representing scene depth through camera poses. These constraints are subsequently extended to multi-view situations \cite{Cai2023pose}, leading to the depth-pose-only (DPO) constraint. The DPO constraint has been applied in SLAM systems \cite{Ge2024pipo,Wang2025pokf}, demonstrating significant computational advantages.

Compared to monocular systems, research on multi-camera systems bundle adjustment is relatively limited. Schneider et al. \cite{Schneider2012bundle} represent the observed and scene points as spherical normalized vectors, transforming the minimization of re-projection errors into the computation of the null space of direction vectors in each iteration. Urban et al. \cite{Urban2017multicol} represent each camera in the multi-camera system with a general camera model \cite{Scaramuzza2006Omnidirectional}, and extend the standard collinearity equations. Based on this, a bundle adjustment method applicable to any multi-camera system is proposed. Recently, the open-source SfM pipeline COLMAP \cite{Johannes2016colmap} has incorporated support for multi-camera systems and provides an interface to jointly optimizes global rig pose and 3D scene points. Mathematically, its objective function is equivalent to that of \cite{Urban2017multicol}, which minimizes the sum of reprojection errors across all constituent cameras. However, none of these methods address the computational efficiency issues arising from multi-camera systems. This paper introduces the concept of pose-only for the first time in the pose optimization of multi-camera systems, effectively addressing the efficiency issues caused by redundant features.

\section{Multi-camera Pose-only Constraint}
\label{sec:mcig}

In this section, we model the multi-camera system as a generalized camera model and derive its projection equation. Based on this, we further propose the pose-only geometry along with the corresponding constraint set.

\subsection{Generalized Camera Model}
\label{sec:mcs}
Without loss of generality, we assume a multi-camera system composed of $C$ cameras, where extrinsic parameters are known. Let us denote the projection function of the $c$-th camera as $\pi_c$, whose intrinsic parameters are also known. The projection process of the $c$-th camera is expressed as:
\begin{equation}
	s \tilde{\mathbf{x}} = \pi_c(\mathbf{X}^c),
\end{equation}
where $s$ is a scale factor, $\tilde{\mathbf{x}} = [u, v, 1]^T$ is the homogeneous form of 2D image pixel corresponding to the 3D scene point $\mathbf{X}^c$ observed by camera. Note that $\pi_c$ is not unique; it corresponds to different camera models, depending on the application scenario. For a classical perspective camera model, $\pi_c(\mathbf{X})$ is expressed as: $\pi_c(\mathbf{X}) = \mathbf{K}_c \mathbf{X}$, where $\mathbf{K}_c$ is the intrinsic matrix of $c$-th camera. For wide-field cameras, $\pi_c$ can be replaced by a fisheye model \cite{Kannala2006generic} or omnidirectional model \cite{Scaramuzza2006Omnidirectional}. The extrinsic parameters of $c$-th camera are denoted as $(\mathbf{R}_c, \mathbf{t}_c)$, where $\mathbf{R}_c \in SO(3)$ is the rotation matrix and $\mathbf{t}_c \in \mathbb{R}^3$ is the translation vector. Extrinsic parameters refer to the transformation from the $c$-th camera to the body coordinate system of multi-camera.

Given a 3D scene point $\mathbf{X}^W$, which is observed by the $c$-th camera in the multi-camera system, the corresponding 2D image pixel is denoted as ${\mathbf{x}}$. At this time, assuming the pose of multi-camera system is $(\mathbf{R}, \mathbf{t})$, the specific expression of the projection equation is:
\begin{equation}
	s \tilde{\mathbf{x}} = \pi_c(\mathbf{R}_c^T(\mathbf{R}\mathbf{X}^W + \mathbf{t}) - \mathbf{R}_c^T \mathbf{t}_c).
	\label{eq:sx}
\end{equation}
Moving the parameters related to the multi-camera system to the left side of the equation, \cref{eq:sx} can be rewritten as:
\begin{equation}
	s \mathbf{R}_c \pi_c^{-1}(\tilde{\mathbf{x}}) + \mathbf{t}_c 
	= \mathbf{R} \mathbf{X}^W + \mathbf{t},
	\label{eq:sRx}
\end{equation}
where $\pi_c^{-1}$ is the back projection function of the $c$-th camera. We define $\mathbf{f} = \mathbf{R}_c \pi_c^{-1}(\tilde{\mathbf{x}})$ and $\mathbf{v} = \mathbf{t}_c$. Here, $\mathbf{f}$ and $\mathbf{v}$ are the normalized direction vector and vertex of the observation ray corresponding to the 2D pixel $\mathbf{x}$, respectively. Finally, the projection equation can be expressed as:
\begin{equation}
	s \mathbf{f} + \mathbf{v} = \mathbf{R} \mathbf{X}^W + \mathbf{t}.
	\label{eq:sfv}
\end{equation}
The projection function \cref{eq:sfv} allows us to treat the multi-camera system as a generalized camera model. Each 2D image pixel $\mathbf{x}$ of every camera uniquely corresponds to a 6D observation ray $(\mathbf{f}, \mathbf{v})$. This projection equation is universal and applicable to complex multi-camera systems with arbitrary configurations and camera models. Additionally, the extrinsic calibration scale of the multi-camera system is directly incorporated into the projection equation, ensuring geometric global consistency.

\subsection{Pose-only Constraint}
\label{sec:mcpoc}
Based on the generalized camera model, we hereby propose the pose-only geometry constraint for multi-camera systems. Consider a 3D scene point $\mathbf{X}^W$ that is observed by a multi-camera system from multiple poses, resulting in \(N\) observations. We denote the $l$-th and $r$-th observations of $\mathbf{X}^W$ as $(\mathbf{f}_l, \mathbf{v}_l)$ and $(\mathbf{f}_r, \mathbf{v}_r)$, which correspond to poses $(\mathbf{R}_l, \mathbf{t}_l)$ and $(\mathbf{R}_r, \mathbf{t}_r)$, respectively. The projection equations for those two observations are expressed as:
\begin{align}
	s_r \mathbf{f}_r  + \mathbf{v}_r &= \mathbf{R}_r\mathbf{X}^W + \mathbf{t}_r, \label{eq:sfvr} \\
	s_l \mathbf{f}_l  + \mathbf{v}_l &= \mathbf{R}_l\mathbf{X}^W + \mathbf{t}_l. \label{eq:sfvl}
\end{align}
From \cref{eq:sfvl}, the point $\mathbf{X}^W$ can be represented by observation $(\mathbf{f}_l, \mathbf{v}_l)$ and pose $(\mathbf{R}_l, \mathbf{t}_l)$ as:
\begin{equation}
	\mathbf{X}^W = \mathbf{R}_l^T(s_l \mathbf{f}_l + \mathbf{v}_l - \mathbf{t}_l).
	\label{eq:XWl}
\end{equation}
Substituting \cref{eq:XWl} into \cref{eq:sfvr}, we get:
\begin{equation}
	s_r \mathbf{f}_r  + \mathbf{v}_r = \mathbf{R}_{lr}(s_l \mathbf{f}_l + \mathbf{v}_l) + \mathbf{t}_{lr},
	\label{eq:sfvr=sfvl}
\end{equation}
where 
\begin{equation*}
	\mathbf{R}_{lr} = \mathbf{R}_r\mathbf{R}_l^T,
	\mathbf{t}_{lr} = \mathbf{t}_r - \mathbf{R}_r\mathbf{R}_l^T \mathbf{t}_l.
\end{equation*}
By left multiplying the skew-symmetric matrix of $\mathbf{f}_r$ on both sides of \cref{eq:sfvr=sfvl}, we obtain:
\begin{equation}
	\begin{aligned}
		[\mathbf{f}_r]_\times \mathbf{v}_r &= [\mathbf{f}_r]_\times [\mathbf{R}_{lr}(s_l \mathbf{f}_l + \mathbf{v}_l) + \mathbf{t}_{lr}] \\
		& = s_l [\mathbf{f}_r]_\times \mathbf{R}_{lr} \mathbf{f}_l + [\mathbf{f}_r]_\times \mathbf{R}_{lr} \mathbf{v}_l + [\mathbf{f}_r]_\times \mathbf{t}_{lr}.
		\label{eq:fxv}
	\end{aligned}
\end{equation}
Then, we can calculate the left scale factor $s_l$ as:
\begin{equation}
	s_l = \frac{||[\mathbf{f}_r]_\times(\mathbf{v}_r - \mathbf{R}_{lr} \mathbf{v}_l - \mathbf{t}_{lr})||}{||[\mathbf{f}_r]_\times \mathbf{R}_{lr}  \mathbf{f}_l || }
	= \frac{\lambda_{lr}}{\theta_{lr}},
	\label{eq:sl}
\end{equation}
where $\lambda_{lr} = \|[\mathbf{f}_r]_\times(\mathbf{v}_r - \mathbf{R}_{lr} \mathbf{v}_l - \mathbf{t}_{lr})\|$ and $\theta_{lr} = \|[\mathbf{f}_r]_\times \mathbf{R}_{lr}  \mathbf{f}_l\|$. Note that $s_l$ is a function of the observations $(l, r)$ and their poses, and has nothing to do with the 3D point $\mathbf{X}^W$. Now, let us consider a new observation $(\mathbf{f}_i, \mathbf{v}_i)$, which corresponds to the $i$-th observation of $\mathbf{X}^W$ from the multi-camera system. The corresponding pose is denoted as $(\mathbf{R}_i, \mathbf{t}_i)$. Similarly, the observation pair $(l, i)$ can also be written in the form of \cref{eq:sfvr=sfvl} as:
\begin{equation}
	s_i \mathbf{f}_i \!+\! \mathbf{v}_i = \mathbf{R}_{li}(s_l \mathbf{f}_l + \mathbf{v}_l) + \mathbf{t}_{li}
	\label{eq:sfvi}
\end{equation}

\noindent{\textit{\textbf{Proposition 1. (Equal-depth Constraint)}}. Let $s_l^{(l, r)}$ and $s_l^{(l, i)}$ denote the scale factors of the left observation in pairs $(l, r)$ and $(l, i)$ respectively. Then: $s_l^{(l, r)} = s_l^{(l, i)}$.}

\noindent \textit{\textbf{Proof.}} Since both observation pairs $(l, r)$ and $(l, i)$ correspond to the same 3D point $\mathbf{X}^W$, the expression for $\mathbf{X}^W$ in terms of the $l$-th observation must be unique. Substituting $s_l^{(l, r)}$ and $s_l^{(l, i)}$ into \cref{eq:XWl} yields:
\begin{align*}
	\mathbf{X}^W = \mathbf{R}_l^T(s_l^{(l, r)} \mathbf{f}_l + \mathbf{v}_l - \mathbf{t}_l) &= \mathbf{R}_l^T(s_l^{(l, i)} \mathbf{f}_l + \mathbf{v}_l - \mathbf{t}_l)\\
	\implies:  s_l^{(l, r)} \mathbf{f}_l &= s_l^{(l, i)} \mathbf{f}_l
\end{align*}
Since $\mathbf{f}_l$ is a nonzero unit vector, the above equation holds only when $s_l^{(l, r)} = s_l^{(l, i)}$.  $\hfill\square$

The equal-depth constraint arises from a fundamental fact: in the generalized camera model, each observation corresponds to a unique 3D point. Moreover, the scale factor of the same observation is unique and does not change with different pairings. 

We can see from \cref{eq:sfvi} that the new observation $(\mathbf{f}_i, \mathbf{v}_i)$ can be fully represented by the observation pair $(l, r)$. It is solely dependent on the pose and observations and is completely independent of the 3D point $\mathbf{X}^W$. 

\noindent{\textit{\textbf{Proposition 2. (Geometric Equivalence)}}. \Cref{eq:sfvi} is geometrically equivalent to the projection equation of observation $(\mathbf{f}_i, \mathbf{v}_i)$.}

\noindent \textit{\textbf{Proof.}} According to the derivation of \cref{eq:sfv} to \cref{eq:sfvi}, the necessity is obviously established.

Assuming that \cref{eq:sfvi} holds, we have:
\begin{equation*}
	\begin{aligned}
		s_i \mathbf{f}_i \!+\! \mathbf{v}_i 
		&= \mathbf{R}_{li}(s_l \mathbf{f}_l + \mathbf{v}_l) + \mathbf{t}_{li} \\
		&= \mathbf{R}_{li}(\mathbf{R}_l\mathbf{X}^W + \mathbf{t}_l) + \mathbf{t}_{li} \\
		&= \mathbf{R}_i\mathbf{R}_l^T \mathbf{R}_l\mathbf{X}^W + \mathbf{R}_i\mathbf{R}_l^T \mathbf{t}_l + \mathbf{t}_i - \mathbf{R}_i\mathbf{R}_l^T \mathbf{t}_l \\
		&= \mathbf{R}_i \mathbf{X}^W + \mathbf{t}_i
	\end{aligned}
\end{equation*}
Both sufficiency and necessity are established.  $\hfill\square$

The equivalence between \cref{eq:sfvi} and the projection equation implies that our derivation process maintains the integrity of the geometric information without any loss. In summary, for all observations corresponding to $\mathbf{X}^W$, we can express them as a function of the observation pair $(l, r)$. For $\mathbf{X}^W$, we can obtain a constraint set:
\begin{equation}
\mathcal{S}_{\mathbf{X}^W}^{(l, r)} = \left\{  s_i \mathbf{f}_i \!+\! \mathbf{v}_i \!=\! \mathbf{R}_{li}\left(\frac{\lambda_{lr}}{\theta_{lr}} \mathbf{f}_l \!+\! \mathbf{v}_l \right) \!+\! \mathbf{t}_{li}  | 1 \le i \le N, i \neq l \right\}
	\label{eq:SXW}
\end{equation}
Since this constraint set is only related to the poses, we call $\mathcal{S}_{\mathbf{X}^W}^{(l, r)}$ the \textbf{multi-camera pose-only constraint} on $\mathbf{X}^W$. The $l$-th and $r$-th observations are defined as left-base and right-base observations of $\mathbf{X}^W$, respectively. The selection algorithm of base observations is detailed in \cref{sec:SBO}.

For a multi-camera system with overlapping fields of view, a 3D point $\mathbf{X}^W$ may be captured by multiple cameras at the same time. As a result, the corresponding base observation pair $(l,r)$ for the $\mathbf{X}^W$ may be generated at the same multi-camera pose, i.e. $\mathbf{R}_r = \mathbf{R}_l$ and $\mathbf{t}_r = \mathbf{t}_l$. In this case, \cref{eq:sl} can be rewritten as:
\begin{equation}
	s_l = \frac{||[\mathbf{f}_r]_\times(\mathbf{v}_r - \mathbf{v}_l)||}{||[\mathbf{f}_r]_\times \mathbf{f}_l || }
	= \frac{\lambda_{lr}}{\theta_{lr}}
	\label{eq:sl2}
\end{equation}
Substituting \cref{eq:sl2} into \cref{eq:SXW}, the multi-camera pose constraint is still valid.

Compared to the projection equation, the multi-camera pose-only constraint effectively eliminates the dependence on explicit 3D points. In other words, the 3D points can be implicitly represented by the base observation pair and their poses. This formulation offers significant advantages. During pose optimization, there is no need to treat numerous 3D points as optimization variables, which significantly reduces the dimensionality of the parameter space and improves the efficiency of the optimization. The following provides a detailed introduction to our pose adjustment algorithm.

\section{Multi-camera Pose Adjustment}
\label{sec:mcpa}
In this section, we introduce the pose adjustment algorithm based on the multi-camera pose-only constraint. Unlike bundle adjustment methods, our algorithm does not explicitly rely on 3D points, which effectively reduces the dimensionality of the parameter space and avoids errors caused by inaccurate 3D points.

\subsection{Pose-Only Residual}
\label{sec:paa}
Visual SLAM and SFM systems \cite{Mur2017orbslam2, Johannes2016colmap} typically recover 3D scene points through triangulation and use these points to associate registered frames and incoming frames. This association is derived from the projection equations (e.g. \cref{eq:sx} or \cref{eq:sfv}). The classic multi-camera bundle adjustment algorithms jointly optimize the 3D points and the poses of camera system by minimizing the reprojection error \cite{Schneider2012bundle,Urban2017multicol}. However, the large number of 3D points leads to a dramatic increase in the parameter space, significantly raising computational costs. We introduce the multi-camera pose-only constraint and demonstrate its equivalence to the projection equations. The proposed constraint in \cref{eq:SXW} demonstrates that each observation can be computed from the left-base and right-base observations. This approach replaces the step of reprojection of 3D points onto the images, eliminating the explicit dependence on 3D points.

Let $\mathbf{X}^W_j$ represent the $j$-th 3D point in the world, which is observed by a calibrated multi-camera system within multiple poses. The left-base and right-base observations of $\mathbf{X}^W_j$ are denoted as $(\mathbf{f}_l, \mathbf{v}_l)$ and $(\mathbf{f}_r, \mathbf{v}_r)$, respectively. Then, we can obtain the constraint set $\mathcal{S}_{\mathbf{X}^W_j}^{(l, r)}$ for $\mathbf{X}^W_j$. Representing the pose of the \( i \)-th observation as $(\mathbf{R}_i, \mathbf{t}_i)$, the $i$-th observation can be calculated as:
\begin{equation}
	s_{ij} \mathbf{f}_{ij} \!+\! \mathbf{v}_{ij} = \mathbf{R}_{li}(\frac{\lambda_{lr}}{\theta_{lr}} \mathbf{f}_l + \mathbf{v}_l) + \mathbf{t}_{li}.
	\label{eq:sfvij}
\end{equation}
From \cref{eq:sRx} and \cref{eq:sfv}, it can be seen that $\mathbf{f}_{ij}$ is a function of the 2D image pixel, and $\mathbf{v}_{ij}$ depends on the known multi-camera extrinsic parameters. Therefore, we move $\mathbf{v}_{ij}$ to the right side of the equation and multiply both sides by $\theta_{lr}$. \cref{eq:sfvij} can be rewritten as:
\begin{equation}
	s_{ij} \theta_{lr} \mathbf{f}_{ij} = 
	\lambda_{lr} \mathbf{R}_{li} \mathbf{f}_l + \theta_{lr} (\mathbf{R}_{li} \mathbf{v}_l + \mathbf{t}_{li} - \mathbf{v}_{ij}).
\end{equation}

The residual in classical bundle adjustment is defined as the distance between the reprojection of 3D points and the actual 2D pixels. In our case, it is the distance between the normalized direction vectors of the reprojection and the actual observations. We define the residuals using normalized spherical error instead of normalized planar error. Prior work \cite{Ye2022coli} demonstrates that normalized spherical error typically exhibits better convergence efficiency. Let $\mathbf{Y}_{ij} = \lambda_{lr} \mathbf{R}_{li} \mathbf{f}_l + \theta_{lr} (\mathbf{R}_{li} \mathbf{v}_l + \mathbf{t}_{li} - \mathbf{v}_{ij})$, our residual function is defined as:
\begin{equation}
	\mathbf{e}_{ij} = \frac{\mathbf{Y}_{ij}}{\left\| \mathbf{Y}_{ij} \right\|} - \hat{\mathbf{f}}_{ij},
	\label{eq:eij}
\end{equation}
where $\hat{\mathbf{f}}_{ij}$ is the actual $i$-th observation corresponding to $\mathbf{X}^W_j$. Given \( M \) 3D scene points in space, which are observed by a multi-camera system from multiple poses, with each 3D point resulting in a total of \( N \) observations, our pose adjustment cost function is defined as:
\begin{equation}
	\mathbf{T} = {\arg \min } \sum_{j=1}^{M} \sum_{i=1}^{N} \|\mathbf{e}_{ij}\|^2,
	\label{eq:argmin}
\end{equation}
where $\mathbf{T} = {\left\{\mathbf{R}_{i}, \mathbf{t}_{i}\right\}_{i=1 \ldots N}}$ is the pose set of multi-camera system. Note that due to the overlapping fields of view of the multi-camera system, different observations may correspond to the same pose. Therefore, the number of poses to be optimized is less than or equal to $N$.

Here, we discuss another residual configuration in the multi-camera pose adjustment algorithm. In \cref{eq:eij}, $\mathbf{Y}_{ij}$ is derived from left- and right-base observations. These two base observations also exhibit a distinction between primary and secondary. In \cref{eq:eij}, the left-base observation is treated as primary, which can be inferred from \cref{eq:sl}. \Cref{eq:sl} calculates the depth of left-base observation $s_l$, ultimately determining $\mathbf{Y}_{ij}$. Similarly, we can also select the right-base observation as primary, which results in an additional residual component. Following the derivation in \cref{eq:sfvr,eq:sfvl,eq:XWl,eq:sfvr=sfvl,eq:fxv,eq:sl}, the depth of right-base observation can be calculated as:
\begin{equation}
	s_r = \frac{\|[\mathbf{f}_l]_\times(\mathbf{v}_l - \mathbf{R}_{rl} \mathbf{v}_r - \mathbf{t}_{rl})\|}{\|[\mathbf{f}_l]_\times \mathbf{R}_{rl}  \mathbf{f}_r\| }
	= \frac{\lambda_{rl}}{\theta_{rl}},
\end{equation}
where
\begin{equation*}
	\mathbf{R}_{rl} = \mathbf{R}_l\mathbf{R}_r^T,
	\mathbf{t}_{rl} = \mathbf{t}_l - \mathbf{R}_l\mathbf{R}_r^T \mathbf{t}_r.
\end{equation*}
Then, we can obtain an additional set of pose-only constraints on $\mathbf{X}^W$:
\begin{equation}
\mathcal{S}_{\mathbf{X}^W}^{(r, l)} = \left\{ s_i \mathbf{f}_i \!+\! \mathbf{v}_i \!=\! \mathbf{R}_{ri}\left(\frac{\lambda_{rl}}{\theta_{rl}} \mathbf{f}_r \!+\! \mathbf{v}_r\right) \!+\! \mathbf{t}_{ri} \,\big|\, 1 \!\le\! i \!\le\! N, i \!\neq\! r \right\}.
	\label{eq:SXWr}
\end{equation}
Let $\mathbf{e}_{ij}^l$ and $\mathbf{e}_{ij}^r$ denote the residual functions using the left- and right-base observations as primary, respectively. $\mathbf{e}_{ij}^l$ is already calculated as \cref{eq:eij}. We can calculate $\mathbf{e}_{ij}^r$ as:
\begin{equation}
	\mathbf{e}_{ij}^r = \frac{\mathbf{Z}_{ij}}{\|\mathbf{Z}_{ij}\|} - \hat{\mathbf{f}}_{ij},
	\label{eq:eijr}
\end{equation}
where $\mathbf{Z}_{ij} = \lambda_{rl} \mathbf{R}_{ri} \mathbf{f}_r + \theta_{rl} (\mathbf{R}_{ri} \mathbf{v}_r + \mathbf{t}_{ri} - \mathbf{v}_{ij})$. Furthermore, the cost function of our multi-camera pose adjustment optimization problem can be redefined as:
\begin{equation}
	\mathbf{T} = {\arg \min } \sum_{j=1}^{M} \sum_{i=1}^{N} \|\mathbf{e}_{ij}^l\|^2 + \|\mathbf{e}_{ij}^r\|^2.
	\label{eq:mineler}
\end{equation}
Compared to \cref{eq:argmin}, the new cost function includes twice as many residual terms, offering additional constraints for optimization. This leads to more accurate pose adjustment but also increases computational complexity. Therefore, it is crucial to select an appropriate cost function for various applications to balance accuracy and efficiency.

\subsection{Jacobian Calculation}
A solution of \cref{eq:argmin} can be obtained from an initial value $\mathbf{T}_0$ using the Levenberg-Marquardt (LM) or Gauss-Newton (GN) iteration. The iterative increment $\Delta \mathbf{T}$ is estimated by constructing and solving the normal equation, and the core step is to calculate the Jacobian matrix $\partial{\mathbf{e}_{ij}}/\partial{\mathbf{T}}$. In this section, we present the specific forms of the Jacobian matrix calculations and introduce several computational techniques to enhance efficiency. The Jacobian matrices for the pose of observation $i$, $l$ and $r$ are expressed as:
\begin{equation}
	\begin{aligned}
		\mathbf{J}_{ij}^{i} &= \frac{\partial{\mathbf{e}_{ij}}}{\partial{\mathbf{Y}_{ij}}} \left( \frac{\partial{\mathbf{Y}_{ij}}}{\partial \boldsymbol{\varphi}_i}, \frac{\partial{\mathbf{Y}_{ij}}}{\partial\mathbf{t}_i} \right), \\
		\mathbf{J}_{ij}^{l} &= \frac{\partial{\mathbf{e}_{ij}}}{\partial{\mathbf{Y}_{ij}}} \left( \frac{\partial{\mathbf{Y}_{ij}}}{\partial\boldsymbol{\varphi}_l}, \frac{\partial{\mathbf{Y}_{ij}}}{\partial\mathbf{t}_l} \right), \\
		\mathbf{J}_{ij}^{r} &= \frac{\partial{\mathbf{e}_{ij}}}{\partial{\mathbf{Y}_{ij}}} \left( \frac{\partial{\mathbf{Y}_{ij}}}{\partial \boldsymbol{\varphi}_r}, \frac{\partial{\mathbf{Y}_{ij}}}{\partial\mathbf{t}_r} \right),
	\end{aligned}
	\label{eq:Jilr}
\end{equation}
{where $\boldsymbol{\varphi}$ is the right perturbation of rotation}. The first part of \cref{eq:Jilr} corresponds to the spherical normalization, which can be calculated as:
\begin{equation}
	\frac{\partial{\mathbf{e}_{ij}}}{\partial{\mathbf{Y}_{ij}}} = \frac{\mathbf{I}_3}{\left\| \mathbf{Y}_{ij} \right\|}  - \frac{\mathbf{Y}_{ij} \mathbf{Y}_{ij}^T}{\left\| \mathbf{Y}_{ij}  \right\|^3}
	\label{eq:dedY}
\end{equation}
where $\mathbf{I}_3$ is the $3 \times 3$ identity matrix. The second part of \cref{eq:Jilr} corresponds to the partial derivative of the poses, with specific results provided in the following.

Constructing the normal equations for the optimization problem is one of the most time-consuming steps. To facilitate the practical application of the pose adjustment in applications such as SLAM, we provide the analytical expression for the Jacobian matrix. For notational simplicity, we use the superscript $^{\wedge}$ to denote the skew-symmetric matrix of a vector. The partial derivative of $\mathbf{Y}_{ij}$ with respect to $(\boldsymbol{\varphi}_i, \mathbf{t}_i)$ is calculated as:
\begin{align*}
	\frac{\partial \mathbf{Y}_{ij}}{\partial \mathbf{t}_i} &= {\theta}_{lr} \mathbf{I}_3, \\
	\frac{\partial \mathbf{Y}_{ij}}{\partial \boldsymbol{\varphi}_i} &= \mathbf{R}_i \left[ \theta_{lr} \! (\mathbf{R}_l^T \mathbf{t}_l)^{\wedge} \! - \lambda_{lr} \! (\mathbf{R}_l^T \mathbf{f}_l)^{\wedge} \! - \theta_{lr} \! (\mathbf{R}_l^T \mathbf{v}_l)^{\wedge} \right].
\end{align*}
Let $\mathbf{a} = \mathbf{t}_{lr} + \mathbf{R}_{lr} \mathbf{v}_l - \mathbf{v}_r$, the partial derivative of $\mathbf{Y}_{ij}$ with respect to $(\boldsymbol{\varphi}_r, \mathbf{t}_r)$ is calculated as:
\begin{align*}
	\frac{\partial \mathbf{Y}_{ij}}{\partial \mathbf{t}_r}
	&= \mathbf{R}_{li} \mathbf{f}_l \left(\frac{\partial \lambda_{lr}}{\partial \mathbf{a}}\right)^T, \\
	\frac{\partial \mathbf{Y}_{ij}}{\partial \boldsymbol{\varphi}_r} &= \mathbf{R}_{li} \mathbf{f}_l \left( \frac{\partial \lambda_{lr}}{\partial \boldsymbol{\varphi}_r} \right)^T + (\mathbf{R}_{li} \mathbf{v}_l + \mathbf{t}_{li} - \mathbf{v}_{ij}) \left( \frac{\partial {\theta}_{lr}}{\partial \boldsymbol{\varphi}_r} \right)^T,
\end{align*}
where 
\begin{align*}
	\frac{\partial \lambda_{lr}}{\partial \mathbf{a}} &= - \frac{1}{\lambda_{lr}} \mathbf{f}^{\wedge}_{r} \mathbf{f}_{r}^{\wedge} (\mathbf{t}_{lr} + \mathbf{R}_{lr} \mathbf{v}_l - \mathbf{v}_r), \\
	\frac{\partial \lambda_{lr}}{\partial \boldsymbol{\varphi}_r} &=
	\left[\mathbf{R}_r (\mathbf{R}_l^T \mathbf{t}_l)^{\wedge} - \mathbf{R}_r (\mathbf{R}_l^T \mathbf{v}_l)^{\wedge}\right]^T \frac{\partial \lambda_{lr}}{\partial \mathbf{a}}, \\
	\frac{\partial {\theta}_{lr}}{\partial \boldsymbol{\varphi}_r} &= \frac{1}{{\theta}_{lr}} \left[\mathbf{f}_{r}^T \mathbf{R}_{lr} \mathbf{f}_l \mathbf{f}_{r}^T \mathbf{R}_r (\mathbf{R}_l^T \mathbf{f}_l)^{\wedge}\right]^T.
\end{align*}
Based on the specific expressions of $\frac{\partial \mathbf{Y}_{ij}}{\partial \boldsymbol{\varphi}_l} $ and $\frac{\partial \mathbf{Y}_{ij}}{\partial \mathbf{t}_l}$, we eliminate redundant matrix operations. The following calculation techniques can be obtained:
\begin{align*}
	\frac{\partial \mathbf{Y}_{ij}}{\partial \boldsymbol{\varphi}_l} &= -\frac{\partial \mathbf{Y}_{ij}}{\partial \boldsymbol{\varphi}_l} - \frac{\partial \mathbf{Y}_{ij}}{\partial \boldsymbol{\varphi}_i}, \\
	\frac{\partial \mathbf{Y}_{ij}}{\partial \mathbf{t}_l} &= -\frac{\partial \mathbf{Y}_{ij}}{\partial \mathbf{t}_r} \mathbf{R}_{lr} - \theta_{lr}\mathbf{R}_{li}.
\end{align*}
The original calculation of derivatives $\frac{\partial \mathbf{Y}_{ij}}{\partial \boldsymbol{\varphi}_l} $ and $\frac{\partial \mathbf{Y}_{ij}}{\partial \mathbf{t}_l}$ is quite complex. We have found some calculation techniques to express their derivatives strictly as combinations of existing matrices. Our calculation techniques reduce the number of matrix operations by half, which helps to improve the construction speed of the normal equation.

\subsection{Selection of Base Observations}
\label{sec:SBO}
In the multi-camera pose-only constraint \cref{eq:SXW}, each 3D scene point is actually represented by a pair of base observations. Thus, a new question arises: how to select the optimal base observation pair for the 3D scene points? Cai et al. \cite{Cai2023pose} chose the two views with the maximum $\theta$ as base views, while Ge et al. \cite{Ge2024pipo} select two frames with the maximum disparity angle. Additionally, we note that there is also a step to choose the main and associate views (anchors) in ParallaxBA \cite{Zhao2015parallaxba}. In contrast to these methods, our generalized camera model discards the concept of views/frames and instead selects the base observations. Inspired by \cite{Beder2006determining}, we propose a selection algorithm based on the uncertainty ellipsoid roundness of the reconstructed 3D scene points. The observation pair with the maximum uncertainty ellipsoid roundness is selected as the base observations. In the following, we introduce the method for calculating the uncertainty ellipsoid roundness. 

\begin{figure}[tbp]
	\includegraphics[width=\linewidth]{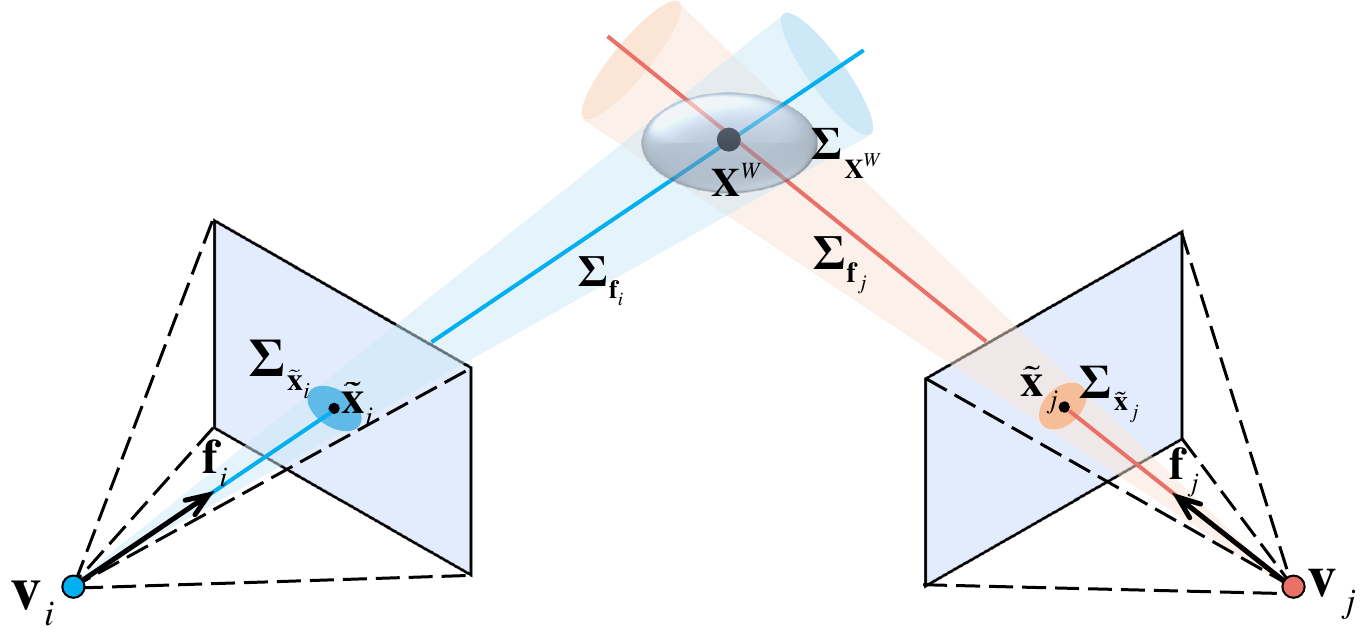} 
	\caption{Schematic diagram of the uncertainty ellipsoid for 3D reconstruction points. The uncertainty ellipsoid $\Sigma_{\mathbf{X}^W}$ for the 3D point $\mathbf{X}^W$ is calculated from the observed rays $(\mathbf{f}_i, \mathbf{v}_i)$ and $(\mathbf{f}_j, \mathbf{v}_j)$, along with their directional covariance $\mathbf{\Sigma}_{\mathbf{f}_i}$ and $\mathbf{\Sigma}_{\mathbf{f}_j}$. The roundness of the uncertainty ellipsoid can be a measure for evaluating the capability of the observed pair in representing the 3D point.
	}
	\label{fig:uncertain_ellipse}
\end{figure}
\begin{algorithm}[t]
	\caption{Base Observations Selection Algorithm} 
	\textbf{Input:} \\
	$\bullet$ Observation set of $\mathbf{X}^W$: $\{(\mathbf{f}_i, \mathbf{v}_i)\}$; \\
	$\bullet$ Corresponding pose set of observations: $\{(\mathbf{R}_i, \mathbf{t}_i)\}$; \\
	$\bullet$ Covariance matrix set of observation directions: $\{\Sigma_{\mathbf{f}_i}\}$;\\
	\textbf{Output:} \\
	$\bullet$ Left- and right-base observations: $(\mathbf{f}_l, \mathbf{v}_l)$ and $(\mathbf{f}_r, \mathbf{v}_r)$;
	\begin{algorithmic}[1]
		\Statex \textcolor{blue}{// Phase 1: Select left-base observation}
		\If{Set $\{\Sigma_{\mathbf{f}_i}\}$ is available}
		\For{$i = 1$ to $N$}
		\State Compute average variance: $\bar{\sigma}_{\mathbf{f}_i}^2 \leftarrow \text{trace}(\boldsymbol{\Sigma}_{\mathbf{f}_i})/3$;
		\EndFor
		\State $l \leftarrow \arg\min_{i} \bar{\sigma}_{\mathbf{f}_i}^2$; \hfill \textcolor{blue}{// Select minimum $\bar{\sigma}_{\mathbf{f}_i}^2$}
		\Else
		\State Initialize covariance matrix set: $\{\Sigma_{\mathbf{f}_i} = \mathbf{I}_3\}$;
		\State $l \leftarrow 1$; \hfill \textcolor{blue}{// Default to first observation}
		\EndIf
		\State Set left-base: $(\mathbf{f}_l, \mathbf{v}_l)$;
		\Statex \textcolor{blue}{// Phase 2: Select right-base observation}
		\State Initialize: $R_{\max} \leftarrow 0$, $r \leftarrow 1$;
		\For{$i = 1$ to $N$}
		\State Calculate the covariance matrix of $\mathbf{X}^W$ using observation pair $(l, i)$ from \cref{eq:AX=0}-\cref{eq:CXw};
		\State Calculate roundness $R_{li}$ using \cref{eq:rij};
		\If{$R_{li} > R_{max}$}
		\State $R_{\max} \leftarrow R_{li}$, $r \leftarrow i$; 
		\EndIf
		\EndFor
		\State Set right-base: $(\mathbf{f}_r, \mathbf{v}_r)$;
		\State \textbf{return} $(\mathbf{f}_l, \mathbf{v}_l)$ and $(\mathbf{f}_r, \mathbf{v}_r)$;
	\end{algorithmic}
	\label{alg:base_obs}
\end{algorithm}
As shown in \cref{fig:uncertain_ellipse}, let $(\mathbf{f}_i, \mathbf{v}_i)$ and $(\mathbf{f}_j, \mathbf{v}_j)$ be two observations of the 3D scene point $\mathbf{X}^W$, with known poses for each observation. Denote the covariance matrices of the direction vectors $\mathbf{f}_i$ and $\mathbf{f}_j$ as $\mathbf{\Sigma}_{\mathbf{f}_i}$ and $\mathbf{\Sigma}_{\mathbf{f}_j}$, respectively. The covariance of the direction vector $\mathbf{f}$ can be obtained through 2D observation $\tilde{\mathbf{x}}$ covariance propagation \cite{Hartley2003multiple}. Assuming the covariance of 2D observations is $\mathbf{\Sigma}_{\tilde{\mathbf{x}}}$, we obtain:
\begin{equation}
	\mathbf{\Sigma}_{\mathbf{f}} = \mathbf{R}_c (\mathbf{J}_{\pi_c^{-1}} \mathbf{\Sigma}_{\tilde{\mathbf{x}}} \mathbf{J}_{\pi_c^{-1}}^T) \mathbf{R}_c^T, 
	\label{eq:sigma_f}
\end{equation}
where $\mathbf{J}_{\pi_c^{-1}}$ is the Jacobian of the back projection function $\pi_c^{-1}$. From the projection equation \cref{eq:sfv}, we obtain the following two equations	:
\begin{align}
	s_i \mathbf{f}_i = \mathbf{P}_i \tilde{\mathbf{X}}^W, \quad s_j \mathbf{f}_j = \mathbf{P}_j \tilde{\mathbf{X}}^W,
	\label{eq:sf=px}
\end{align}
where $\mathbf{P}_i = [\mathbf{R}_i, \mathbf{t}_i - \mathbf{v}_i]$,  $\mathbf{P}_j = [\mathbf{R}_j, \mathbf{t}_j - \mathbf{v}_j]$ and $\tilde{\mathbf{X}}^W$ is the homogeneous form of $\mathbf{X}^W$. {For a given vector $\mathbf{v} = [v_1, v_2, v_3]^T$, we define an operator $\mathcal{K}(\mathbf{v})$ that extracts the first two rows of the skew-symmetric matrix $[\mathbf{v}]_\times$. From \cref{eq:sf=px}, we obtain the following linear system:
\begin{equation}
	\begin{bmatrix} \mathcal{K}(\mathbf{f}_i) \mathbf{P}_i \\ \mathcal{K}(\mathbf{f}_j) \mathbf{P}_j
	\end{bmatrix} \tilde{\mathbf{X}}^W = \mathbf{0}.
	\label{eq:AX=0}
\end{equation}
This system can be solved in a least-squares sense under the unit norm constraint $\|\tilde{\mathbf{X}}^W\| - 1 = 0$. Consequently, the Gauss–Helmert model \cite{helmert1872ausgleichungsrechnung,Corrochano2005uncertainty,forstner2016photogrammetric} with constraint can be employed to estimate the covariance matrix of $\tilde{\mathbf{X}}^W$. Following the notation in \cite{forstner2016photogrammetric}, we define the two constraint functions as:
\begin{equation}
	\boldsymbol{g}(\mathbf{l}, \tilde{\mathbf{X}}^W) = \begin{bmatrix} \mathcal{K}(\mathbf{f}_i) \mathbf{P}_i \\ \mathcal{K}(\mathbf{f}_j) \mathbf{P}_j.
	\end{bmatrix} \tilde{\mathbf{X}}^W = \mathbf{0}, \boldsymbol{h}(\tilde{\mathbf{X}}^W) = \|\tilde{\mathbf{X}}^W\| - 1 = 0,
\end{equation}
where $\mathbf{l} = [\mathbf{f}_i^T, \mathbf{f}_j^T]^T$ denotes the observation vector with its associated covariance matrix:
\begin{equation}
	\mathbf{\Sigma}_{\mathbf{l}} = \begin{bmatrix} \mathbf{\Sigma}_{\mathbf{f}_i} & \mathbf{0} \\ \mathbf{0} & \mathbf{\Sigma}_{\mathbf{f}_j}\end{bmatrix}.
\end{equation}
Specifically, $\boldsymbol{g}(\mathbf{l}, \tilde{\mathbf{X}}^W)$ represents the implicit functional constraints involving both unknown parameter and observations, and $\boldsymbol{h}(\tilde{\mathbf{X}}^W)$ imposes constraints solely on the unknowns. By performing a first-order Taylor expansion at $\widehat{\mathbf{X}}^W$, we linearize these functions as:
\begin{equation}
	\begin{aligned}
		&\boldsymbol{g}(\mathbf{l}, \tilde{\mathbf{X}}^W) = \boldsymbol{g}(\mathbf{l}, \widehat{\mathbf{X}}^W) + \mathbf{E} \Delta \tilde{\mathbf{X}}^W + \mathbf{F} \Delta{\mathbf{l}} = \mathbf{0},\\
		&\boldsymbol{h}(\tilde{\mathbf{X}}^W) = \boldsymbol{h}(\widehat{\mathbf{X}}^W) + \mathbf{H}^T \Delta \tilde{\mathbf{X}}^W =\mathbf{0},
	\end{aligned}
	\label{eq:linearize}
\end{equation}
where 
\begin{equation}
	\begin{aligned}
		\mathbf{E} &= \left. \frac{\partial \boldsymbol{g}(\mathbf{l}, \tilde{\mathbf{X}}^W)}{\partial \tilde{\mathbf{X}}^W}\right|_{\widehat{\mathbf{X}}^W} = \begin{bmatrix} \mathcal{K}(\mathbf{f}_i) \mathbf{P}_i \\ \mathcal{K}(\mathbf{f}_j) \mathbf{P}_j
		\end{bmatrix},\\
		\mathbf{F} &= \left. \frac{\partial \boldsymbol{g}(\mathbf{l}, \tilde{\mathbf{X}}^W)}{\partial \mathbf{l}} \right|_{\widehat{\mathbf{X}}^W} = \begin{bmatrix} \mathcal{K}(\mathbf{P}_i \widehat{\mathbf{X}}^W) & \mathbf{0} \\ \mathbf{0} & \mathcal{K}(\mathbf{P}_j \widehat{\mathbf{X}}^W)
		\end{bmatrix},\\
		\mathbf{H} &= \left. \frac{\partial \boldsymbol{h}(\tilde{\mathbf{X}}^W)}{\partial \tilde{\mathbf{X}}^W}\right|_{\widehat{\mathbf{X}}^W} = \widehat{\mathbf{X}}^W.
	\end{aligned}
\end{equation}
Here, $\widehat{\mathbf{X}}^W$ is the estimated value obtained from \cref{eq:AX=0}. Since the derivation of Gauss–Helmert solution is relatively lengthy, we omit  the intermediate steps and refer to \cite{forstner2016photogrammetric} for further technical details and proofs. Let $\mathbf{N} = \mathbf{E}^T(\mathbf{F} \mathbf{\Sigma}_{\mathbf{l}} \mathbf{F}^T)^{-1}\mathbf{E}$, the resulting covariance matrix for $\tilde{\mathbf{X}}^W$ is given by:
\begin{equation}
	\mathbf{\Sigma}_{\tilde{\mathbf{X}}^W} = \mathbf{N}^{-1} - \mathbf{N}^{-1}\mathbf{H}(\mathbf{H}^T\mathbf{N}^{-1}\mathbf{H})^{-1}\mathbf{H}^T\mathbf{N}^{-1}.
\end{equation}
Since $\mathbf{\Sigma}_{\tilde{\mathbf{X}}^W}$ is singular, we need to convert the homogeneous vector $\tilde{\mathbf{X}}^W$ to ${\mathbf{X}}^W$. The covariance of ${\mathbf{X}}^W$ is calculated as:

\begin{equation}
	\mathbf{\Sigma}_{{\mathbf{X}}^W} = \frac{\partial{\tilde{\mathbf{X}}^W}}{\partial{{\mathbf{X}}^W}} \mathbf{\Sigma}_{\tilde{\mathbf{X}}^W} \left(\frac{\partial{\tilde{\mathbf{X}}^W}}{\partial{{\mathbf{X}}^W}}\right)^T, \frac{\partial{\tilde{\mathbf{X}}^W}}{\partial{{\mathbf{X}}^W}} = \frac{1}{\tilde{X}_4^W}[\mathbf{I}_3, -\mathbf{X}^W]
	\label{eq:CXw}
\end{equation}
{where $\tilde{X}_4^W$ is the fourth component of $\tilde{\mathbf{X}}^W$.} Thus, we have recovered the covariance matrix of the 3D scene point from the observations. The matrix $\mathbf{\Sigma}_{{\mathbf{X}}^W}$ geometrically represents the spatial uncertainty of ${\mathbf{X}}^W$, reflecting the reconstruction quality from the observation pair $(i, j)$. The uncertainty ellipsoid can be approximately obtained from $\mathbf{\Sigma}_{{\mathbf{X}}^W}$. We further perform the eigendecomposition on $\mathbf{\Sigma}_{{\mathbf{X}}^W} =  \mathbf{U} \text{diag}(\lambda_1, \lambda_2, \lambda_3) \mathbf{V}^T$. The uncertainty ellipsoid roundness of ${\mathbf{X}}^W$ reconstructed from observation pair $(i,j)$ is calculated as:
\begin{equation}
	R_{ij} = \sqrt{\lambda_3/\lambda_1}. 
	\label{eq:rij}
\end{equation}
The roundness $R_{ij}$ ranges from 0 to 1 and exhibits scale invariance. In the following, we propose a base observations selection algorithm based on the uncertainty ellipsoid roundness. Since exhaustively checking all observation pairs is computationally expensive, our algorithm adopts a two-stage approach. First, the observation with the minimum average variance is selected as the left-base observation. The average variance of observation $\mathbf{f}$ is calculated as:
\begin{equation}
	\bar{\sigma}_{\mathbf{f}}^2 = \frac{1}{3} \text{trace}(\mathbf{\Sigma}_{\mathbf{f}}).	
	\label{eq:avg_var}
\end{equation}
Then, we iterate through the remaining observations. The observation that, when combined with the left-base observation, forms an uncertainty ellipsoid with the maximum roundness is selected as the right-base observation. \Cref{alg:base_obs} summarizes the detailed steps for selecting base observations.

\subsection{Statistical Optimal Reconstruction}
The proposed algorithm focuses solely on pose optimization for multi-camera systems. However, 3D scene reconstruction is also a necessary component of a complete 3D visual geometric computing and plays a crucial role in applications such as SLAM \cite{Urban2016multicol} and SFM \cite{Cui2023mcsfm}. In this section, we introduce a statistically optimal triangulation method based on the generalized camera model, aiming at 3D scene reconstruction. Our algorithm accounts for the covariance matrix of the observations and incorporates it into the cost function.

Let $\mathbf{X}^W$ denote the 3D scene point to be reconstructed, which corresponds to the observation set $\{(\mathbf{f}_i, \mathbf{v}_i) | 1 \leq i \leq N \}$ from the multi-camera system. The accurate pose of each observation is obtained using pose adjustment algorithm. Then, we can determine $\mathbf{X}^W$ by minimizing the sum of its Euclidean distances to all observation rays:
\begin{equation}
	\mathbf{X}^W = {\arg \min } \sum_{i=1}^{N} \| (\mathbf{I}_3 - \mathbf{f}_i \mathbf{f}_i^T)(\mathbf{R}_i \mathbf{X}^W + \mathbf{t}_i - \mathbf{v}_i) \|^2.
	\label{eq:eXW}
\end{equation}
However, this solution is geometrically optimal but not statistically optimal for the reconstruction problem. In our triangulation method, the uncertainty of the observation direction $\mathbf{f}$ is considered. The covariance matrix of $\mathbf{f}_i$ is given by $\mathbf{\Sigma}_{\mathbf{f}_i}$, which can be computed via \cref{eq:sigma_f}. However, the covariance matrix $\mathbf{\Sigma}_{\mathbf{f}_i}$ is inherently singular because the spherically normalized vector $\mathbf{f}_i$ possesses only two degrees of freedom. Directly using  $\mathbf{\Sigma}_{\mathbf{f}_i}$ to model the three residual components in \cref{eq:eXW} would introduce a redundant dimension, leading to theoretical inconsistency.

To eliminate the redundancy in both the covariance information and the distance residuals, we introduce the null space of spherically normalized vector \cite{forstner2010minimal}. Specifically, the null space of $\mathbf{f}_i$ is computed as:
\begin{equation}
	\mathbf{J}_{\mathbf{f}_i} = \text{null}(\mathbf{f}_i^T) = \begin{bmatrix}
		\mathbf{r} & \mathbf{s}
	\end{bmatrix}.
\end{equation} 
where the function $\text{null}(\cdot)$ performs a singular value decomposition of vector and extracts the right singular vectors corresponding to the two zero eigenvalues, yielding an $3\times2$ orthogonal matrix. This null space spans the tangent plane of the unit sphere at $\mathbf{f}_i$, i.e., the two-dimensional subspace consisting of all directions orthogonal to $\mathbf{f}_i$. Within this tangent space, the original 3D residual can be projected into a 2D residual:
\begin{equation}
	\mathbf{e}_i = \mathbf{J}_{\mathbf{f}_i}^T \mathbf{f}_i = \mathbf{J}_{\mathbf{f}_i}^T (\mathbf{R}_i \mathbf{X}^W + \mathbf{t}_i - \mathbf{v}_i).
	\label{eq:eiJf}
\end{equation}
The covariance matrix is accordingly transformed to:
\begin{equation}
	\mathbf{\Sigma}_{\mathbf{e}_i} = \mathbf{J}_{\mathbf{f}_i}^T \mathbf{\Sigma}_{\mathbf{f}_i} \mathbf{J}_{\mathbf{f}_i}.
\end{equation}

We then define an optimization problem that aims at determining the 3D point $\mathbf{X}^W$ by minimizing the sum of Mahalanobis distances between the 3D point and all observation rays. The cost function is defined as:
\begin{equation}
	{\mathbf{X}}^W = {\arg \min } \sum_{i=1}^{N} \| \mathbf{e}_i \|_{\Sigma_{\mathbf{e}_i}^{-1}}^2 = {\arg \min } \sum_{i=1}^{N} \mathbf{e}_i^T \Sigma_{\mathbf{e}_i}^{-1} \mathbf{e}_i,
	\label{eq:eCe}
\end{equation}
where $\|\cdot\|_{\Sigma}$ denotes Mahalanobis distance with covariance $\Sigma$. Substituting \cref{eq:eiJf} into \cref{eq:eCe} and setting the derivative of the cost function with respect to $\mathbf{X}^W$ to $\mathbf{0}$, we have:
\begin{equation}
	\sum_{i=1}^{N} \mathbf{A}_i^T \Sigma_{\mathbf{f}_i}^{-1} (\mathbf{A}_i \mathbf{X}^W + \mathbf{B}_i) = \mathbf{0},
\end{equation}
{where $\mathbf{A}_i = \mathbf{J}_{\mathbf{f}_i}^T \mathbf{R}_i$ and $\mathbf{B}_i = \mathbf{J}_{\mathbf{f}_i}^T (\mathbf{t}_i - \mathbf{v}_i)$.} Let 
\begin{equation*}
	\mathbf{A} = \sum_{i=1}^{N} \mathbf{A}_i^T \Sigma_{\mathbf{e}_i}^{-1} \mathbf{A}_i, \quad \mathbf{B} = \sum_{i=1}^{N} \mathbf{A}_i^T \Sigma_{\mathbf{e}_i}^{-1} \mathbf{B}_i,
\end{equation*}
the 3D point $\mathbf{X}^W$ is determined as:
\begin{equation}
	\mathbf{X}^W = -\mathbf{A}^{-1} \mathbf{B}.
\end{equation}
Our method provides a statistically optimal closed-form solution to the reconstruction problem for the generalized camera model. The geometric error, as defined by \cref{eq:eXW}, is equivalent to the residual of the midpoint method for generalized camera models proposed in \cite{Ramalingam2006generic}. When covariance is ignored, our method achieves the same accuracy as the midpoint method. However, we avoid complex derivations and large-scale matrix operations. Incorporating the covariance of the observations into the cost function enables the proposed method to achieve statistical optimality.

\section{Experiments}
\label{sec:experiment}
In this section, we evaluate the proposed method on both synthetic data and real datasets. The experiments are performed on a laptop with an Intel Core i5-12500H @ 2.5GHz CPU and 16GB of memory. We compare the following pose optimization algorithms in our evaluation.
\begin{itemize}
	\item \texttt{MultiCol} \cite{Urban2017multicol} is the standard multi-camera bundle adjustment method, and its cost function is defined as the sum of reprojection errors on 2D images.
	\item \texttt{BACS} \cite{Schneider2012bundle} represents both the scene and image points as minimum homogeneous form, with its cost function defined as the sum of distances between the spherical normalized vectors.
	\item \texttt{MCPA} only uses the left-base observation as primary to construct the residual term, with its cost function explicitly defined in \cref{eq:argmin}. 
	\item \texttt{MCPALR} uses both left- and right-base observations as primary to construct residual term, with its cost function explicitly defined in \cref{eq:mineler}. 
\end{itemize}
\texttt{MCPA} and \texttt{MCPALR} are both proposed methods in this paper, differing only in the residual term construction step. 
In the experiments, \texttt{MultiCol} employs the pinhole camera model instead of the omnidirectional model \cite{Scaramuzza2006Omnidirectional}. This change simplifies the calculation process while maintaining the reliability and validity of the evaluation results.

For quantitative evaluation of pose optimization accuracy, we use two standard metrics: average rotation error $\bar{\varepsilon}_\mathbf{R}$ and average translation error $\bar{\varepsilon}_\mathbf{t}$. Given the ground truth of pose set $\{(\mathbf{R}_i^{gt}, \mathbf{t}_i^{gt}) | 1 \leq i \leq N\}$ and the estimated pose set $\{(\mathbf{R}_i, \mathbf{t}_i) | 1 \leq i \leq N\}$, average rotation and translation errors are calculated separately as follows: $\bar{\varepsilon}_\mathbf{R} = \frac{1}{N}\sum_{i=1}^{N} \arccos((\text{trace}(\mathbf{R}_i^{gt}\mathbf{R}_i^T)-1)/2)$ and $\bar{\varepsilon}_\mathbf{t} = \frac{1}{N}\sum_{i=1}^{N} \|(\mathbf{t}_i^{gt} - \mathbf{t}_i)\| / \|\mathbf{t}_i^{gt}\|$. The average reprojection error $\bar{\varepsilon}_{p}$ is used as an indirect metric to evaluate the reconstruction quality. In synthetic data experiments, the reconstruction accuracy is quantified by the average spatial distance error $\bar{\varepsilon}_{\mathbf{X}}$ between the reconstructed points and the ground truth.

\subsection{Results on Synthetic Data}
In synthetic experiments, a virtual multi-camera system observing 3D scene points is simulated along predefined trajectories to generate observation data.

\subsubsection{Data Generation}
The virtual multi-camera system designed for the experiments consists of four cameras. For convenience, each camera is represented using the pinhole model, with a focal length set to 540 pixels and a resolution set to 1080×960 pixels. Two extrinsic configurations are designed, referred to as the \textit{forward} and the \textit{omnidirectional}. In the forward configuration, the four cameras are arranged side-by-side with their optical axes aligned in the same direction, as shown on the left of \cref{fig:Forward_Omni}. This configuration provides rich stereo vision details, which are beneficial for 3D reconstruction tasks. In the omnidirectional configuration, the four cameras are positioned at different locations, with their optical axes pointing in four different directions, as shown on the right of \cref{fig:Forward_Omni}. This configuration provides a broader field of view for autonomous driving scenarios, which helps to improve the robustness of localization. 
\begin{figure}[tbp]
	\includegraphics[width=\linewidth]{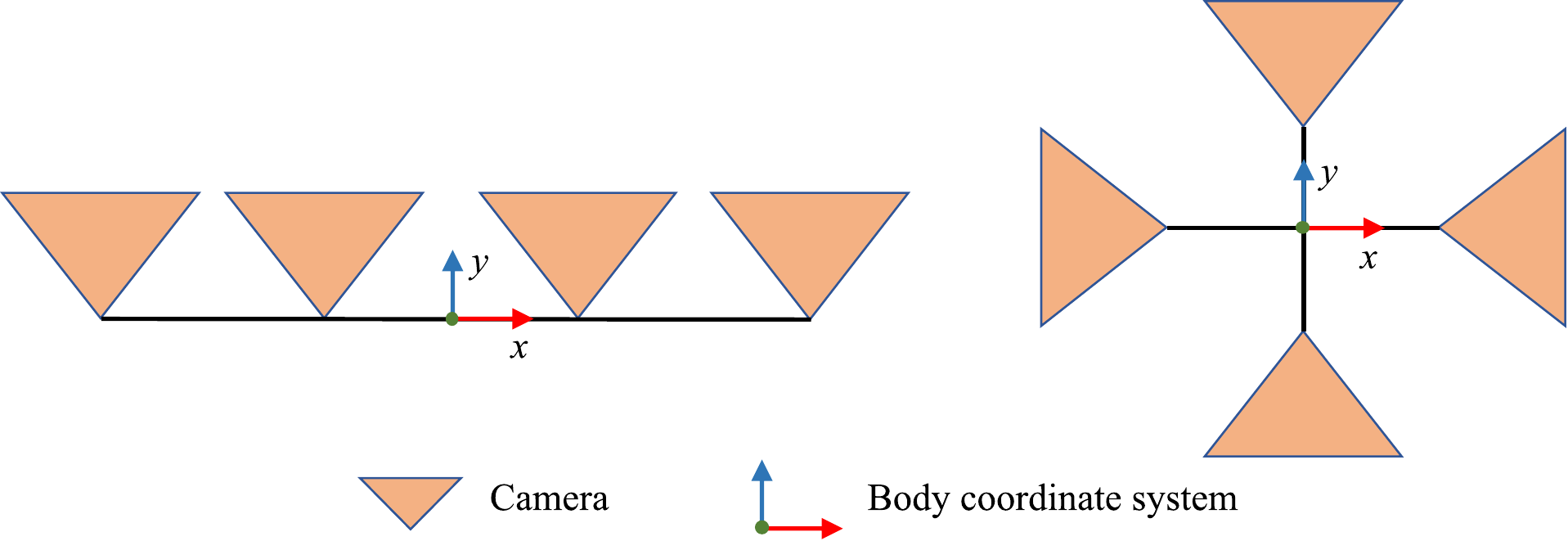} 
	\caption{Two types of extrinsic configurations for the multi-camera system used in the synthetic data experiments. \textbf{Left} is the forward configuration, where all cameras observe the same direction. \textbf{Right} is the omnidirectional configuration, where four cameras observe four different directions.}
	\label{fig:Forward_Omni}
\end{figure}
\begin{figure}[tbp]
	\includegraphics[width=\linewidth]{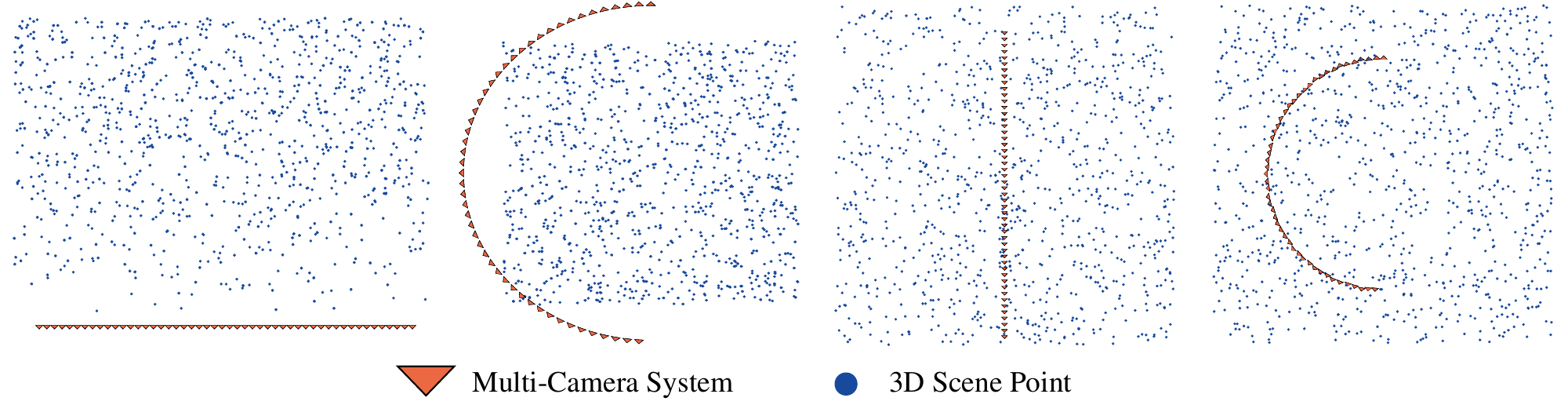} 
	\caption{Multi-camera system motion trajectories and 3D scene points used in the synthetic data experiments. The combination of two multi-camera system configurations and two motion trajectories results in four synthetic datasets. From left to right, they are labeled \textit{Forward-Linear}, \textit{Forward-Curve}, \textit{Omni-Linear}, and \textit{Omni-Curve} respectively.}
	\label{fig:Four_conf}
\end{figure}

In the 3D space $[-500m, 500m]^3$,  scene points are randomly generated, and the multi-camera system moves along predefined trajectories to capture image observations. We defined two types of trajectories: \textit{linear} and \textit{curve} motion. Linear motion requires the multi-camera system to move along a straight-line trajectory, while curve motion mandates movement along a circular trajectory. Based on these two motion types and different multi-camera system configurations, we generated four synthetic datasets: \textit{Forward-Linear}, \textit{Forward-Curve}, \textit{Omni-Linear}, and \textit{Omni-Curve}. \Cref{fig:Four_conf} illustrates the configurations of four synthetic datasets. 

Subsequently, the 3D scene points are projected onto the image according to the predefined multi-camera system parameters and motion trajectories. Considering practical conditions, we add zero-mean Gaussian noise with varying magnitude to the 2D image points. The noise standard deviation $\sigma$ is randomly sampled from a uniform distribution $\mathcal{U}(0, \sigma_{max})$, where $\sigma_{max}$ denotes the maximum standard deviation. Accordingly, the covariance matrix of the image point is $\mathrm{diag}(\sigma^2, \sigma^2)$. Following \cite{Ye2022coli}, the initial poses are obtained by perturbing the ground truth with noise. We add a perturbation of $2^\circ$ with an arbitrary direction to the rotation, and Gaussian noise of $0.5m$ to the translation. 
For fair comparison, the initial 3D scene points are obtained by multi-view triangulation \cite{Hartley2003multiple} using the perturbed poses, rather than by directly adding noise.

\subsubsection{Base Observations Selection Results}
This experiment validates the advantages of our proposed base observations selection strategy and reveals the reasons why our pose adjustment achieves better accuracy than traditional bundle adjustment. We compare four base observation selections strategies. The \texttt{Random-Selection} strategy selects two observations randomly from all available ones as the base observations. The \texttt{Maximum-Disparity} strategy, adopted in PIPO-SLAM \cite{Ge2024pipo} and ParallaxBA \cite{Zhao2015parallaxba}, selects two observations with the largest disparity angle as base. The \texttt{Maximum-Theta} strategy, proposed in \cite{Cai2023pose}, selects two observations with the largest $\theta_{lr}$ as the bases, where $\theta_{lr}$ is calculated using \cref{eq:sl}. The proposed strategy is named \texttt{Maximum-Roundness}. Triangulation is performed using two base observations \cite{Ramalingam2006generic}. The average Euclidean distance between the triangulated 3D points and their ground truth is used to evaluate the performance of selection strategies quantitatively. Additionally, we compare the results of triangulating 3D points using all observations, called \texttt{Multi-Observation}. This method corresponds to the case where 3D points are represented by using all observations within bundle adjustment. We conduct validations on \textit{Forward-Linear} and \textit{Forward-Curve} synthetic datasets. The maximum standard deviation $\sigma_{max}$ of Gaussian noise added to the 2D image points varies from 1 to 10 pixels. For each maximum noise standard deviation, we perform 1000 independent trials and report the median results. 
\begin{figure}[tbp]
	\centering
	\includegraphics[width=0.8\linewidth]{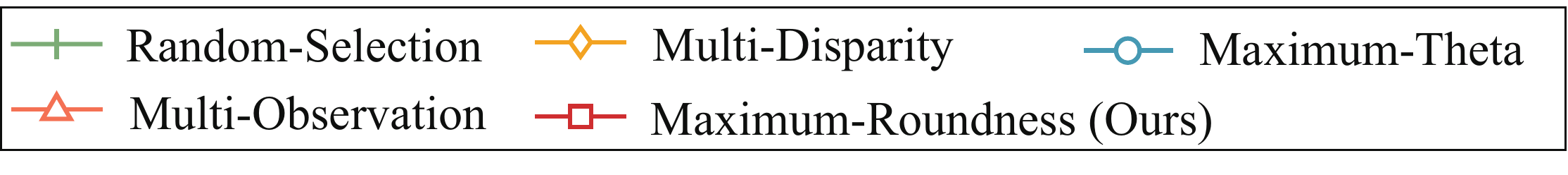} \\
	\subfloat[Forward-Linear\label{fig:Base_Linear}]{%
		\centering
		\includegraphics[width=0.48\linewidth]{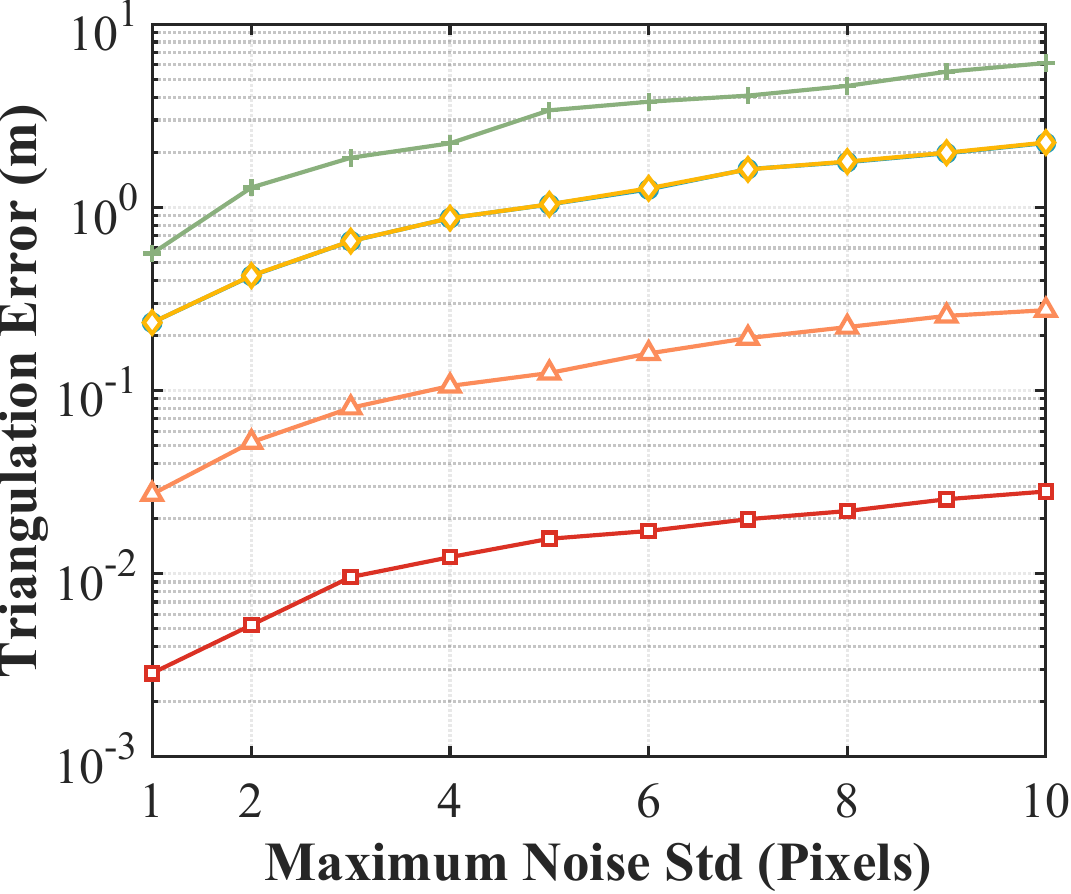}%
	}
	\subfloat[Forward-Curve\label{fig:Base_Curve}]{%
		\centering
		\includegraphics[width=0.48\linewidth]{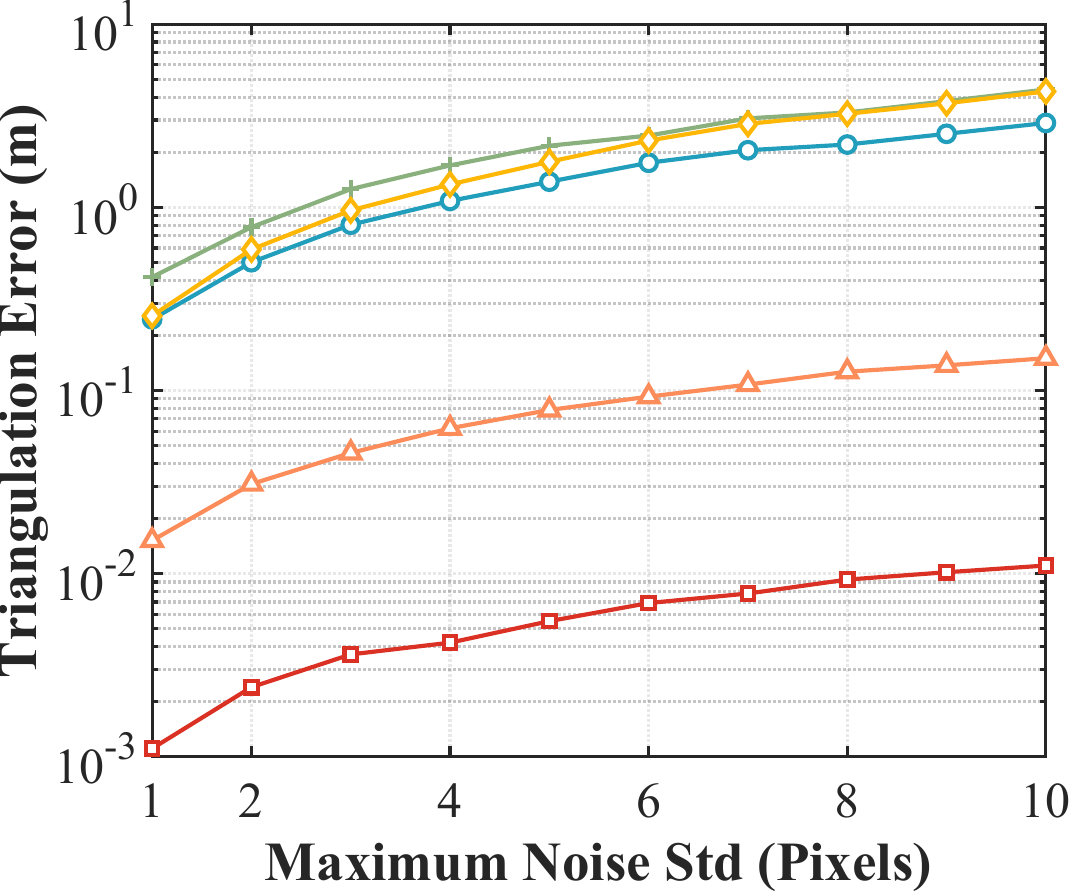}%
	}
	\caption{{Comparison results of four base observation selection strategies on the (a) \textit{Forward-Linear} and (b) \textit{Forward-Curve} datasets. Results show the median triangulation error over 1,000 independent trials, with the y-axis in logarithmic scale. Our method significantly outperforms the other three strategies and surpasses joint triangulation using all observations, providing experimental validation for pose-only constraint.}}
	\label{fig:Base_Selection}
\end{figure}

The comparison results of the base observation selection experiments are presented in \cref{fig:Base_Selection}. The overall trend indicates that the triangulation error increases progressively with the rise of the maximum noise standard deviation. Across all noise levels, the proposed strategy outperforms other strategies, maintaining triangulation errors at consistently low levels. Even under high noise levels, our method effectively identifies the best two base observations.
Strategies \texttt{Maximum-Disparity} and \texttt{Maximum-Theta} significantly outperform \texttt{Random-Selection}, but their performance remains inferior to our proposed strategy. The comparison with \texttt{Multi-Observation} validates a non-intuitive fact that 3D points triangulated from two well-selected observations can achieve higher accuracy than those triangulated from multiple observations. Thus, it can be inferred that our base observation selection strategy provides experimental justification for the pose adjustment to achieve accuracy superior to bundle adjustment.

\subsubsection{{Reconstruction Results}}
{This experiment aims to validate the accuracy advantages of the proposed statistically optimal reconstruction. We evaluate our proposed statistically optimal triangulation (\texttt{SOT (Ours)}) against six mainstream and widely adopted triangulation approaches on \textit{Forward-Linear} and \textit{Forward-Curve} synthetic datasets. The evaluated baselines include the \texttt{MidPoint}\cite{Ramalingam2006generic} as implemented in OpenSfM; the linear solver \texttt{LinearLS} used in COLMAP\cite{Johannes2016colmap}, originally proposed by Hartley et al.\cite{Hartley1997triangulation}; and the \texttt{DLT} method \cite{Hartley2003multiple}, commonly adopted in systems such as ORB-SLAM2\cite{Mur2017orbslam2}. Additionally, we include the nonlinear optimization approach (\texttt{NonLin}), the iteratively reweighted least-squares midpoint method (\texttt{IRMP})\cite{yang2019iteratively}, and the \texttt{RANSAC} triangulation strategy commonly used for robust estimation. Gaussian noise with the maximum standard deviation ($\sigma_{max}$) ranging from 1 to 10 pixels is added to the image points. For each noise level, we perform 1,000 independent Monte Carlo trials and report the median error.}
\begin{figure}[tbp]
	\centering
	\subfloat[Forward-Linear\label{fig:Triang_Linear}]{%
		\centering
		\includegraphics[width=0.48\linewidth]{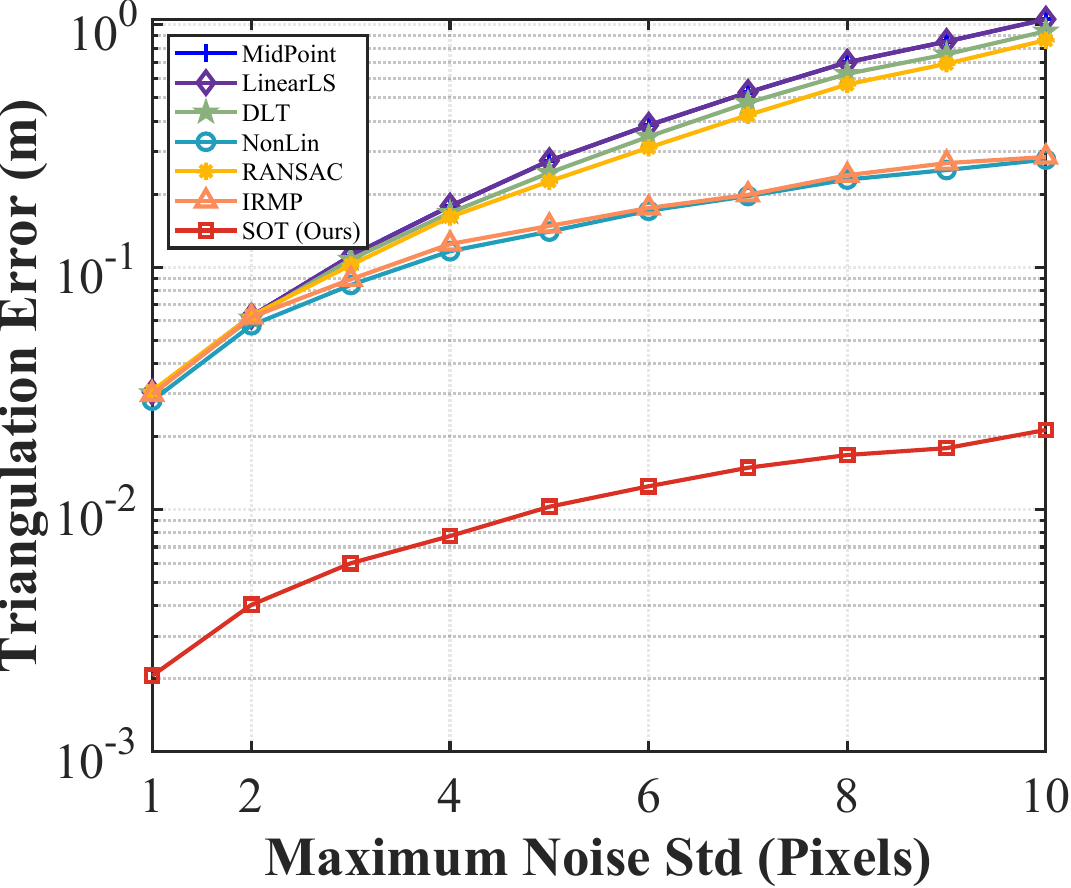}%
	}
	\subfloat[Forward-Curve\label{fig:Triang_Curve}]{%
		\centering
		\includegraphics[width=0.48\linewidth]{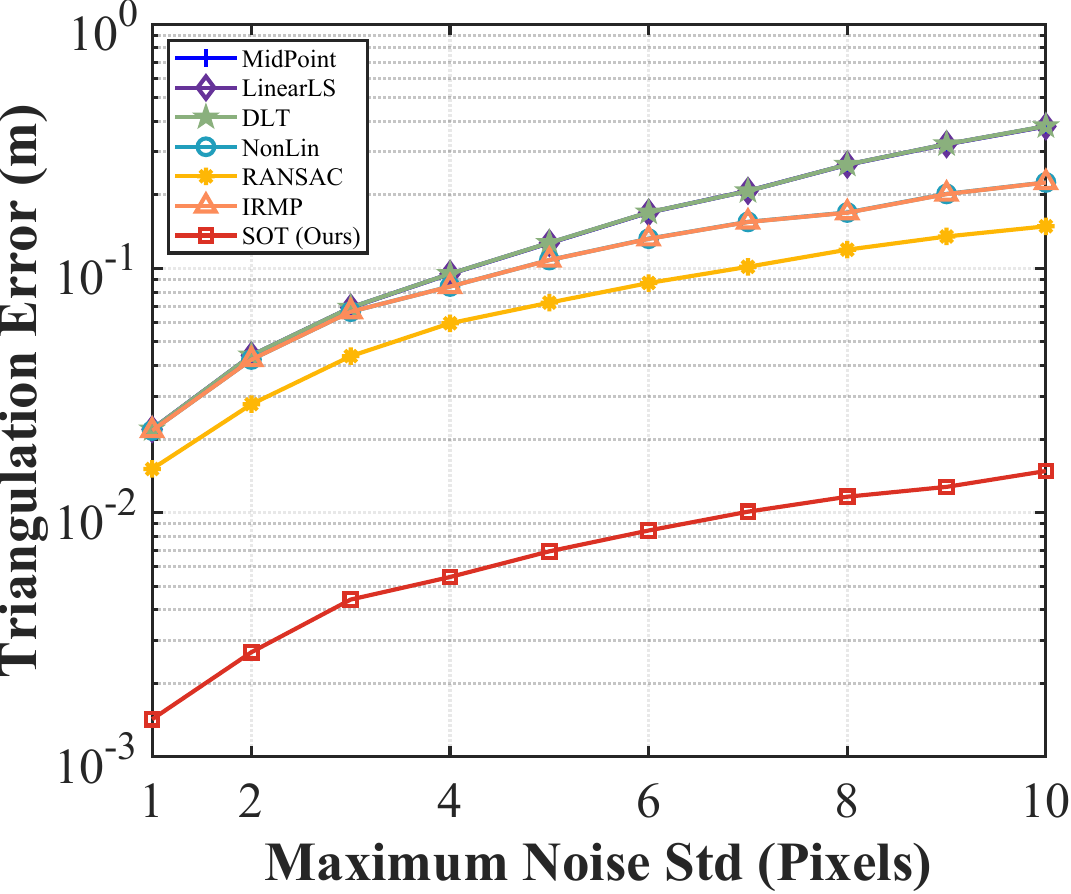}%
	}
	\caption{{Accuracy comparison of triangulation methods on the (a) \textit{Forward-Linear} and (b) \textit{Forward-Curve} datasets. Results show the median triangulation error over 1,000 independent trials, with the y-axis in logarithmic scale. The proposed method achieves the best accuracy across all noise levels.}}
	\label{fig:triang_result}
\end{figure}

{The comparison of reconstruction results is shown in \cref{fig:triang_result}. Across both datasets, the proposed statistically optimal reconstruction consistently achieves the lowest error across all noise levels, with its performance advantage becoming increasingly pronounced as the noise intensifies. Under the high-noise condition ($\sigma_{max} = 10$), \texttt{SOT (Ours)} yields an improvement in precision by an order of magnitude compared to the second-best alternatives, \texttt{NonLin} and \texttt{IRMP}. The performance gain stems from the core design of our algorithm. We explicitly incorporate the full covariance of 2D observations into the optimization objective via the Mahalanobis distance, which enables the reconstruction to adaptively suppress measurements with high uncertainty.}

\subsubsection{Pose Optimization Results}
In this experiment, we compare the performance of four pose optimization algorithms on synthetic datasets under different image noise levels. In each dataset, the motion trajectory of multi-camera system consists of 50 sampling poses, resulting in $4 \times 50$ images. A total of 1,000 3D scene points are randomly and uniformly generated within the space. Gaussian noise with maximum standard deviations $\sigma_{max}$ of $[1, 2, 4, 6, 8, 10]$ pixels is added to the 2D image points. For each noise level, 100 independent Monte Carlo trials are conducted, and the statistical results are reported. 
\begin{figure*}[tbp]
	\centering
	\includegraphics[width=0.6\linewidth]{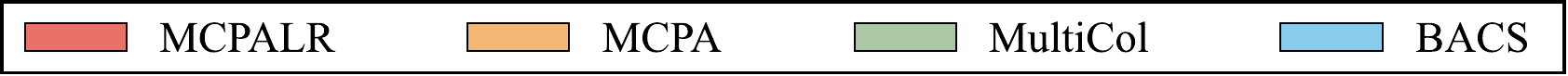} \\
	\begin{subfigure}{0.85\linewidth}
		\centering
		\includegraphics[width=0.3\linewidth]{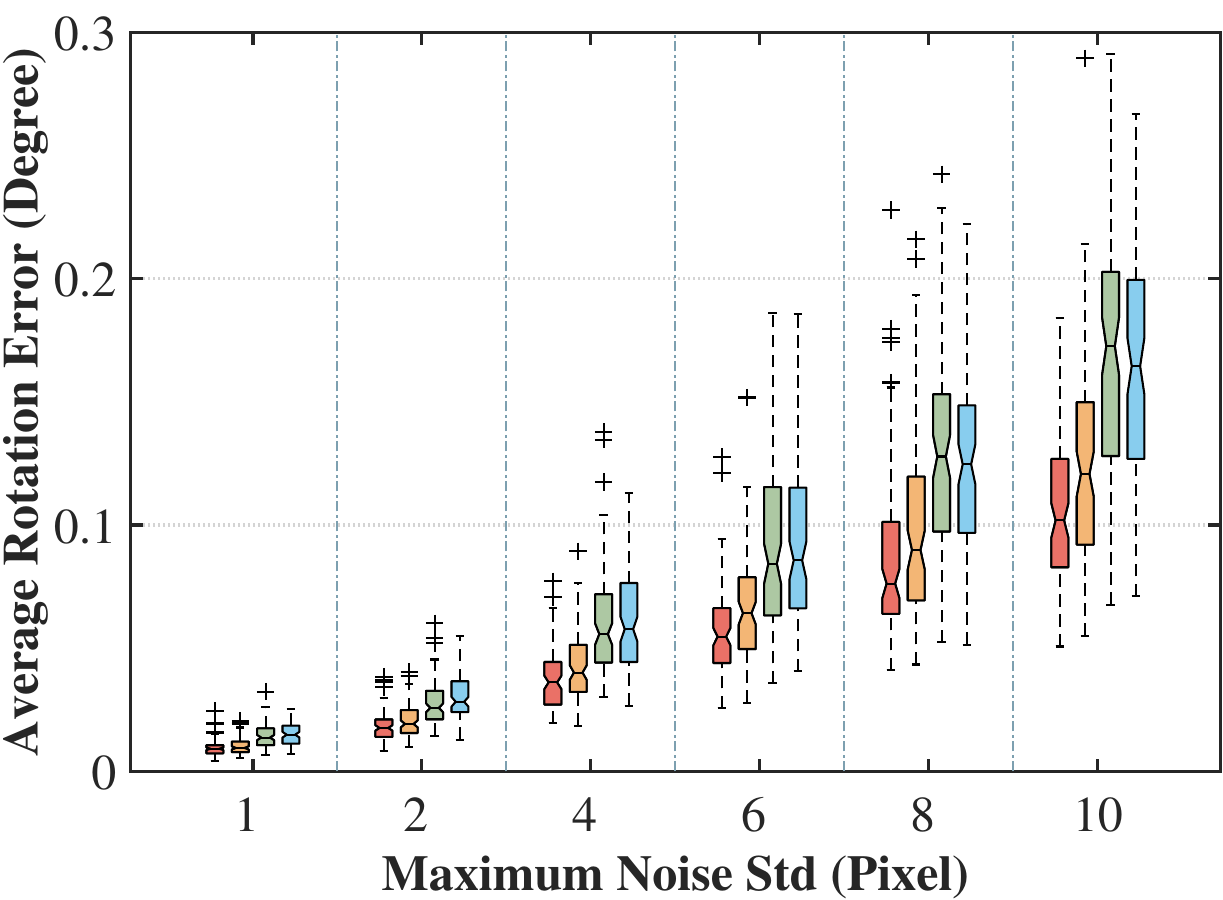} \qquad
		\includegraphics[width=0.3\linewidth]{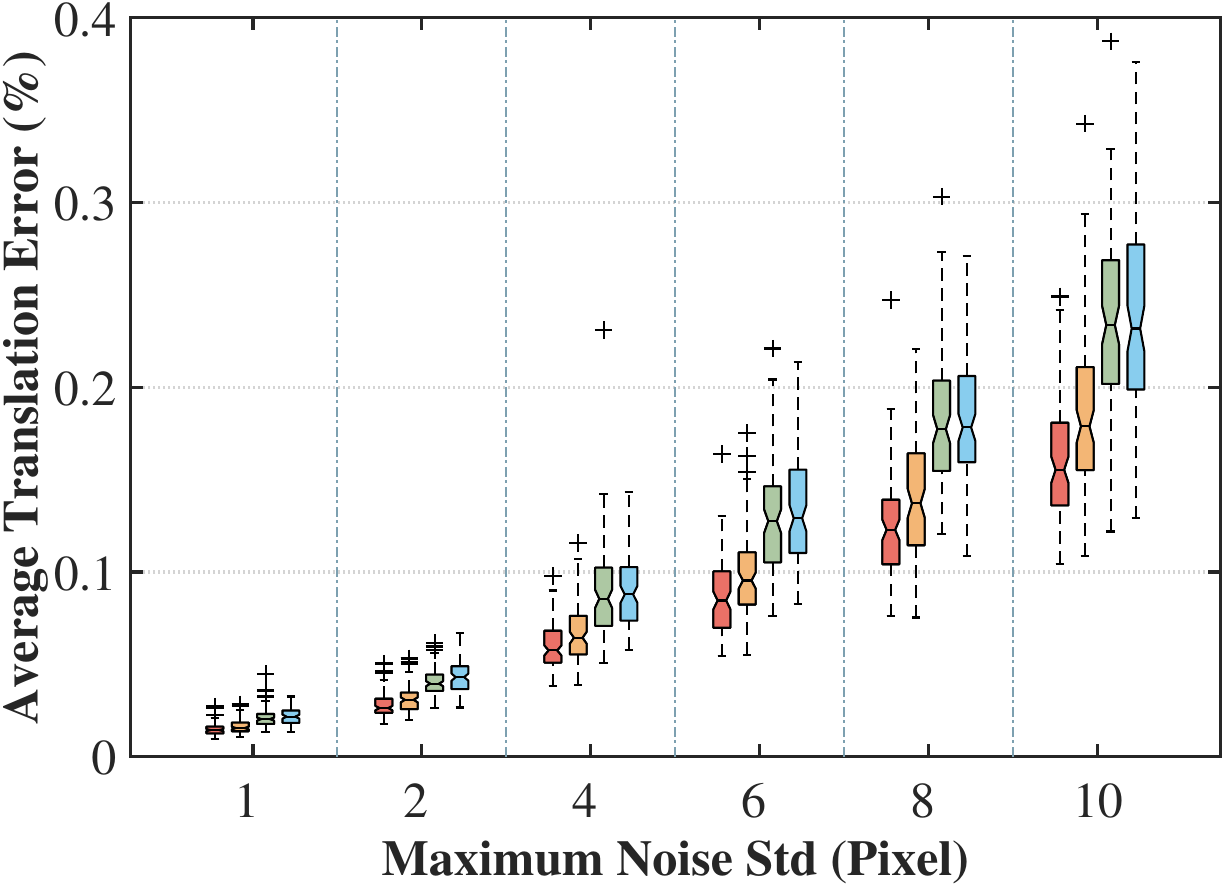} \qquad
		\includegraphics[width=0.3\linewidth]{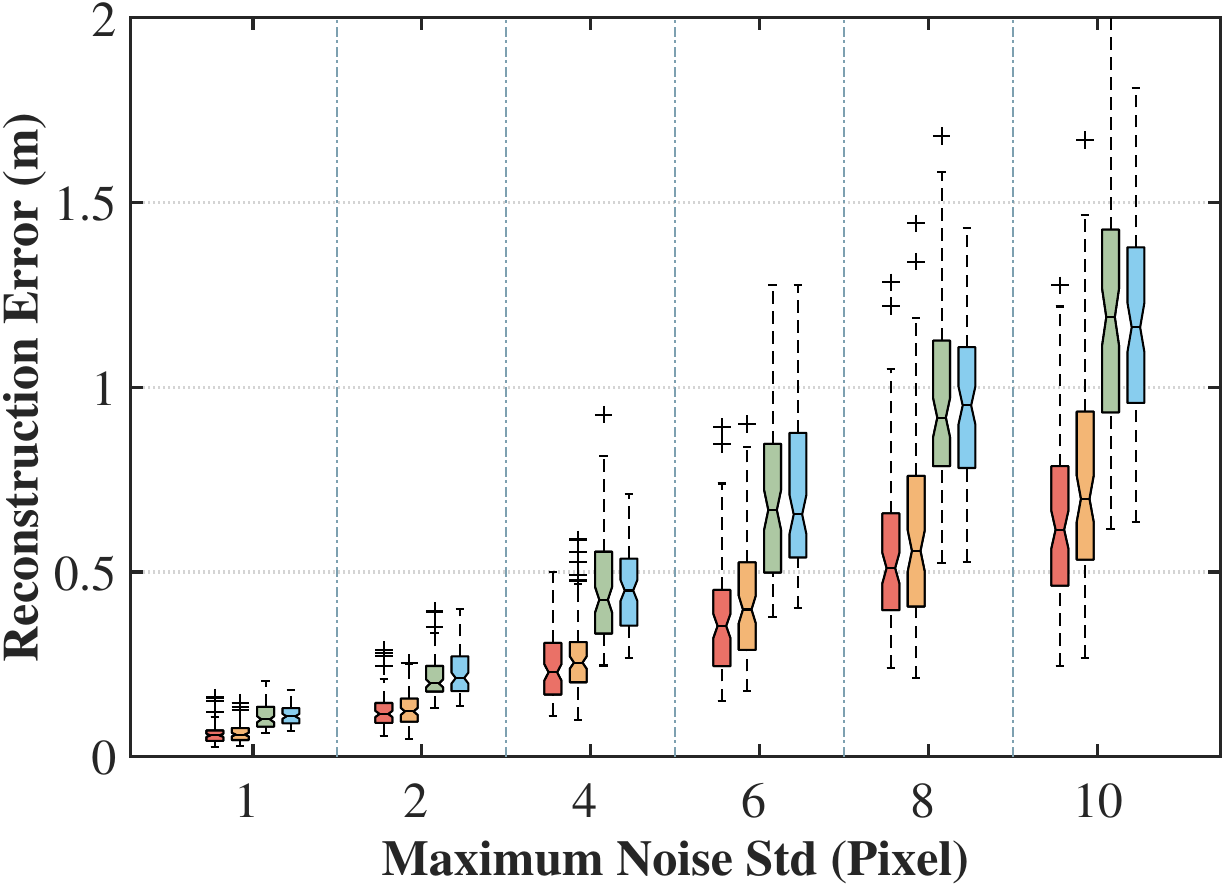} 
		\caption{Algorithm accuracy comparison in \textit{Forward-Linear}}
		\label{fig:Forward_Linear_Noise}
	\end{subfigure}
	\begin{subfigure}{0.85\linewidth}
		\centering
		\includegraphics[width=0.3\linewidth]{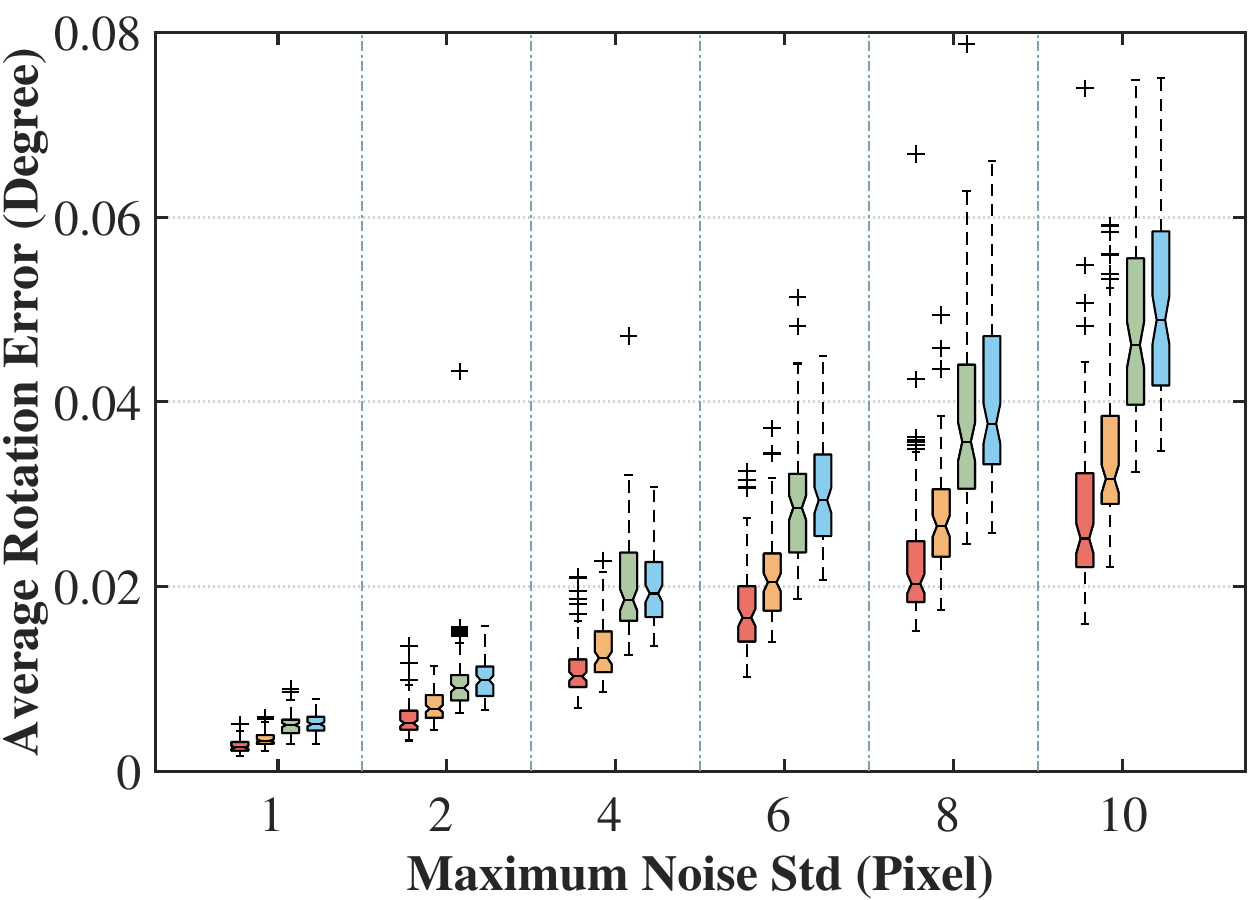} \qquad
		\includegraphics[width=0.3\linewidth]{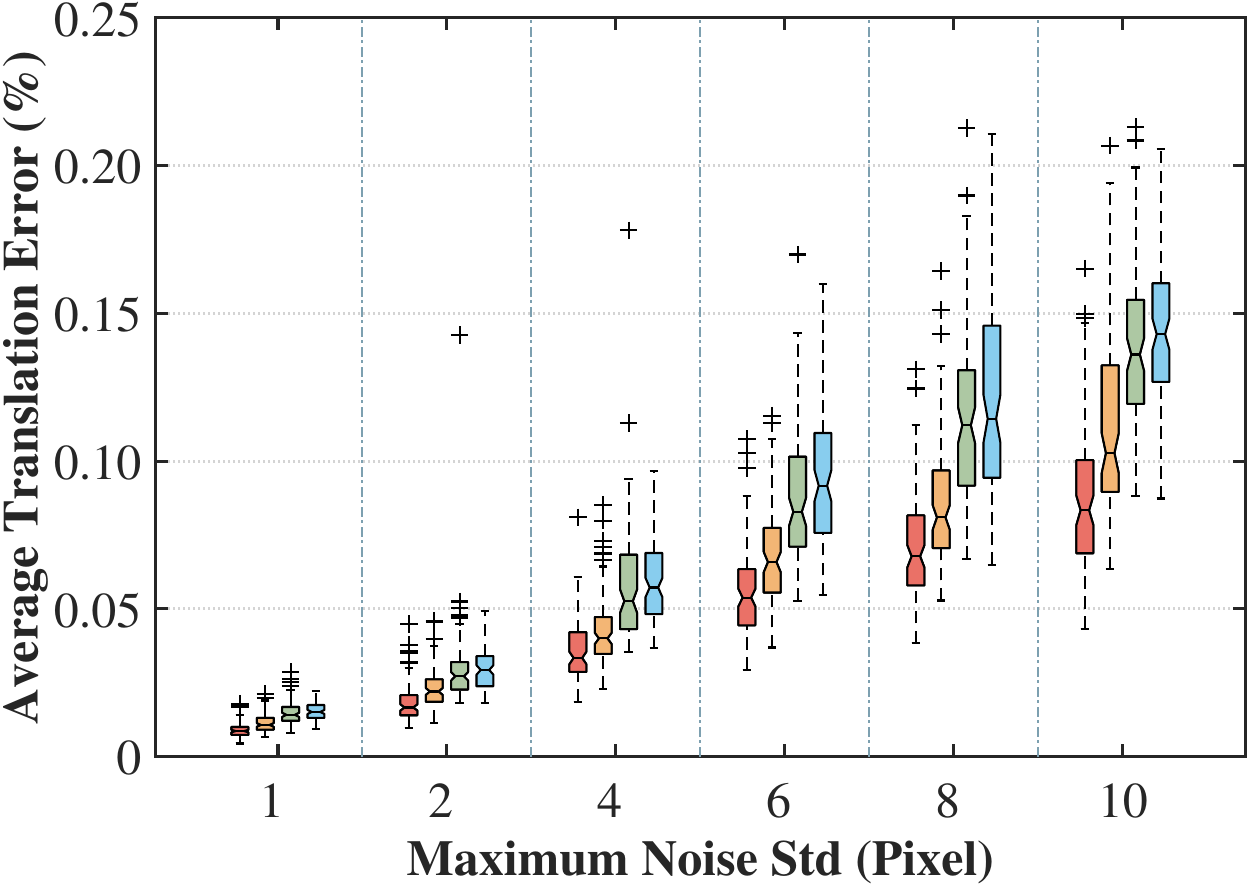} \qquad
		\includegraphics[width=0.3\linewidth]{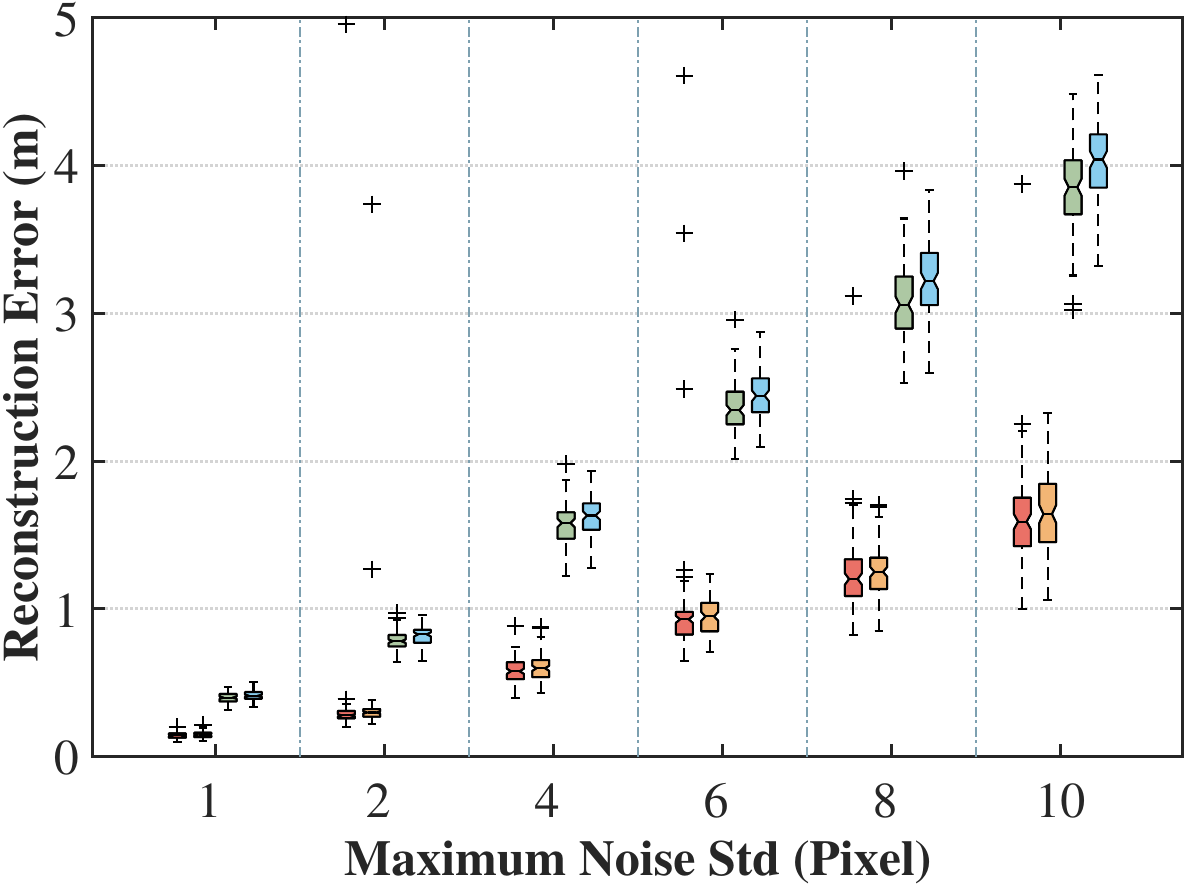} 
		\caption{Algorithm accuracy comparison in \textit{Forward-Curve}}
		\label{fig:Forward_Curve_Noise}
	\end{subfigure}
	\begin{subfigure}{0.85\linewidth}
		\centering
		\includegraphics[width=0.3\linewidth]{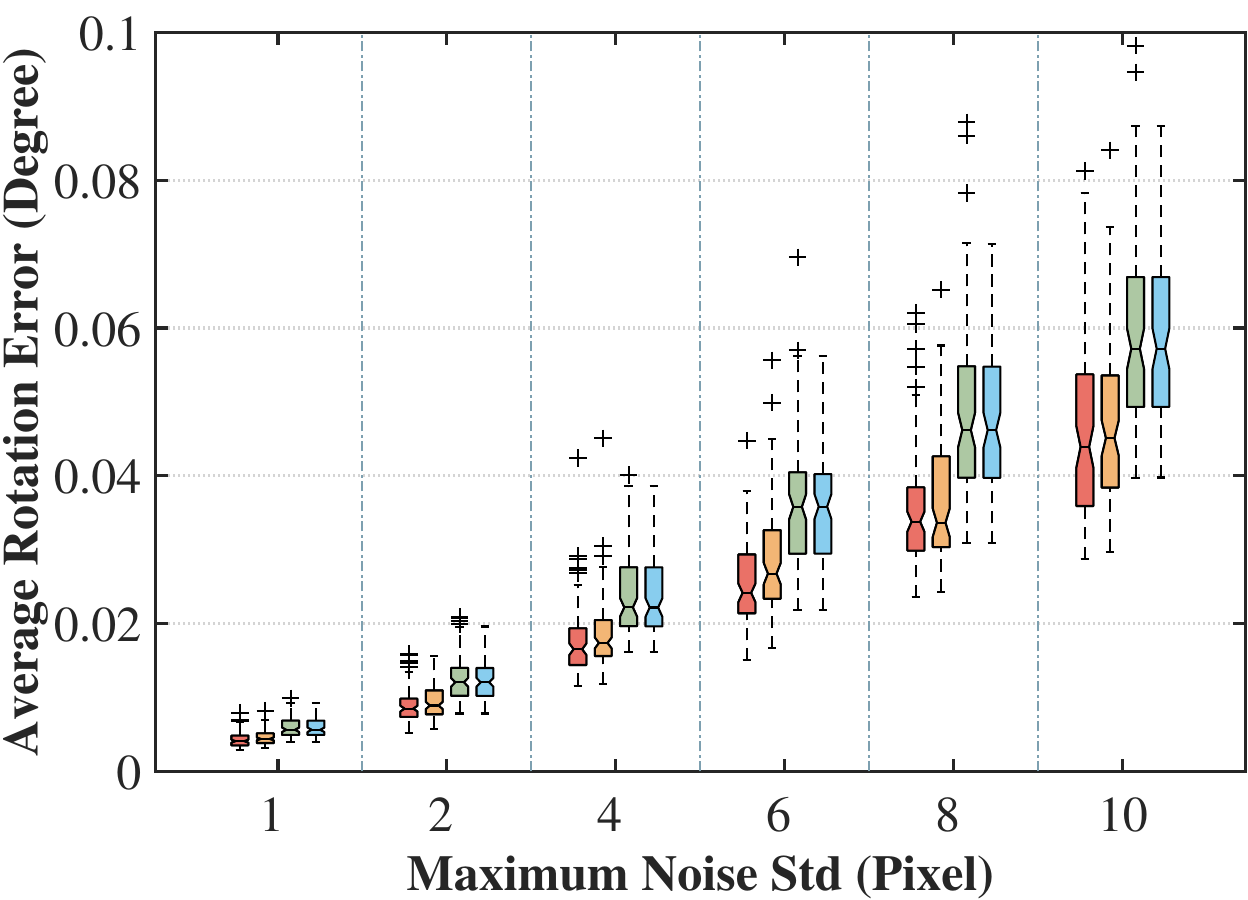} \qquad
		\includegraphics[width=0.3\linewidth]{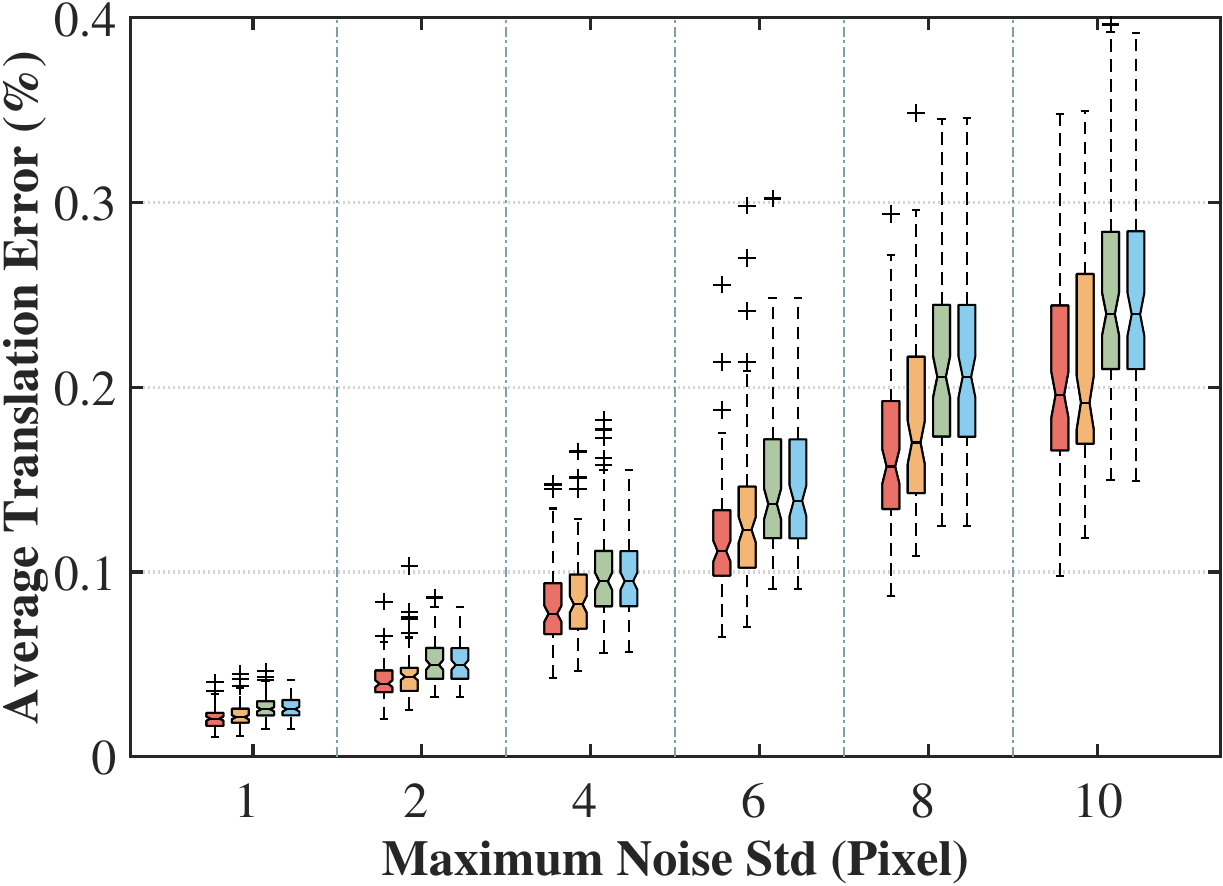} \qquad
		\includegraphics[width=0.3\linewidth]{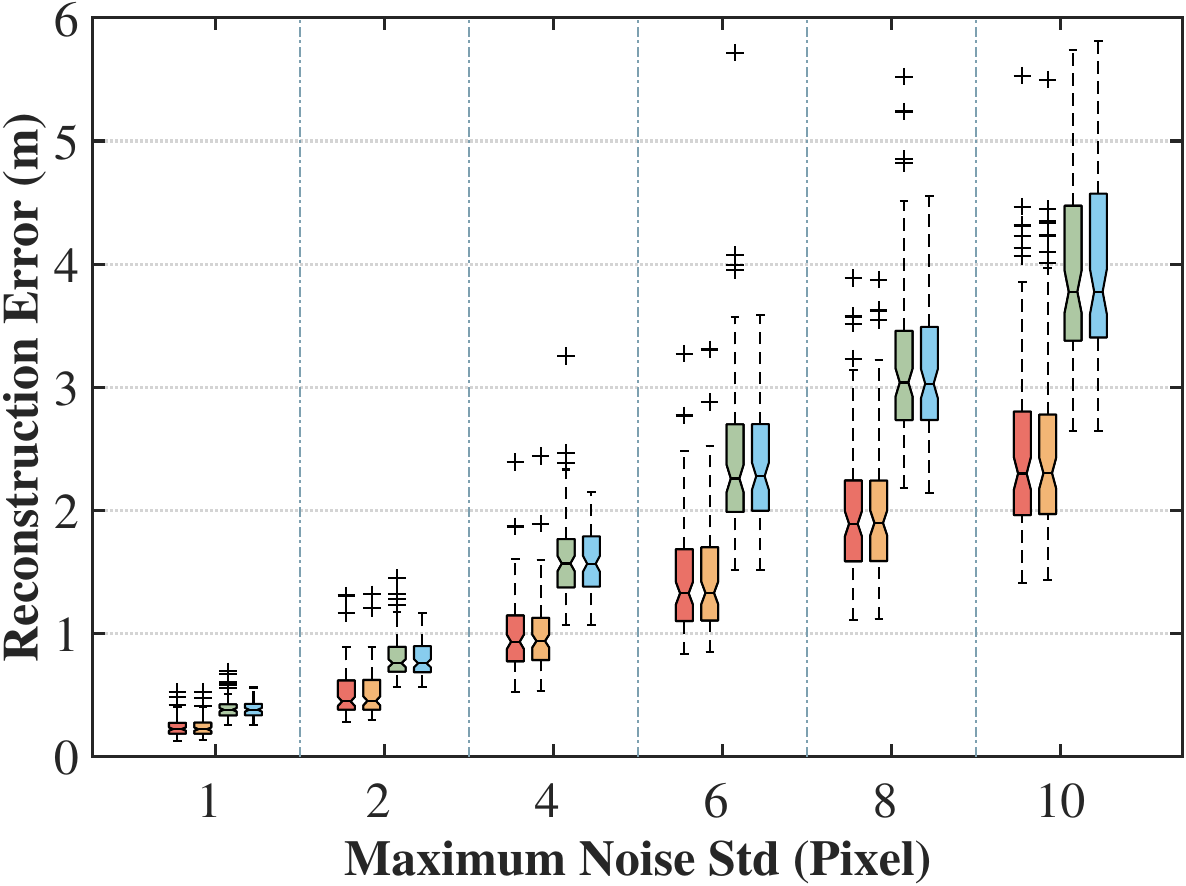} 
		\caption{Algorithm accuracy comparison in \textit{Omni-Linear}}
		\label{fig:Omni_Linear_Noise}
	\end{subfigure}
	\begin{subfigure}{0.85\linewidth}
		\centering
		\includegraphics[width=0.3\linewidth]{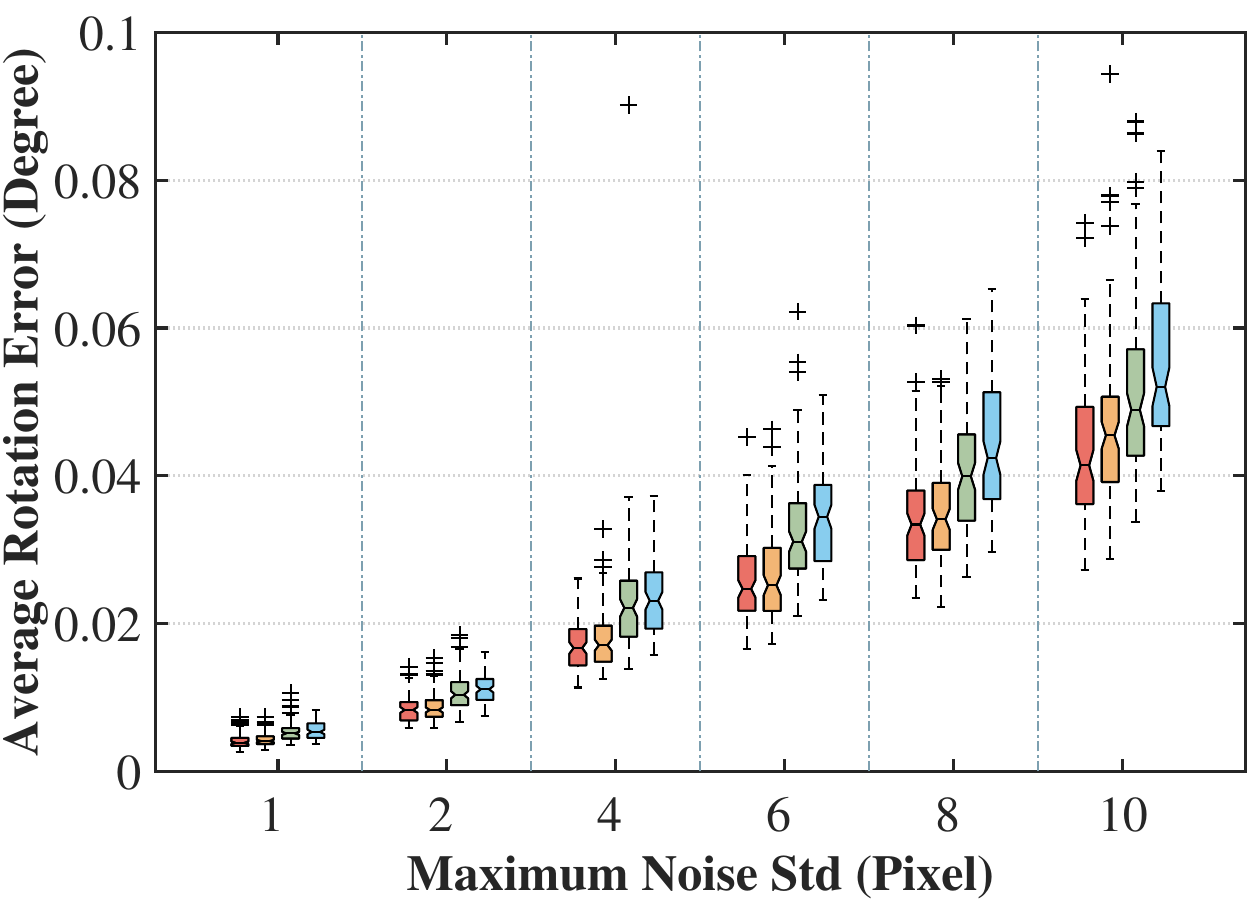} \qquad
		\includegraphics[width=0.3\linewidth]{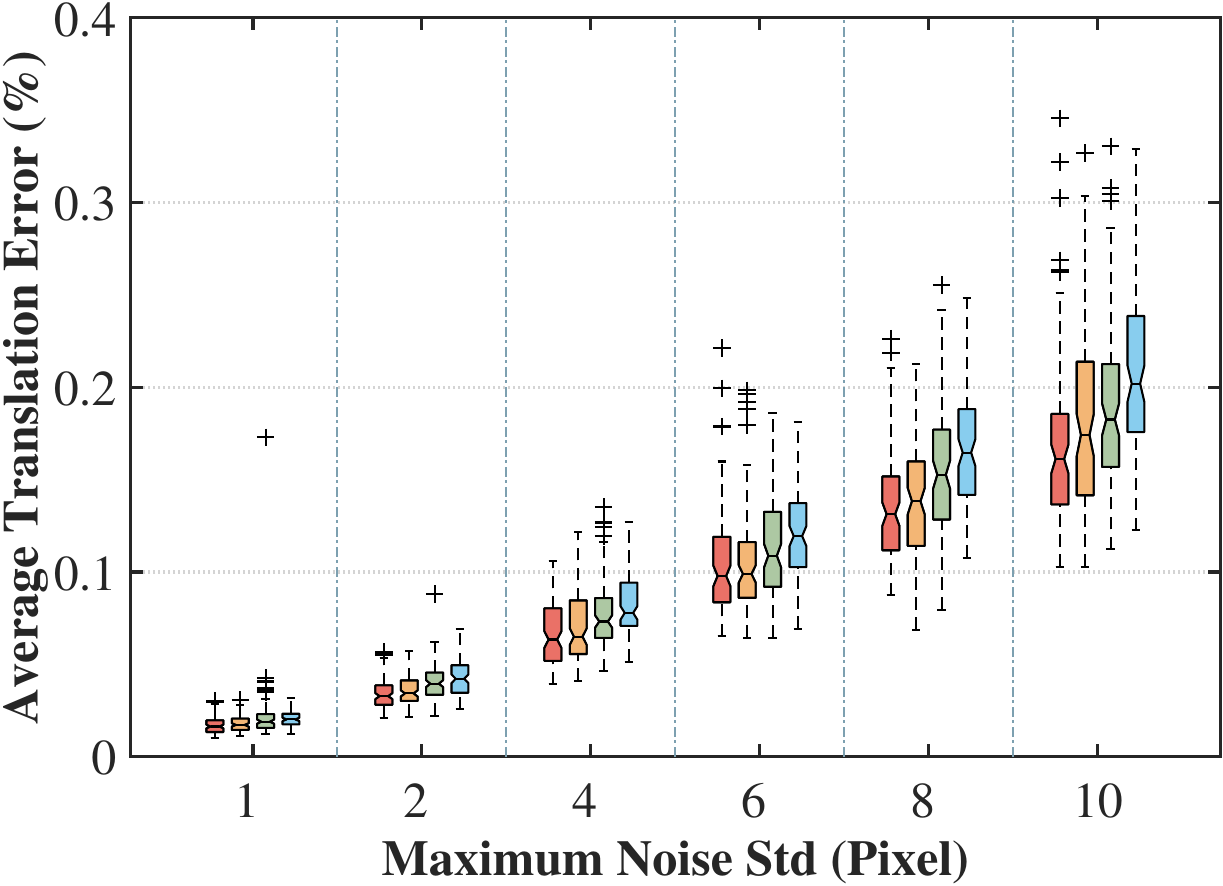} \qquad
		\includegraphics[width=0.3\linewidth]{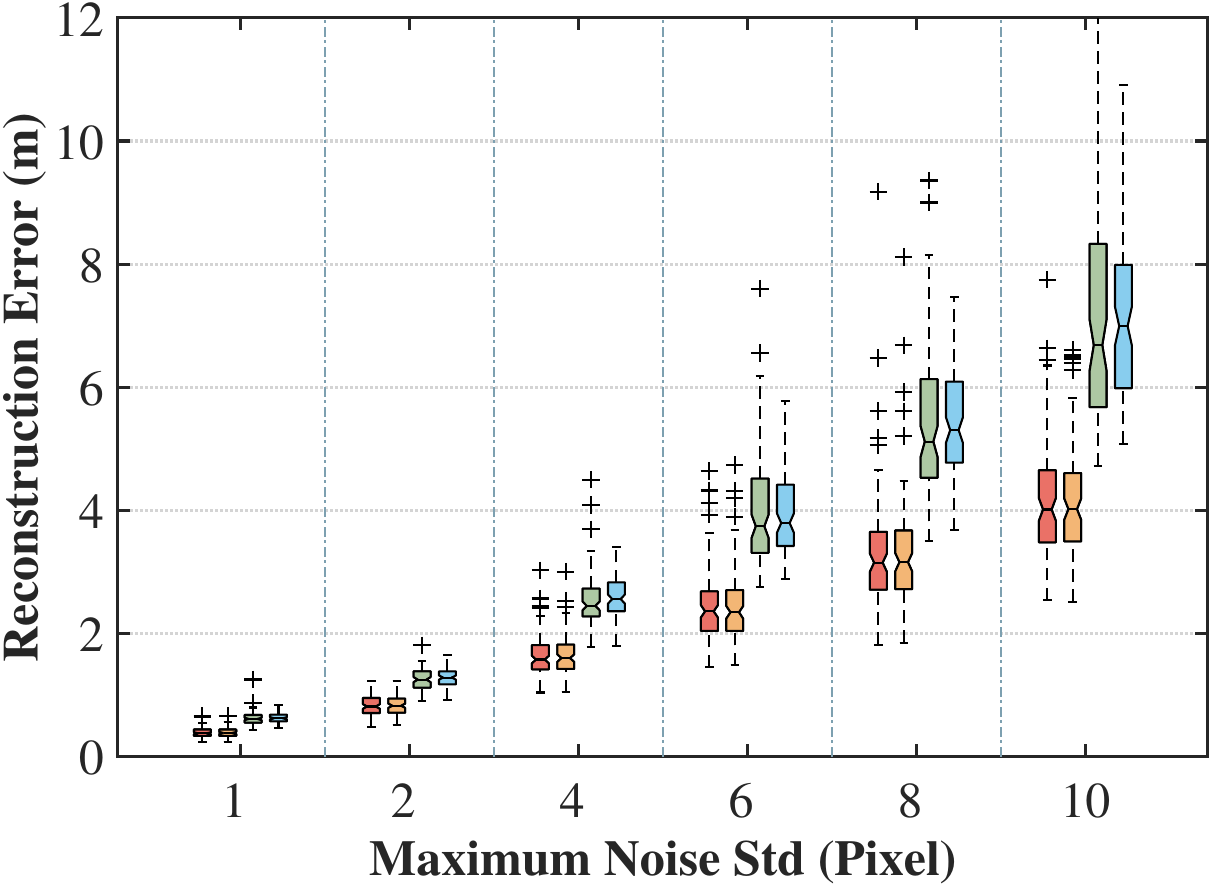}
		\caption{Algorithm accuracy comparison in \textit{Omni-Curve}}
		\label{fig:Omni_Curve_Noise}
	\end{subfigure}
	\caption{Accuracy comparison of four pose optimization algorithms on four synthetic datasets. The charts show the average rotation error (first column), average translation error (second column), and reconstruction error (third column) over 50 poses in the trajectory under different noise levels. Results indicate that the proposed algorithms \texttt{MCPALR} and \texttt{MCPA} consistently achieve lower errors than \texttt{MultiCol} and \texttt{BACS} across all noise levels.}
	\label{fig:Four_Conf_Noise}
\end{figure*}

The pose optimization results are shown in \cref{fig:Four_Conf_Noise}. The proposed methods, \texttt{MCPA} and \texttt{MCPALR}, consistently outperform the baseline algorithms \texttt{MultiCol} and \texttt{BACS} in pose estimation accuracy across different noise levels. We attribute the advantage to two key aspects. On the one hand, \texttt{MCPA} and \texttt{MCPALR} effectively decouple multi-camera poses from 3D scene points using the pose-only constraint, allowing them to focus solely on pose optimization. This renders our algorithms entirely immune to noise originating from the 3D scene points. On the other hand, the proposed algorithms construct multi-camera pose-only constraint using two base observations and their corresponding poses. Our base observation selection strategy optimally selects two observations to provide an analytical representation of the 3D points elegantly. As shown in \cref{fig:Four_Conf_Noise}, under low-noise levels $(\sigma_{max} < 2)$, the proposed algorithms achieve pose estimation accuracy comparable to traditional bundle adjustment methods \texttt{MultiCol} and \texttt{BACS}. However, as the noise level increases $(\sigma_{max} > 2)$, the proposed methods demonstrate significantly enhanced robustness. This trend can be inferred from the results \cref{fig:Base_Selection}. Compared to bundle adjustment methods, which rely on multiple observations triangulation, the proposed base observations selection strategy consistently maintains higher accuracy, even in high noise levels.

Among the two proposed methods, \texttt{MCPALR} slightly outperforms \texttt{MCPA} in pose optimization accuracy. This result is mainly attributed to the former adopting a more comprehensive set of constraints, as defined in \cref{eq:SXW} and \cref{eq:SXWr}, which effectively strengthens the constraint force during the pose optimization process. Note that in the \textit{Forward-Curve} dataset (see \cref{fig:Forward_Curve_Noise}), \texttt{MCPALR} exhibits a more noticeable improvement in accuracy. This is mainly due to the continuous observation of the 3D scene by the multi-camera system. Each scene point corresponds to more image observations, which allows the algorithm to select the two base observations more elegantly. As shown in the third column of \cref{fig:Four_Conf_Noise}, the lower reconstruction error confirms the advantage of our statistical optimal triangulation method proposed in this paper. 

\subsubsection{{Runtime Results}}
\begin{table*}[thb]
	\centering
	\LARGE
	\caption{Performance comparison of \texttt{MCPALR}, \texttt{MCPA} and \texttt{MultiCol} algorithms.}
	\resizebox{\textwidth}{!}{ 
		\begin{tabular}{cccccccccccc}
			\toprule
			\multirow{2}{*}{Poses} & \multirow{2}{*}{Points} & \multirow{2}{*}{Obs} & \multicolumn{3}{c}{{Runtime}(s)} & \multicolumn{3}{c}{Memory(MB)} & \multicolumn{3}{c}{$\bar{\varepsilon}_\mathbf{R}(^\circ)$, $\bar{\varepsilon}_\mathbf{t}(\%)$, $\bar{\varepsilon}_\mathbf{X}(m)$} \\
			{} & {} & {} & \texttt{MCPALR} & \texttt{MCPA} & \texttt{MultiCol} & \texttt{MCPALR} & \texttt{MCPA} & \texttt{MultiCol} & \texttt{MCPALR} & \texttt{MCPA} & \texttt{MultiCol} \\
			\midrule
			\multirow{3}{*}{50} & 1000 & 52,921 & 0.36 & \textbf{0.25} & 0.37 & \textbf{0.43} & \textbf{0.43} & 3.67 & \textbf{0.0128, 0.0217, 0.0820} & 0.0181, 0.0229, 0.1182 & 0.0231, 0.0380, 0.1725 \\
			{} & 3000 & 161,884 & 1.02 & \textbf{0.70} & 1.16 & \textbf{0.44} & \textbf{0.44} & 11.19 & \textbf{0.0093, 0.0146, 0.0573} & 0.0127, 0.0179, 0.0674 & 0.0147, 0.0242, 0.1263\\
			{} & 5000 & 270,165 & 1.51 & \textbf{1.12} & 1.86 & \textbf{0.45} & \textbf{0.45} & 18.66 & \textbf{0.0047, 0.0114, 0.0462} & 0.0061, 0.0110, 0.0599 & 0.0164, 0.0207, 0.1697\\
			\midrule
			\multirow{3}{*}{100} & 3000 & 167,861 & 1.26 & \textbf{0.89} & 1.38 & \textbf{1.01} & \textbf{1.01} & 11.70 & \textbf{0.0275, 0.0274, 0.2980} & 0.0350, 0.0301, 0.3931 & 0.0366, 0.0322, 0.4657\\
			{} & 5000 & 279,387 & 1.59 & \textbf{1.39} & 2.27 & \textbf{1.02} & \textbf{1.02} & 19.46 & \textbf{0.0187, 0.0157, 0.0224} & 0.0226, 0.0225, 0.2610 & 0.0351, 0.0315, 0.4065\\
			{} & 8000 & 446,521 & 2.99 & \textbf{2.08} & 4.39 & \textbf{1.03} & \textbf{1.03} & 31.06 & \textbf{0.0077, 0.0115, 0.0600} & 0.0134, 0.0155, 0.0725 & 0.0297, 0.0265, 0.3388\\
			\midrule
			\multirow{3}{*}{200} & 5000 & 669,417 & 4.41 & \textbf{3.29} & 7.27 & \textbf{4.85} & \textbf{4.85} & 46.27 & \textbf{0.0107, 0.0158, 0.2222} & 0.0129, 0.0142, 0.3685 & 0.0179, 0.0203, 0.4243\\
			{} & 8000 & 1,061,646 & 7.77 & \textbf{5.57} & 13.31 & \textbf{4.93} & \textbf{4.93} & 73.38 & \textbf{0.0161, 0.0135, 0.4205} & 0.0191, 0.0148, 0.5156 & 0.0238, 0.0190, 0.6346\\
			{} & 10000 & 1,376,647 & 9.84 & \textbf{7.11} & 17.94 & \textbf{4.99} & \textbf{4.99} & 95.13 & \textbf{0.0098, 0.0086, 0.2623} & 0.0122, 0.0104, 0.3019 & 0.0233, 0.0197, 0.6197\\
			\bottomrule
		\end{tabular}
	}
	\begin{tablenotes}
		\footnotesize
		\item \hspace{-3em} \textbf{Bold} values indicate the best results. 
	\end{tablenotes}
	\label{tab:Time_Memory}
\end{table*}
This experiment evaluates the computational efficiency of the algorithms across optimization problems of varying scales. We compare the performance of \texttt{MCPALR}, \texttt{MCPA}, and \texttt{MultiCol} algorithms with varying numbers of trajectory poses and 3D scene points. All three algorithms are implemented in the C++17 environment, while the open-source code for \texttt{BACS}, written in MATLAB, is not included in this experiment. Additionally, the experimental results in \cite{Urban2017multicol} have demonstrated that \texttt{MultiCol} is faster than \texttt{BACS} in terms of computation time. The iterative process employs the Levenberg-Marquardt (LM) algorithm, and the linear solver utilizes the Schur complement. All methods employ the $L_2$ norm error and the same termination criteria. The maximum iteration number is set to 10. All algorithms share the same initial pose obtained by adding perturbations to the ground truth. For each problem size, we conducted 100 independent experiments and reported the median results.

Following \cite{Ge2024pipo}, we report the problem size (including the number of poses, 3D points, and observations), the total time for optimization, the memory size of Hessian matrix, and the optimized errors in \cref{tab:Time_Memory}. The optimization time does not include the data reading and writing processes. Results show that our \texttt{MCPA} and \texttt{MCPALR} algorithms significantly outperform \texttt{MultiCol} in terms of runtime. With the same number of poses, the runtime of \texttt{MultiCol} increases substantially with more scene points, while \texttt{MCPA} and \texttt{MCPALR} are much more efficient, especially for large-scale data. On a dataset with 200 poses and 10,000 points, memory usage is reduced by 19 times and optimization speed is improved by 2.5 times. This advantage is mainly attributed to the introduction of multi-camera pose-only constraint, which reduces the number of optimization variables and reduces the complexity of the Hessian matrix.
{\texttt{MCPALR} slightly outperforms \texttt{MCPA} in accuracy, but incurs a modest increase in runtime due to its more complex cost function. Therefore, a trade-off in cost function design is necessary to balance speed and accuracy in practical applications}.

\subsection{Results on Real Dataset}
We conducted a comprehensive evaluation of the proposed algorithms on two publicly available multi-camera system datasets: ETH3D \cite{Thomas2017ETH3D} and KITTI Odometry Dataset \cite{Geiger2012KITTI}.

\subsubsection{ETH3D Dataset}
\begin{table*}[thb]
	\centering
	\captionsetup{width=.8\textwidth}
	\caption{Problem Sizes for Each Scene in ETH3D Dataset and Runtime for Optimization Algorithms.}
	\label{tab:eth_time}
	\resizebox{0.8\textwidth}{!}{
		\begin{tabular}{l l r r c c c c}
			\toprule
			&  &  & & \multicolumn{4}{c}{\textbf{{Runtime(s)}}} \\
			\cmidrule(lr){5-8}
			\textbf{Datasets}& \textbf{Images} & \textbf{Points} & \textbf{Observations} & \texttt{MCPALR} & \texttt{MCPA} & \texttt{MultiCol} & \texttt{RigBA}\\
			\midrule
			Delivery      & $4\times237$ & 21,318  & 1,012,916 & \underline{24.96}  & \textbf{14.00}  & 60.11  & 46.17\\
			Electro       & $4\times300$ & 29,668  & 1,018,049 & \underline{54.16}  & \textbf{20.66}  & 65.05  & 56.10\\
			Lakeside      & $4\times266$ & 28,426  & 1,396,698 & \underline{34.20}  & \textbf{24.01}  & 57.06  & 51.18\\
			Forest        & $4\times257$ & 53,039  & 2,266,880 & \underline{49.06}  & \textbf{37.23}  & 69.79  & 57.91\\
			Playground    & $4\times240$ & 40,322  & 1,266,495 & \underline{40.66}  & \textbf{25.21}  & 72.16  & 43.41\\
			Terrains      & $4\times165$ & 32,059  & 1,127,498 & \underline{17.94}  & \textbf{9.51}   & 56.85  & 26.38\\
			Storage Room  & $4\times199$ & 38,768  & 977,012   & {37.75}  & \textbf{25.15}  & 39.22  & \underline{26.41}\\
			Storage Room 2& $4\times208$ & 14,776  & 603,844   & \underline{19.25}  & \textbf{7.28}   & 25.32  & 30.23\\
			Sand Box      & $4\times278$ & 175,120 & 3,790,419 & \underline{97.58}  & \textbf{61.48}  & 158.99 & 111.48\\
			Tunnel        & $4\times352$ & 182,544 & 5,665,514 & \underline{281.73} & \textbf{187.54} & 341.31 & 1800.85\\
			\bottomrule
		\end{tabular}
	}
	\begin{tablenotes}
		\footnotesize
		\item \hspace{5em} \textbf{Bold} values indicate the shortest runtime, and \underline{underline} values indicate the second shortest.
	\end{tablenotes}
\end{table*}

\begin{table*}[thb]
	\centering
	\caption{Accuracy Comparison of Five Multi-Camera Pose Optimization Algorithms on the ETH3D Dataset.}
	\label{tab:eth_err}
	\small
	\setlength{\tabcolsep}{3pt} 
	\begin{tabular}{l *{15}{c}}
		\toprule
		& \multicolumn{3}{c}{\texttt{MCPALR}} & \multicolumn{3}{c}{\texttt{MCPA}} & \multicolumn{3}{c}{\texttt{MultiCol}} & \multicolumn{3}{c}{\texttt{BACS}} & \multicolumn{3}{c}{\texttt{RigBA}}\\
		\cmidrule(lr){2-4} \cmidrule(lr){5-7} \cmidrule(lr){8-10} \cmidrule(lr){11-13} \cmidrule(lr){14-16}
		\textbf{Datasets} & $\bar{\varepsilon}_{\mathbf{R}}(^\circ)$ & $\bar{\varepsilon}_{\mathbf{t}}(\%)$ & $\bar{\varepsilon}_{\mathbf{p}}(px)$ & 
		$\bar{\varepsilon}_{\mathbf{R}}(^\circ)$ & $\bar{\varepsilon}_{\mathbf{t}}(\%)$ & $\bar{\varepsilon}_{\mathbf{p}}(px)$ & 
		$\bar{\varepsilon}_{\mathbf{R}}(^\circ)$ & $\bar{\varepsilon}_{\mathbf{t}}(\%)$ & $\bar{\varepsilon}_{\mathbf{p}}(px)$ & 
		$\bar{\varepsilon}_{\mathbf{R}}(^\circ)$ & $\bar{\varepsilon}_{\mathbf{t}}(\%)$ & $\bar{\varepsilon}_{\mathbf{p}}(px)$ & 
		$\bar{\varepsilon}_{\mathbf{R}}(^\circ)$ & $\bar{\varepsilon}_{\mathbf{t}}(\%)$ & $\bar{\varepsilon}_{\mathbf{p}}(px)$ \\
		\midrule
		Delivery & \underline{0.062} & \textbf{0.397} & \underline{0.677} & \textbf{0.055} & \underline{0.454} & \underline{0.677} & 0.181 & 1.083 & 0.756 & 0.103 & 0.635 & 0.687 & 0.190 & 2.471 & \textbf{0.432}\\
		Electro & \underline{0.089} & \underline{0.510} & {0.612} & \textbf{0.065} & \textbf{0.477} & 0.616 & 0.643 & 4.029 & 0.951 & 0.201 & 0.851 & \textbf{0.417} & 0.422 & 2.897 & \underline{0.545}\\
		Lakeside & \textbf{0.012} & \underline{0.062} & \textbf{0.406} & \underline{0.013} & \textbf{0.058} & \textbf{0.406} & 0.029 & 0.103 & \underline{0.410} & 0.029 & 0.106 & \underline{0.410} & 0.450 & 0.068 & 0.978\\
		Forest  & \textbf{0.037} & \textbf{0.101} & \textbf{0.423} & \underline{0.042} & \underline{0.117} & \underline{0.423} & 0.085 & 0.238 & 0.429 & 0.071 & 0.197 & 0.429 & 0.141 & 1.605 & 0.565\\
		Playground & \textbf{0.094} & \underline{0.596} & \textbf{0.364}  & \underline{0.098} & \textbf{0.570} & 0.368 & 0.167 & 0.901 & \underline{0.365} & 0.910 & 7.367 & 0.387 & 0.290 & 4.584 & 0.504\\
		Terrains & \textbf{0.171} & \underline{0.300} & \underline{0.649} & \underline{0.184} & \textbf{0.273} & 0.660 & 0.252 & 0.616 & 0.730 & 0.318 & 0.522 & \underline{0.654} & 0.977 & 2.237 & \textbf{0.310}\\
		Storage Room & \textbf{0.032} & \textbf{0.333} & \textbf{0.456} & \underline{0.048} & \underline{0.424} & 0.496 & 0.061 & 0.576 & \underline{0.461} & 0.119 & 0.841 & 0.500 & 0.200 & 3.164 & 0.556\\
		Storage Room 2& \textbf{0.051} &\textbf{0.785} & \underline{0.508} & \underline{0.056} & \underline{0.836} & 0.510 & 0.165 & 3.272 & 0.593 & 0.081 & 1.216 & \textbf{0.504} & 0.154 & 2.884 & 0.514\\
		Sand Box & \underline{0.030} & \underline{0.113} & \textbf{0.342} & \textbf{0.026} & \textbf{0.111} & 0.349 & 0.048 & 0.196 & \underline{0.348} & 0.051 & 0.215 & 0.349 & 0.500 & 4.515 & 0.644\\
		Tunnel & \underline{0.051} & \textbf{0.246} & {0.349} & \textbf{0.046} & 0.260 & \underline{0.340} & 0.202 & 0.938 & 0.399 & 0.073 & 0.458 & 0.352 & 0.076 & 0.397 & \textbf{0.329}\\
		\bottomrule
	\end{tabular}
	\begin{tablenotes}
		\footnotesize
		\item \hspace{-1.5em} \textbf{Bold} values indicate the smallest error, and \underline{underline} values indicate the second smallest.
	\end{tablenotes}
\end{table*}
\begin{figure*}[t]
	\centering
	\includegraphics[width=0.8\linewidth]{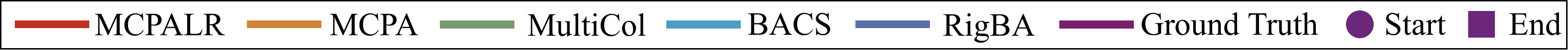}\\
	\vspace{0.5em}
	\subfloat[Electro\label{fig:Electro_traj}]{
		\centering
		\includegraphics[width=0.24\linewidth]{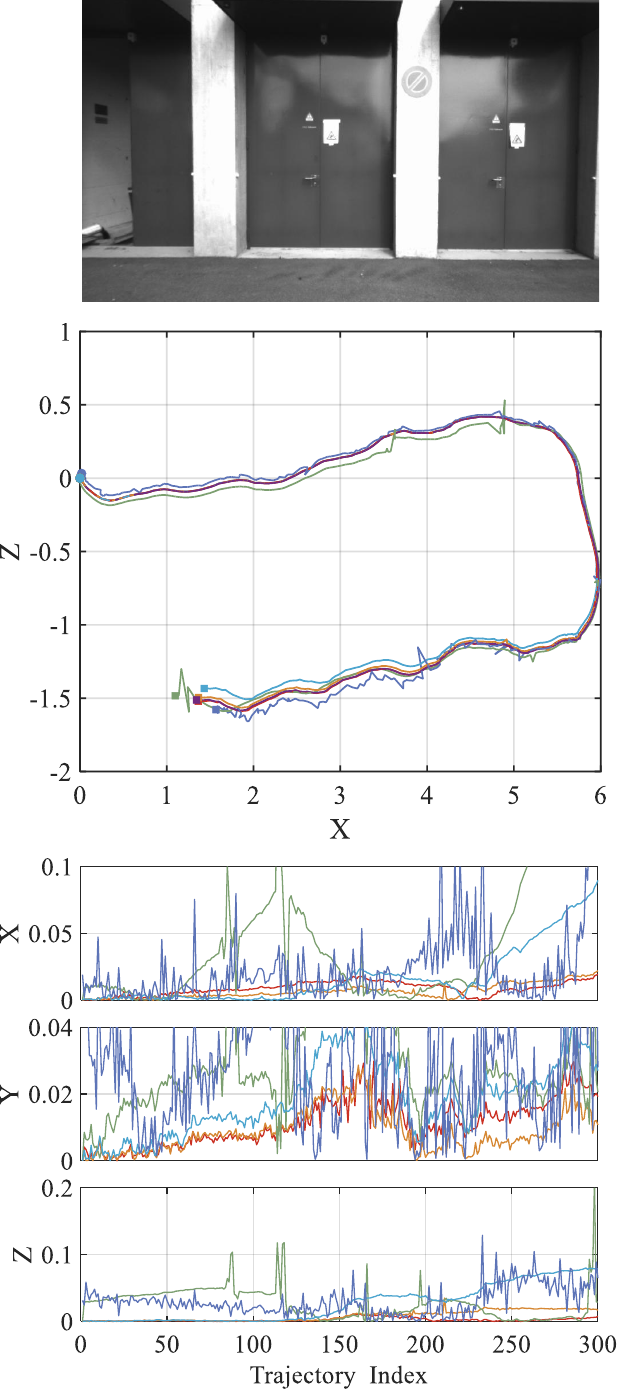}%
	}
	\subfloat[Forest\label{fig:Forest_traj}]{
		\centering
		\includegraphics[width=0.24\linewidth]{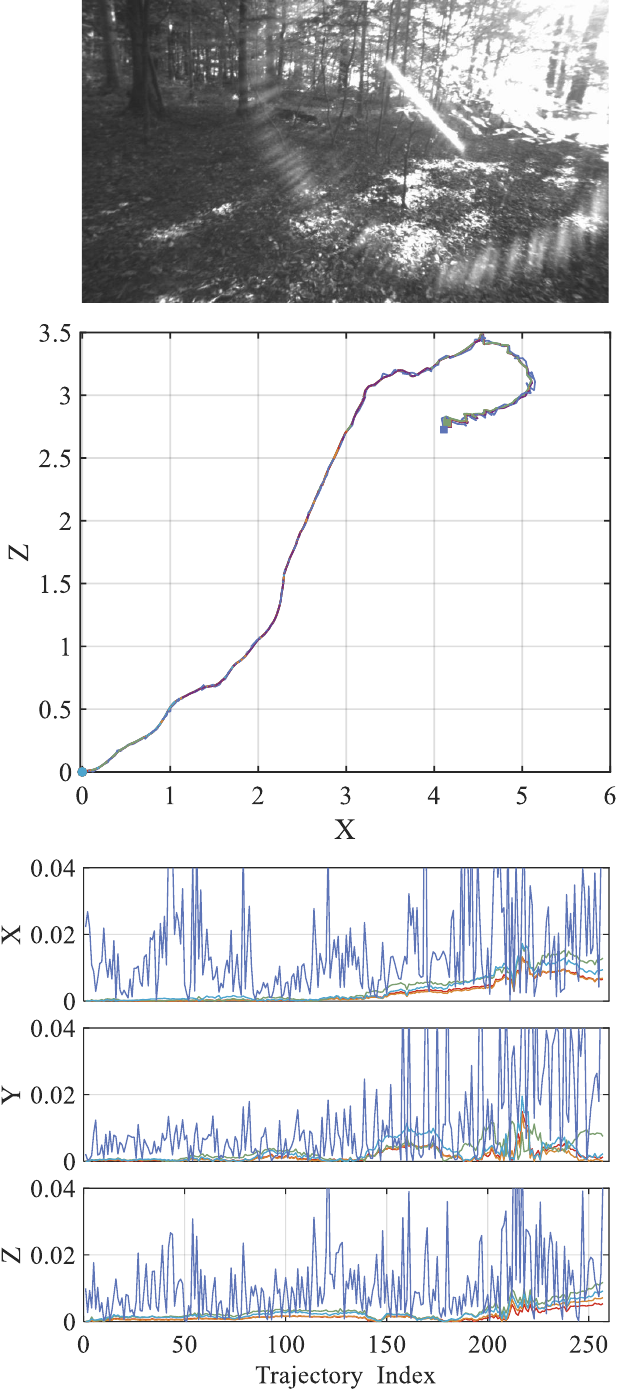}%
	}
	\subfloat[Terrains\label{fig:Terrains_traj}]{
		\centering
		\includegraphics[width=0.24\linewidth]{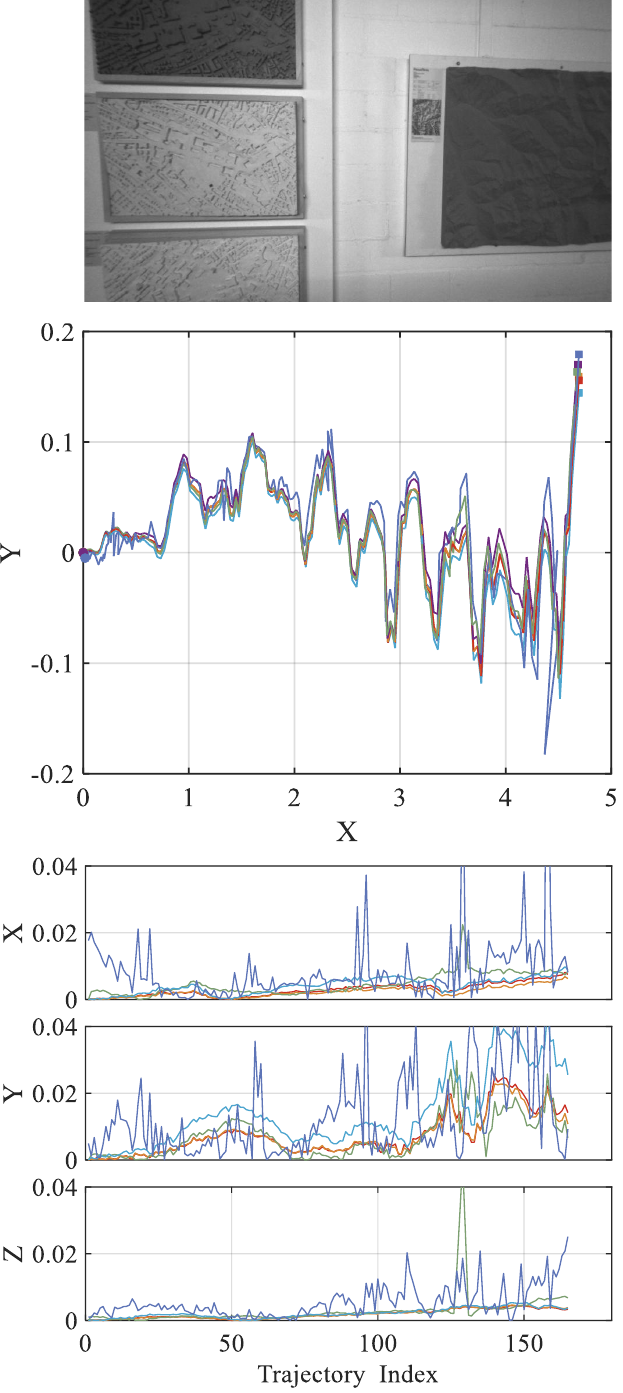}%
	}
	\subfloat[Playground\label{fig:Playground_traj}]{
		\centering
		\includegraphics[width=0.24\linewidth]{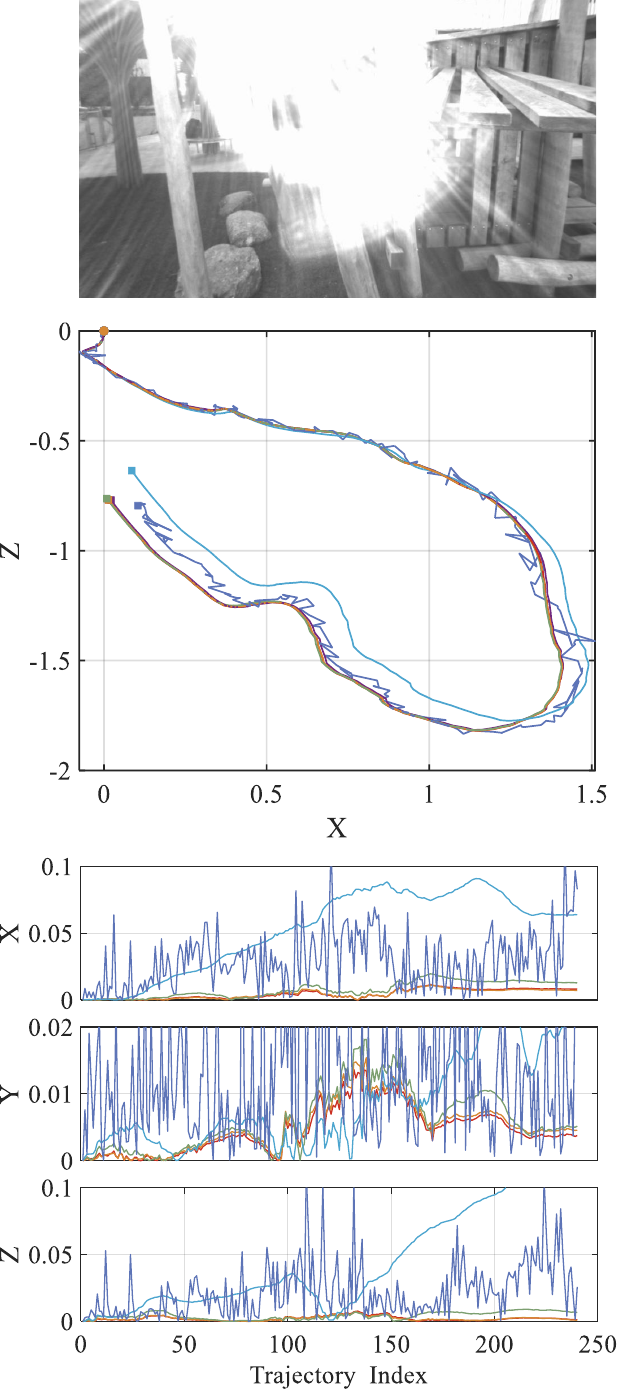}%
	}
	\caption{{Comparison of optimized trajectories for representative scenes Electro, Forest, Terrains, and Playground from the ETH3D dataset. \textbf{Top row} displays example images of the four scenes, which encompass low-texture, outdoor, indoor, and extreme exposure conditions. \textbf{Middle row} shows the comparison of the optimized trajectories of four algorithms against the ground truth. \textbf{Bottom row} illustrates the performance of the absolute trajectory error in the XYZ components.}}
	\label{fig:ETH3D_traj}
\end{figure*}

The ETH3D dataset \cite{Thomas2017ETH3D} is a widely recognized 3D reconstruction dataset consisting of ten common scenes that cover challenging environments such as indoor, outdoor, complex lighting, and low-texture areas, with a total of over 10,000 images. The multi-camera system in the dataset consists of four cameras, all observing the same direction, with a layout similar to the forward configuration shown in \cref{fig:Forward_Omni}. The dataset provides camera intrinsics but not extrinsics. We select the leftmost camera as the body coordinate frame and compute each camera’s extrinsics as its average relative pose to this frame. Since the ground truth lacks absolute scale, the extrinsics are scale-free; however, pose optimization accuracy can still be evaluated under a consistent relative scale. To provide a comprehensive evaluation against modern state-of-the-art pipeline, we also incorporate the \texttt{rig\_bundle\_adjuster} (denoted as \texttt{RigBA}) from COLMAP \cite{Johannes2016colmap} as a key baseline.

The feature extraction and matching of the images are obtained from COLMAP, following the default settings. The feature covariance can be calculated using existing methods \cite{Zeisl2009estimation,Muhle2023learning}. To facilitate the verification of the proposed algorithm, we simplify the covariance computation by back-propagating the reprojection error \cite{Hartley2003multiple}. The initial poses input for the optimization algorithm, similar to \cite{Ye2022coli}, is obtained by adding Gaussian noise to the ground truth, resulting in an initial average reprojection error of approximately 10 pixels. For \texttt{RigBA}, we initialize its solver with the same perturbed poses and fixed extrinsic parameters to ensure a fair comparison. All five algorithms share the same termination criteria: either reaching a maximum number of 30 iterations or the cost change falling below the threshold.

\Cref{tab:eth_time} reports the problem sizes and runtime for 10 scenes in the ETH3D dataset. The results indicate that the proposed two algorithms consistently outperform the \texttt{MultiCol} and \texttt{RigBA} in terms of computation time. Specifically, \texttt{MCPA} exhibits exceptional efficiency, achieving a maximum of 5.98 times (Terrains) faster runtime and an average of 2.29 times faster compared to \texttt{MultiCol}. Although \texttt{MCPALR} has a slightly lower running speed than \texttt{MCPA}, it still achieves a maximum runtime reduction of 3.17 times (Terrains) and an average reduction of 1.44 times compared to \texttt{MultiCol}. This speedup primarily stems from the reduced number of optimization parameters and the simplified complexity of solving the normal equations. The optimization accuracy comparison results are presented in \cref{tab:eth_err}. Results demonstrate that the proposed algorithms significantly outperform three baseline methods in terms of accuracy. Among them, \texttt{MCPALR} achieves the best accuracy in most scenarios, followed by \texttt{MCPA}. In some scenes, \texttt{MultiCol}, \texttt{BACS} and \texttt{RigBA} may exhibit the smallest or second smallest reprojection errors even when the pose accuracy is poor. This is mainly attributed to the fact that these algorithms primarily focus on minimizing the reprojection error. In contrast, the cost function we propose emphasizes pose optimization.

We selected four representative scenes from the ETH3D dataset: Electro, Forest, Terrains, and Playground, which represent low-texture, outdoor, indoor, and extreme exposure conditions, respectively. \Cref{fig:ETH3D_traj} presents example images of these scenes (top row), comparison of the optimized multi-camera system trajectories (middle row), and the XYZ components of the absolute trajectory error (bottom row). The trajectory comparison results validate the advantages of the proposed \texttt{MCPALR} and \texttt{MCPA} algorithms, attributed to the focus on pose-specific optimization. Additionally, the combination of multi-camera pose-only constraint and base-observations selection strategy implicitly increases the weight of base observations while effectively reducing the impact of outliers. In contrast, bundle adjustment methods, \texttt{MulticCol}, \texttt{BACS} and \texttt{RigBA}, treat all points as having equal contributions to accuracy and heavily rely on the initial values of the 3D points. \texttt{RigBA} shows significant trajectory oscillations. We attribute this behavior to the insufficient exploitation of extrinsic constraints in the optimization, making fail to local optimum.

\subsubsection{KITTI Odometry Dataset}
\begin{table}[thb]
	\centering
	\setlength{\tabcolsep}{3pt} 
	\caption{Problem Sizes for Each Sequence in KITTI Odometry Dataset and Comparison of Runtime for Optimization Algorithms.}
	\label{tab:kitti_time}
		\begin{tabular}{c l r r c c c}
			\toprule
			& & & & \multicolumn{3}{c}{\textbf{{Runtime(s)}}} \\
			\cmidrule(lr){5-7}
			\textbf{Seq.}& \textbf{Images} & \textbf{Points} & \textbf{Obs.} & \texttt{MCPALR} & \texttt{MCPA} & \texttt{MultiCol} \\
			\midrule
			00        &$2\times4541$ & 1,168,922 & 14,665,565 & \underline{511.66} & \textbf{235.98} & 664.30 \\
			01        & $2\times1101$ & 233,978  & 2,358,701 & \underline{125.15}  & \textbf{95.49}  & 184.75 \\
			02        & $2\times4661$ & 1,483,015 & 16,358,049 & 767.70 & \textbf{536.07} & \underline{729.16} \\
			03        & $2\times801$ & 389,703  & 4,423,236 & \underline{186.28}  & \textbf{129.83}  & 226.75  \\
			04        & $2\times271$ & 108,799  & 1,074,786 & \underline{28.78} & \textbf{15.30}  & 48.55  \\
			05 & $2\times2761$ & 1,146,276  & 11,794,689 & \underline{441.48} & \textbf{269.01} & 613.56  \\
			06 & $2\times1101$ & 449,410  & 3,943,665 & \underline{185.10} & \textbf{106.42} & 192.81  \\
			07 & $2\times1101$ & 432,890  & 4,772,387   & \underline{102.15}  & \textbf{61.44}   & 230.30  \\
			08 & $2\times4071$ & 1,113,508 & 13,602,975 & \underline{273.92} & \textbf{171.14} & 654.01  \\
			09 & $2\times1591$ & 708,950 & 5,975,625 & 250.51  & \textbf{145.67}  & \underline{239.97} \\
			10 & $2\times1201$ & 549,428 & 4,823,916 & 219.44  & \textbf{145.39}  & \underline{213.82} \\
			\bottomrule
		\end{tabular}
	\begin{tablenotes}
		\footnotesize
		\item  \textbf{Bold} values indicate the shortest runtime, and \underline{underline} values indicate the second shortest.
	\end{tablenotes}
\end{table}

\begin{figure*}[tbp]
	\centering
	\subfloat{
		\centering
		\includegraphics[width=0.33\linewidth]{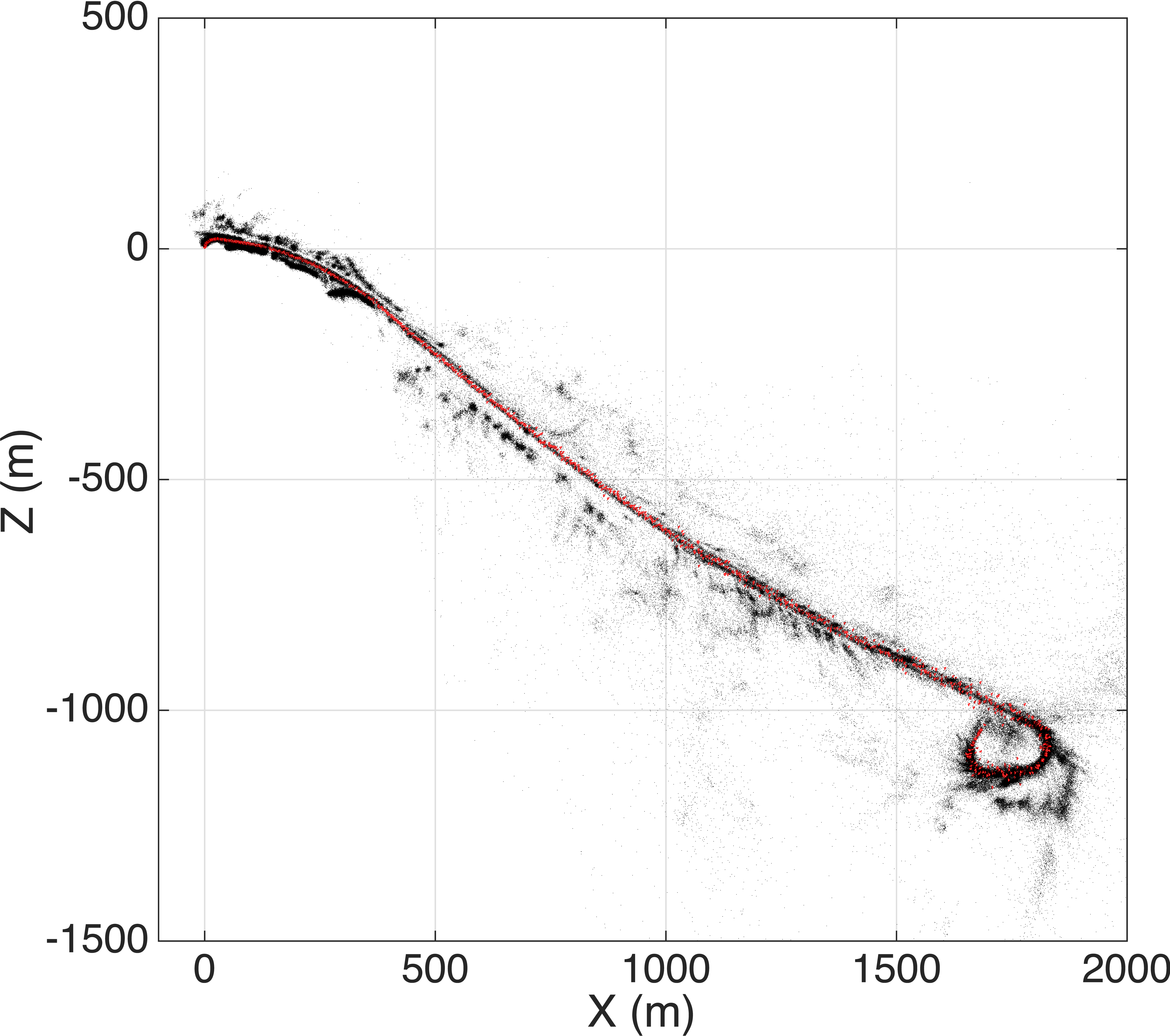} 
	}
	\subfloat{
		\centering
		\includegraphics[width=0.33\linewidth]{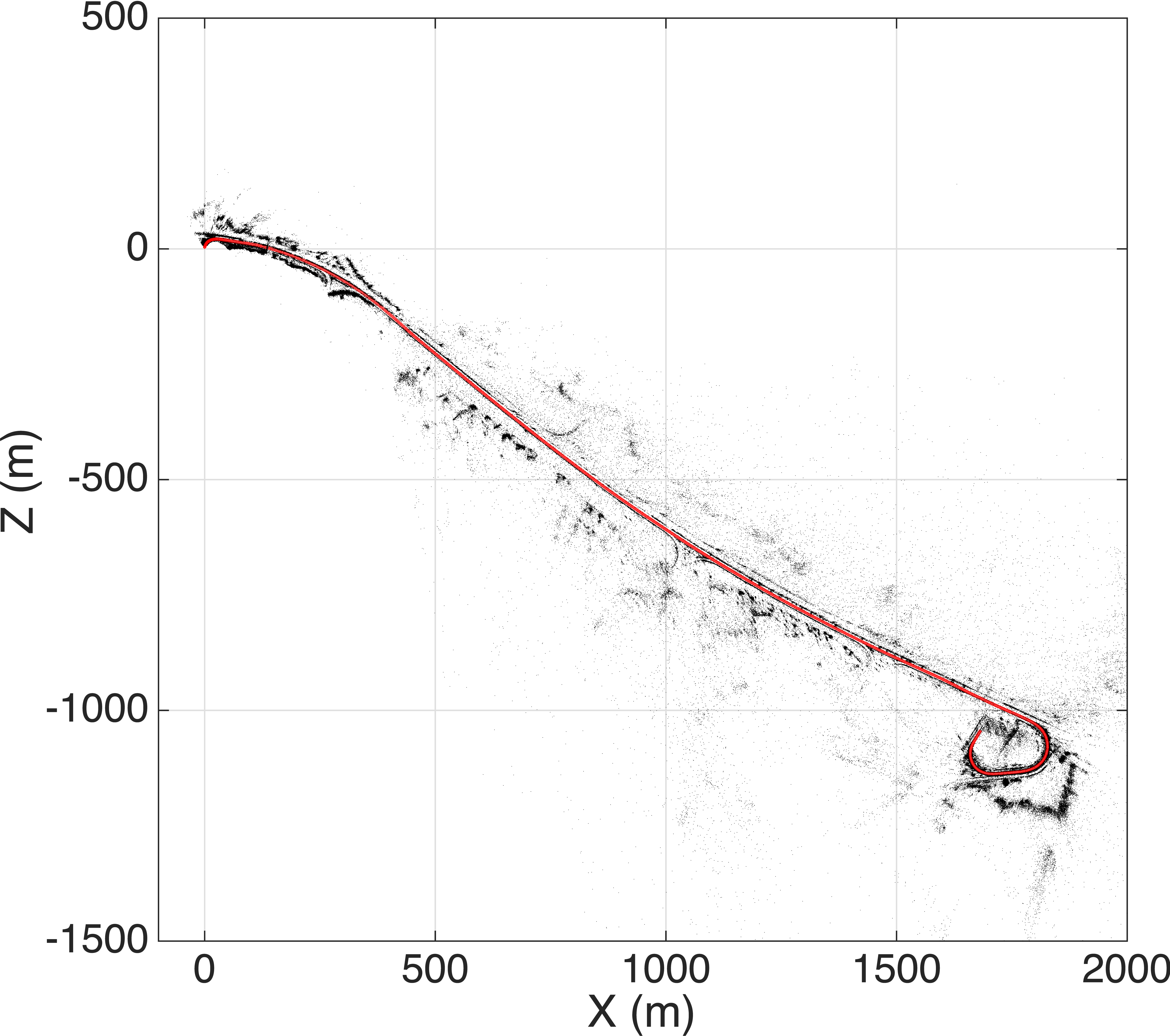} 
	}
	\subfloat{
		\centering
		\includegraphics[width=0.33\linewidth]{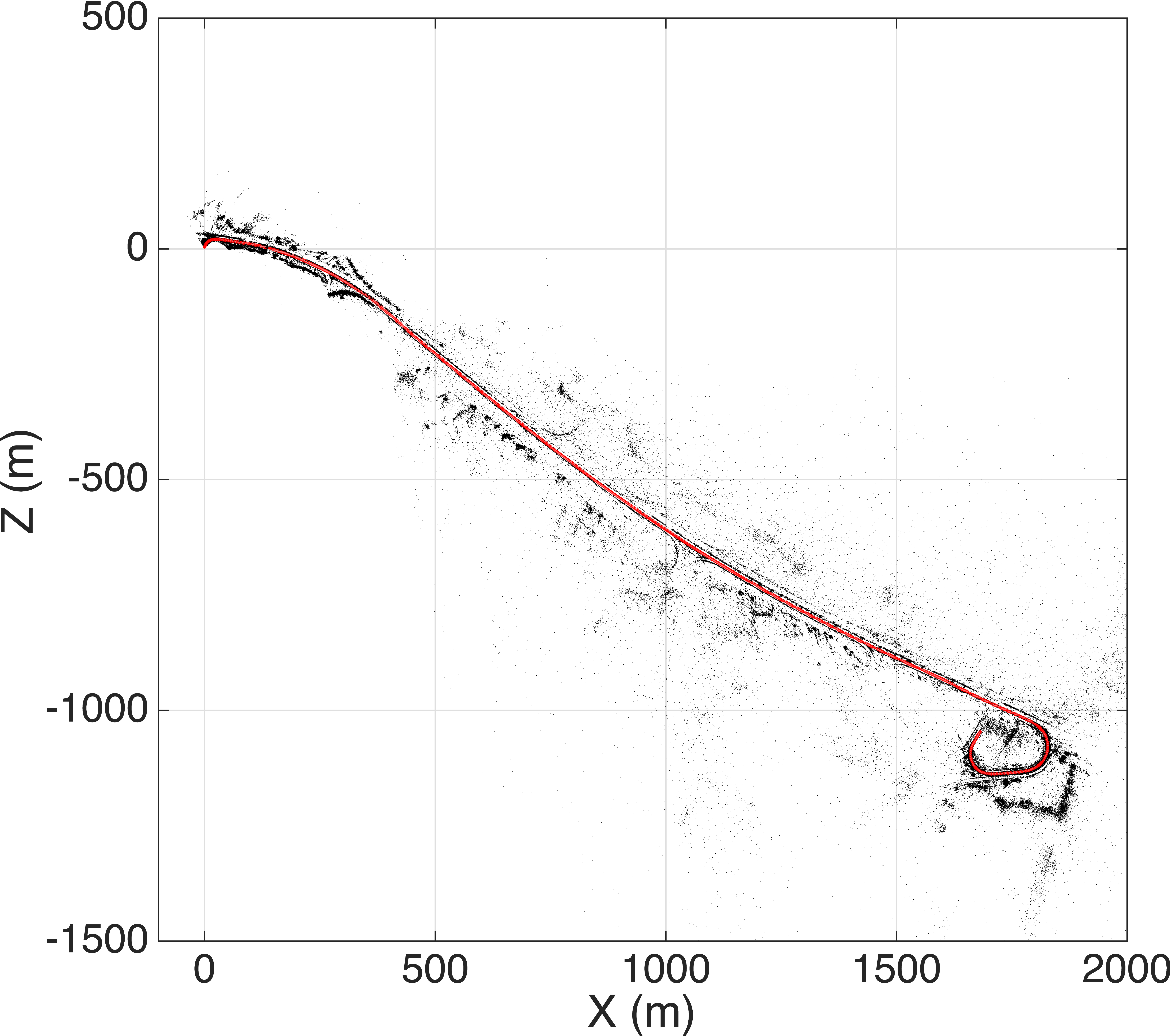} 
	}\\
	\setcounter{subfigure}{0}
	\subfloat[Initial\label{fig:Init_07_recon}]{
		\centering
		\includegraphics[width=0.33\linewidth]{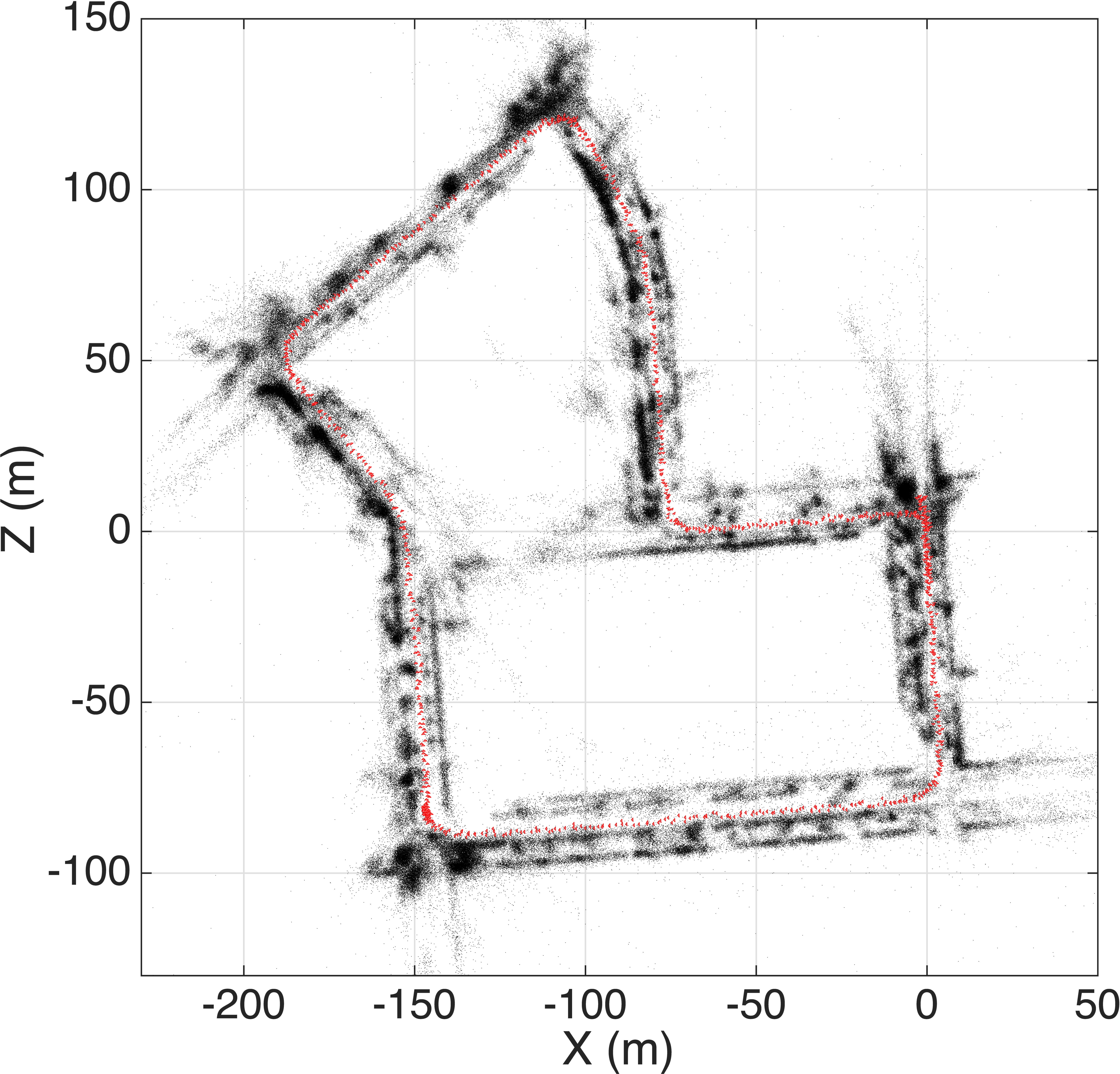} 
	}
	\subfloat[MCPALR\label{fig:MCPALR_07_recon}]{
		\centering
		\includegraphics[width=0.33\linewidth]{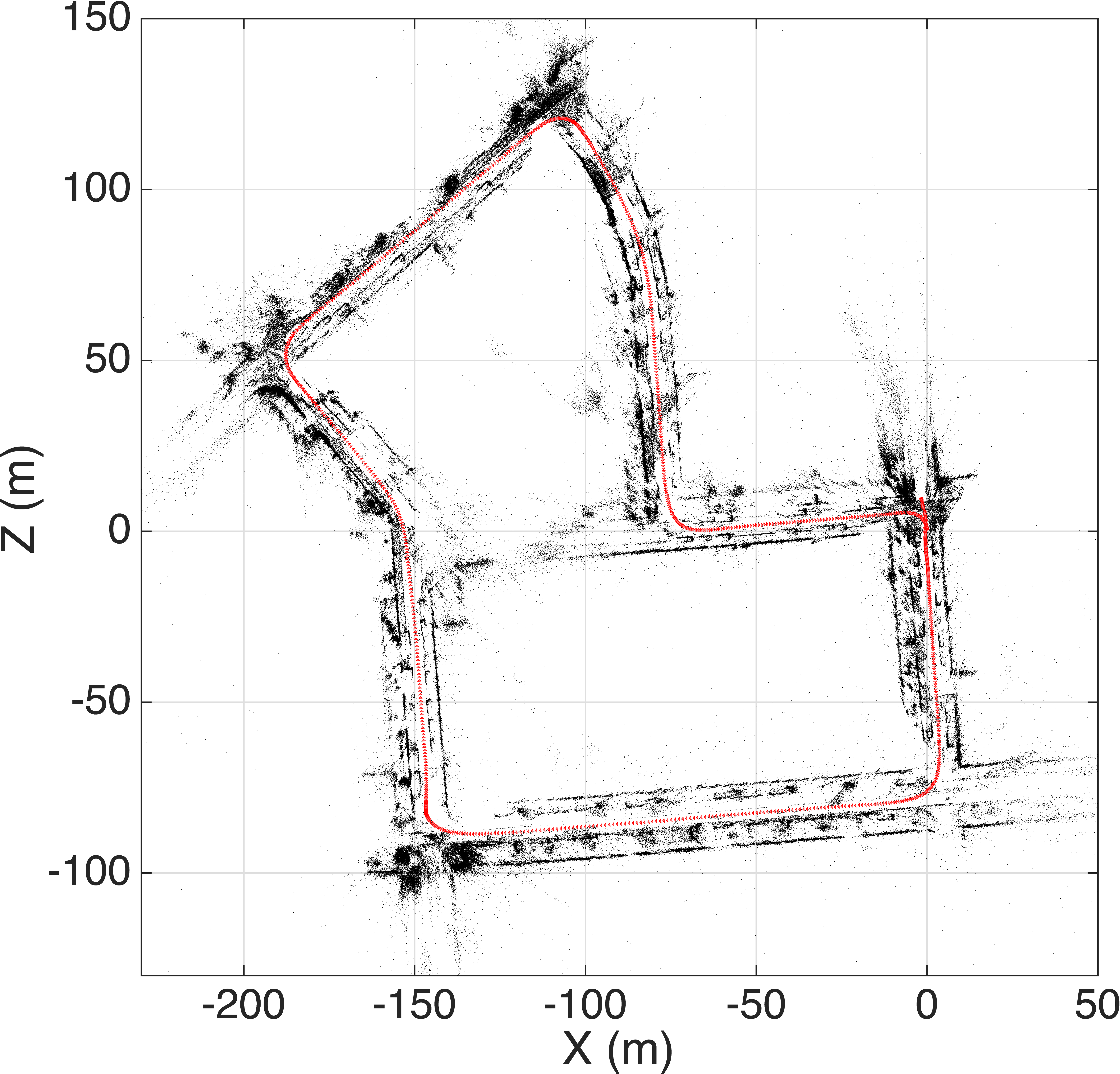} 
	}
	\subfloat[MCPA\label{fig:MCPA_07_recon}]{
		\centering
		\includegraphics[width=0.33\linewidth]{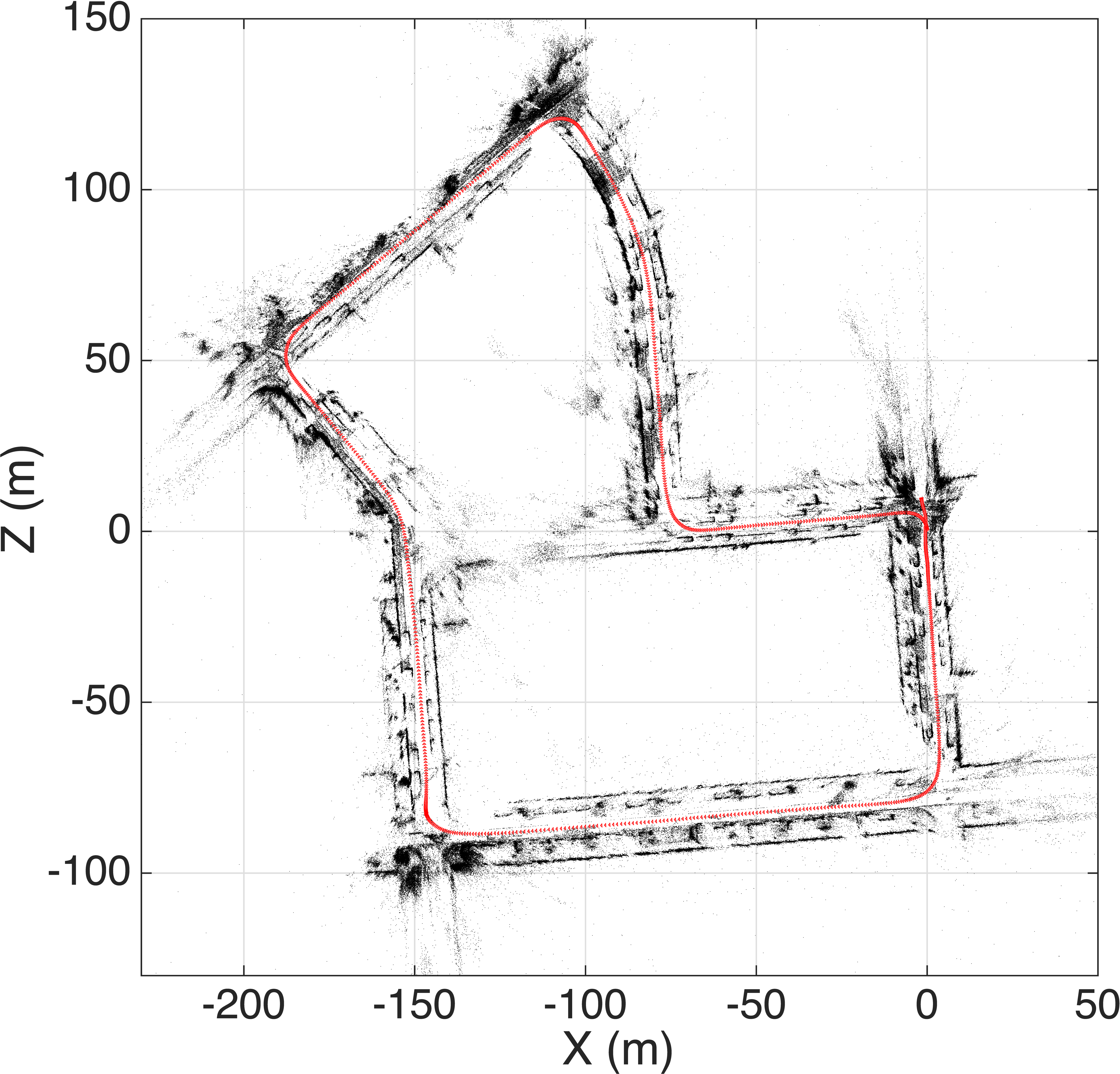}
	}
	\caption{Trajectory optimization and scene reconstruction results for Seq.01 (top row) and Seq.07 (bottom row) of the KITTI dataset. The first column shows the initial trajectories and point cloud structures, which exhibit noticeable blurriness and noise. The second and third columns display the trajectories and reconstructed point clouds after optimization with \texttt{MCPALR} and \texttt{MCPA}, respectively, demonstrating significant improvements in clarity and completeness.}
	\label{fig:kitti_010507_recon}
\end{figure*}
\begin{figure*}[tbp]
	\centering
	\subfloat[MCPALR\label{fig:MCPALR_traj}]{
		\centering
		\includegraphics[width=0.23\linewidth]{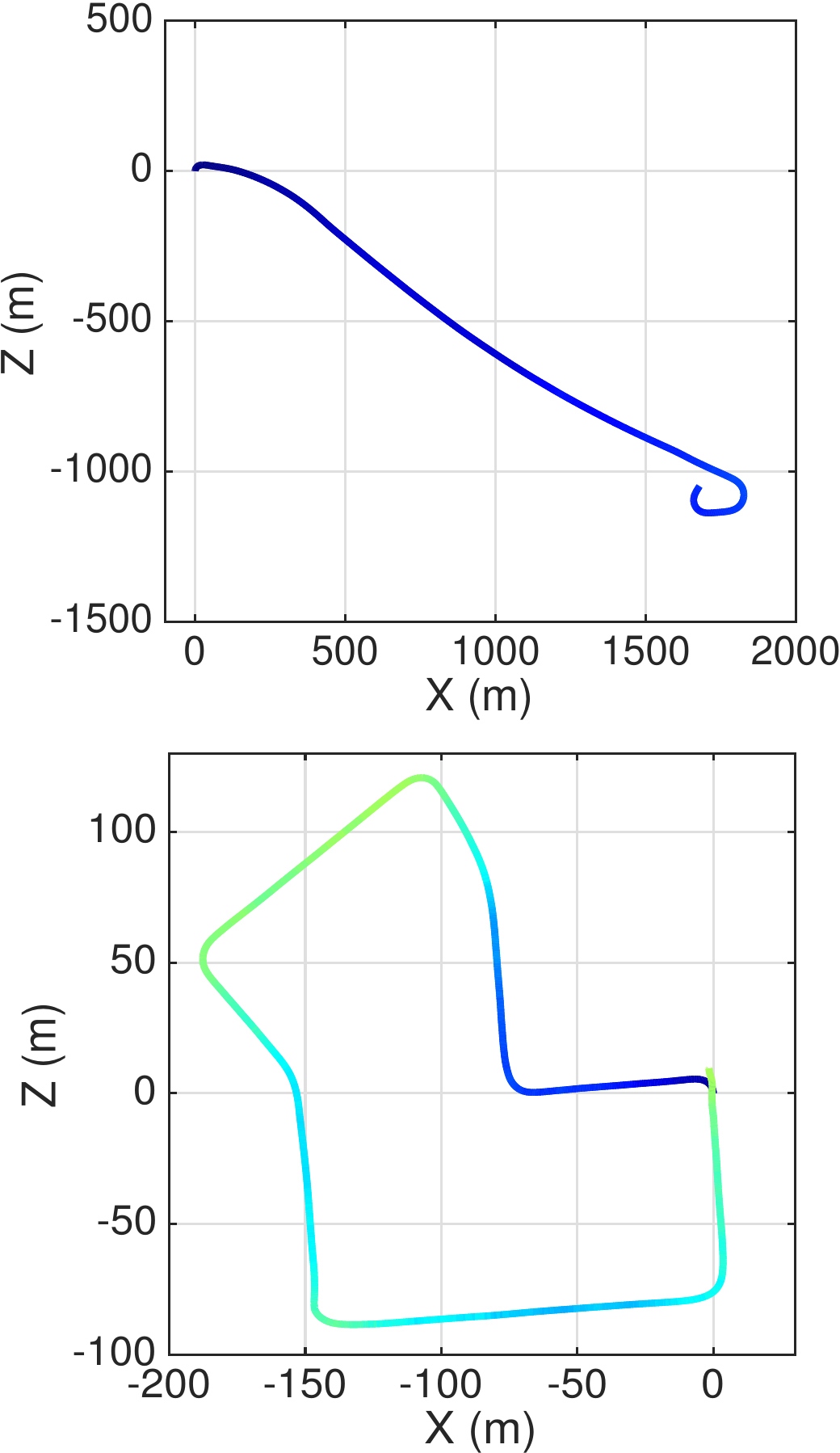}
	}
	\subfloat[MCPA\label{fig:MCPA_traj}]{
		\centering
		\includegraphics[width=0.23\linewidth]{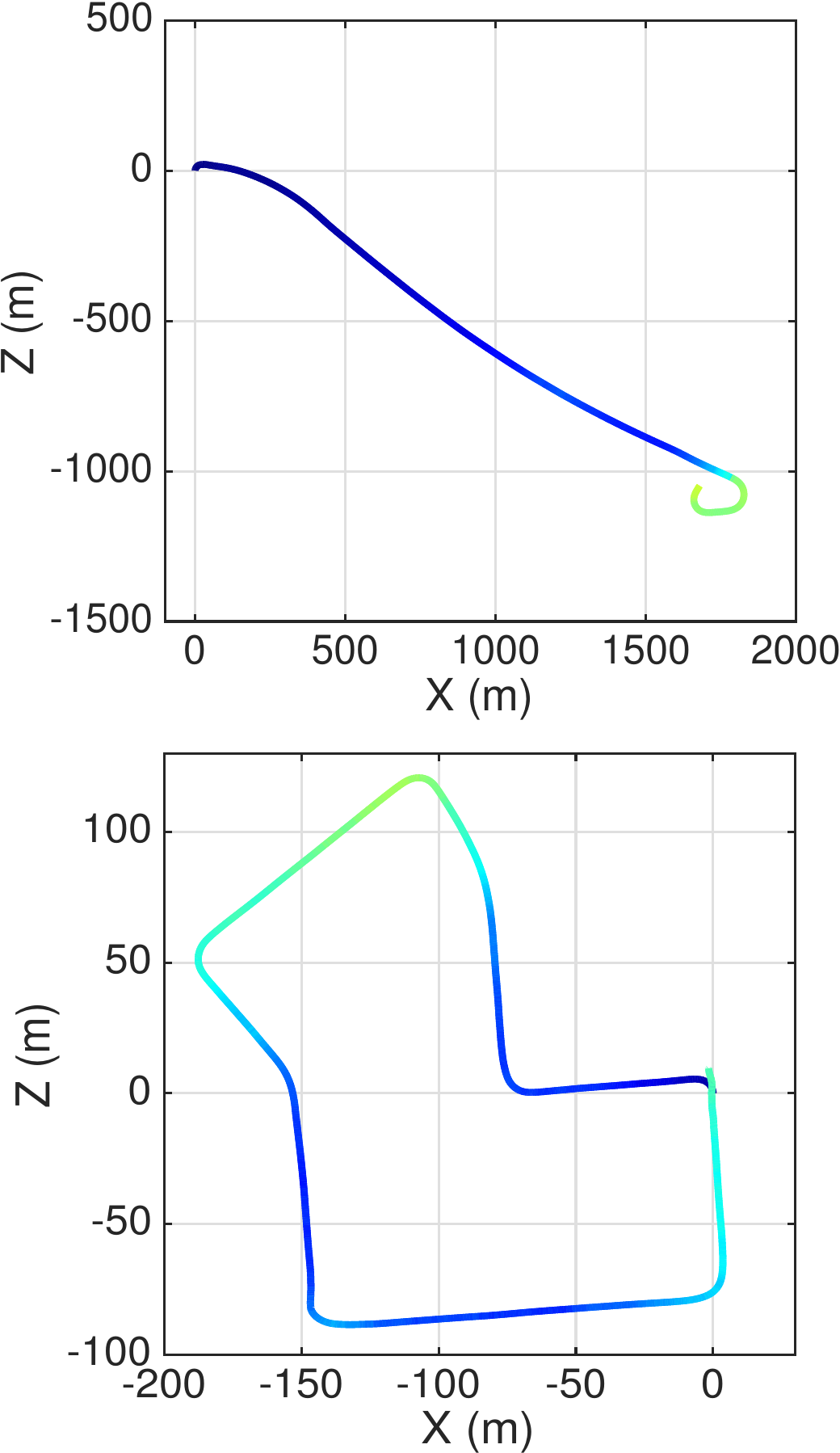}%
	}
	\subfloat[MultiCol\label{fig:MultiCol_traj}]{
		\centering
		\includegraphics[width=0.23\linewidth]{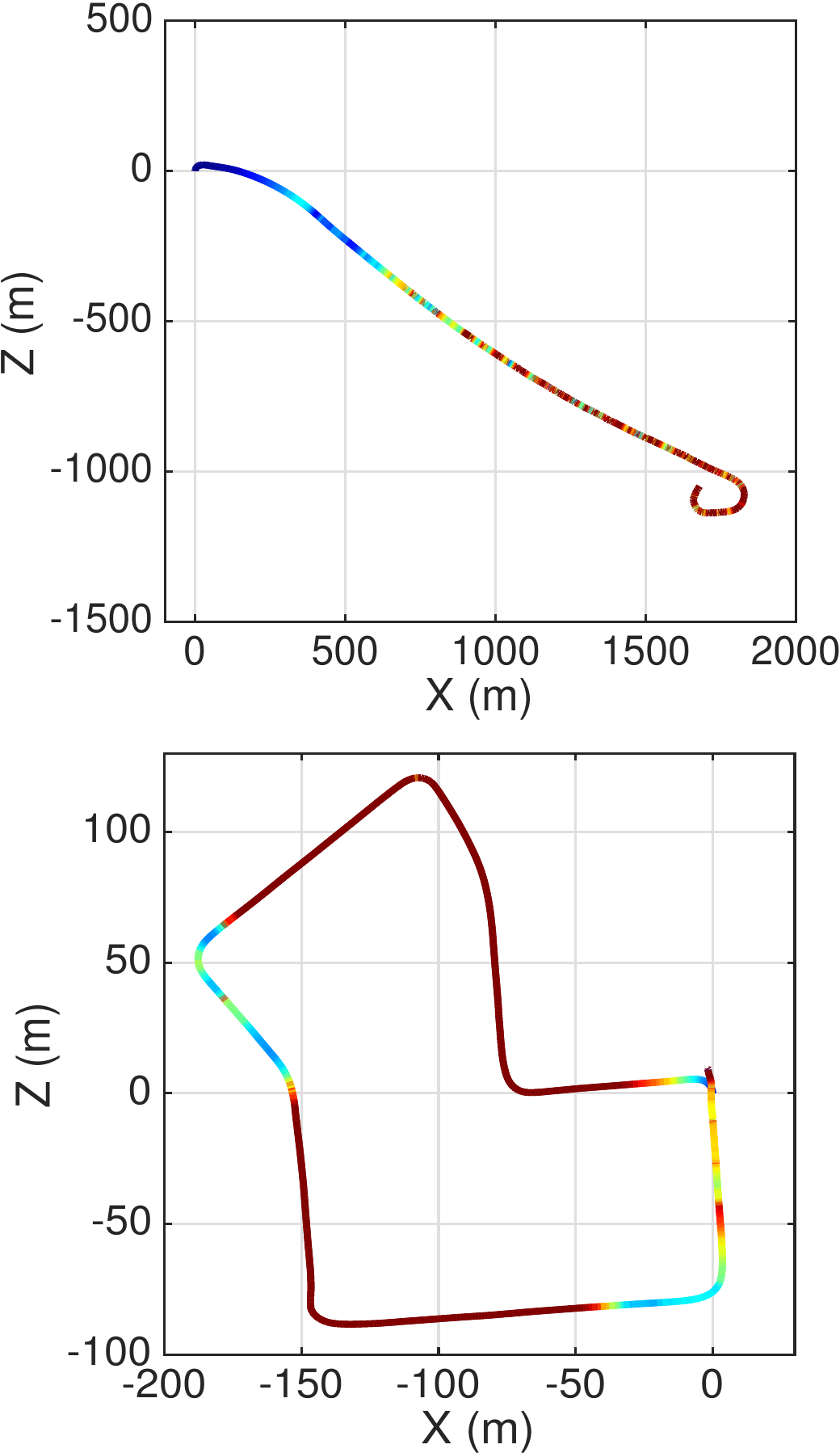}%
	}
	\subfloat[BACS\label{fig:BACS_traj}]{
		\centering
		\includegraphics[width=0.263\linewidth]{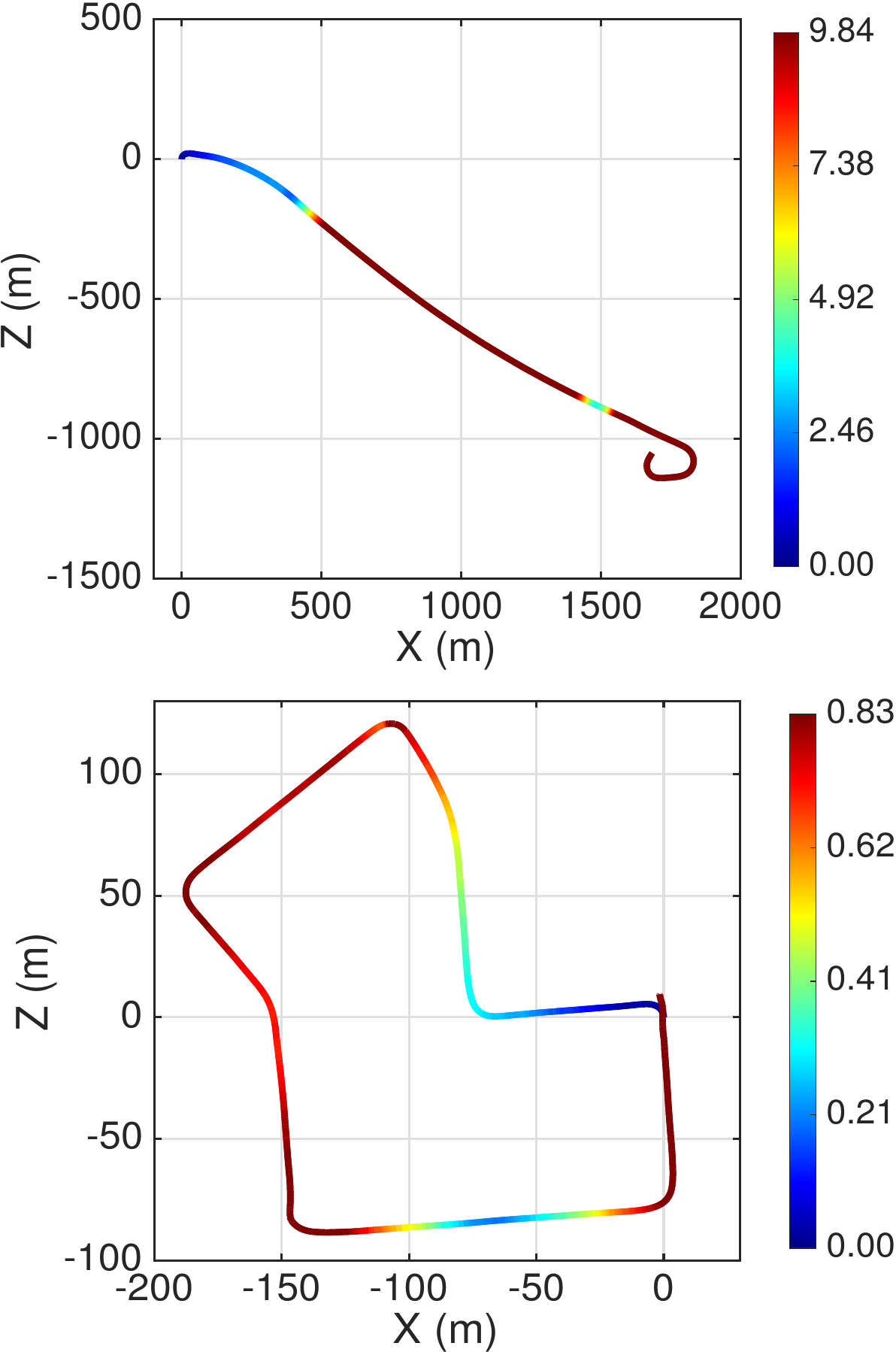}%
	}
	\caption{Comparison of trajectory optimization results for sequences Seq.01 (top row) and Seq.07 (bottom row) of the KITTI dataset under different algorithms. The trajectories are color-coded to represent the absolute trajectory errors, with dark red indicating larger errors and dark blue indicating smaller errors.}
	\label{fig:kitti_0107_traj}
\end{figure*}
The KITTI odometry dataset \cite{Geiger2012KITTI} is a widely used benchmark for visual odometry in the field of autonomous driving. It encompasses a variety of real-world driving scenarios, including city, rural, and highway. In our experiments, we select 11 sequences from the odometry subset, specifically sequences 00-10, totaling over 46,000 images. The multi-camera system used in the KITTI is a stereo vision setup, where intrinsic and extrinsic parameters of each camera have been accurately calibrated. Additionally, the dataset provides a high-precision reference ground truth trajectory. Compared to ETH3D dataset, the KITTI odometry dataset presents greater challenges in several aspects. The KITTI dataset is significantly larger than the ETH3D dataset, with increased data volume and scene complexity. Furthermore, the multi-camera system in the KITTI exhibits forward linear motion in most scenarios, resulting in a small parallax angle, which introduces additional challenges for scene reconstruction.

The feature extraction and matching of images are still performed using COLMAP \cite{Johannes2016colmap}. In the matching process, we customized the match pairs following the temporal sequence. The image sequences are cross-matched among the 50 overlapping poses. Similar to the operation used in ETH3D, we added Gaussian noise to the ground truth poses to obtain initial values for iterative optimization. All algorithms share the same termination criteria for iteration, with a maximum iteration count set to 30. The open-source code of \texttt{BACS} fails in addressing large-scale optimization problems. Therefore, we re-implemented \texttt{BACS} with the objective function focused on minimizing the spherical normalized reprojection error.

{We report the size of the optimization problems for all sequences in the KITTI odometry dataset and the runtime of different algorithms in \cref{tab:kitti_time}.} Compared to the ETH3D dataset, the KITTI dataset has a larger problem scale, with 11.49 times more 3D points and 8.43 times more poses. 
{As shown in the comparison of runtime in \cref{tab:kitti_time}}, the proposed \texttt{MCPA} algorithm demonstrates a significant speedup. Compared to \texttt{MultiCol}, the optimization speed improved by up to 3.82 times (Seq.08), with an average increase of approximately 2.09 times. In most scenes, \texttt{MCPALR} achieves notable speed improvements over \texttt{MultiCol}, with a maximum increase of 2.39 times (Seq.08) and an average increase of 1.29 times. This result validates the effectiveness of the proposed algorithms in handling large-scale pose optimization problems.
\begin{table*}[thb]
	\centering
	\normalsize
	\caption{Accuracy Comparison of Four Multi-Camera Pose Optimization Algorithms on the KITTI Odometry Dataset.}
	\label{tab:kitti_err}
		\begin{tabular}{c ccc ccc ccc ccc}
			\toprule
			& \multicolumn{3}{c}{\texttt{MCPALR}} & \multicolumn{3}{c}{\texttt{MCPA}} & \multicolumn{3}{c}{\texttt{MultiCol}} & \multicolumn{3}{c}{\texttt{BACS}} \\
			\cmidrule(lr){2-4} \cmidrule(lr){5-7} \cmidrule(lr){8-10} \cmidrule(lr){11-13}
			\textbf{Seq.} & $\bar{\varepsilon}_{\mathbf{R}}(^\circ)$ & $\bar{\varepsilon}_{\mathbf{t}}(\%)$ & $\bar{\varepsilon}_{\mathbf{p}}(px)$ & 
			$\bar{\varepsilon}_{\mathbf{R}}(^\circ)$ & $\bar{\varepsilon}_{\mathbf{t}}(\%)$ & $\bar{\varepsilon}_{\mathbf{p}}(px)$ & 
			$\bar{\varepsilon}_{\mathbf{R}}(^\circ)$ & $\bar{\varepsilon}_{\mathbf{t}}(\%)$ & $\bar{\varepsilon}_{\mathbf{p}}(px)$ & 
			$\bar{\varepsilon}_{\mathbf{R}}(^\circ)$ & $\bar{\varepsilon}_{\mathbf{t}}(\%)$ & $\bar{\varepsilon}_{\mathbf{p}}(px)$ \\
			\midrule
			00  &\textbf{0.0057} & \textbf{0.0140} & \textbf{0.4287} & \underline{0.0082} & 0.0224 & 0.4339 & 0.0120 & 0.0144 & 0.4305 & \textbf{0.0057} & \underline{0.0142} & \underline{0.4304} \\
			01 & \underline{0.0173} & \textbf{0.0090} & \textbf{0.6063} & \textbf{0.0166} & \underline{0.0185} & \underline{0.6081} & 0.1147 & 0.1460 & 2.9201 & 0.1253 & 0.2116 & 1.0827 \\
			02 & \textbf{0.1185} & \textbf{0.1464} & \underline{0.8141} & 0.2380 & 0.2202 & \textbf{0.8021} & \underline{0.1902} & \underline{0.1858} & 2.221 & 0.2057 & 0.3149 & 0.8743 \\
			03 & {0.0156} & 0.0238 & \underline{1.2369} & \textbf{0.0117} & \underline{0.0184} & \textbf{1.2071} & 0.1805 & 0.2748 & 3.0899 & \underline{0.0135} & \textbf{0.0163} & 1.2402 \\
			04   & \textbf{0.1621} & \underline{0.3754} & \textbf{0.5507} & \underline{0.1933} & \textbf{0.3584} & \underline{0.5792} & 0.3946 & 0.6969 & 1.1203 & 0.2007 & 0.3998 & 0.5829 \\
			05  & \underline{0.0410} & \textbf{0.0691} & \textbf{0.9040} & 0.0459 & 0.1082 & 0.9145 & 0.1116 & 0.1935 & 1.0871 & \textbf{0.0396} & \underline{0.1077} & \underline{0.9116} \\
			06 & \underline{0.0344} & \textbf{0.2609} & \textbf{0.6087} & \textbf{0.0335} & \underline{0.2967} & 0.6134 & 0.2141 & 0.3312 & 1.0263 & 0.0771 & 1.0251 & \underline{0.6104} \\
			07 & \underline{0.0127} & \underline{0.1255} & 0.8143 & \textbf{0.0114} & \textbf{0.1078} & \textbf{0.8034} & 0.1640 & 0.2653 & 0.8336 & 0.0390 & 0.2909 & \underline{0.8110} \\
			08   & 0.0102 & \textbf{0.0312} & 0.4941 & \textbf{0.0084} & 0.0546 & \textbf{0.4880} & \underline{0.0101} & 0.0363 & 0.4900 & 0.0097 & \underline{0.0315} & \underline{0.4899} \\
			09   & 0.0321 & \underline{0.1435} & 0.6318 & \textbf{0.0238} & \textbf{0.0274} & \textbf{0.6223} & 0.1336 & 0.1668 & 1.4857 & \underline{0.0266} & 0.1779 & \underline{0.6232} \\
			10 & \textbf{0.0649} & \textbf{0.0691} & \textbf{0.6096} & 0.1114 & 0.0856 & \underline{0.6195} & 0.1698 & 0.2209 & 2.4011 & \underline{0.0697} & \underline{0.0724} & 0.6792\\
			\bottomrule
		\end{tabular}
	\begin{tablenotes}
		\footnotesize
		\item \hspace{-1.5em} \textbf{Bold} values indicate the smallest error, and \underline{underline} values indicate the second smallest.
	\end{tablenotes}
\end{table*}
The qualitative results of pose optimization for Seq.01 and Seq.07 in the KITTI odometry dataset are shown in \cref{fig:kitti_010507_recon}. \Cref{fig:Init_07_recon} displays the initial trajectories of the multi-camera system and scene point clouds. The initial trajectories exhibit significant errors, characterized by drift and jitter, and the point cloud structures appear blurry. \Cref{fig:MCPALR_07_recon} and \Cref{fig:MCPA_07_recon} show the results after pose optimization and reconstruction using the proposed \texttt{MCPALR} and \texttt{MCPA} algorithms, respectively. After optimization, the trajectories become smoother and more stable, and the scene structures are clearer. \Cref{fig:kitti_0107_traj} shows the trajectories of the multi-camera system optimized by four different algorithms, with color coding used to represent the absolute trajectory errors along the trajectories visually. The comparison indicates that the trajectories optimized by our proposed \texttt{MCPALR} and \texttt{MCPA} algorithms (\cref{fig:MCPALR_traj} and \cref{fig:MCPA_traj}) exhibit significantly smaller absolute trajectory errors. The quantitative comparison of the final optimization accuracy for the four algorithms is presented in \cref{tab:kitti_err}. In terms of accuracy, \texttt{MCPA} and \texttt{MCPALR} demonstrate the best optimization performance on most sequences. In summary, these experimental results demonstrate that our algorithm significantly outperforms existing baselines in terms of computational efficiency, and exhibits strong competitiveness in pose optimization accuracy.

{To comprehensively evaluate the proposed algorithm in realistic SfM or SLAM scenarios, we conducted comparative experiments using initial poses obtained from a real pose estimation pipeline. Unlike adding Gaussian noise to ground-truth poses, this pipeline more realistically captures the effects of accumulated drift and observation noise inherent in practical systems. We evaluate our method on two loop-closing sequences Seq.07 and Seq.09. Specifically, we employ COLMAP for feature extraction and matching. The multi-camera system is modeled as a generalized camera using the given calibration parameters. Relative poses between adjacent frames are estimated using the 17-point algorithm \cite{li2008linear} with RANSAC for outlier rejection, followed by nonlinear optimization. The pose estimation solvers are implemented using the OpenGV library \cite{kneip2014opengv}. These relative poses are then integrated to obtain global frame poses in a world coordinate system defined by the first frame. Owing to the generalized camera formulation, the estimated poses inherently include scale information. The 3D scene points required by \texttt{MultiCol} and \texttt{BACS} are obtained through triangulation using the estimated poses. All other experimental settings remain consistent.}

{Qualitative results are presented in \cref{fig:kitti_pose_0709_traj}. The initial trajectory obtained from the pose estimation pipeline (gray dashed line) deviates significantly from the ground-truth trajectory (black solid line) due to error accumulation. After optimization by all methods, the trajectories (color-coded solid lines) show clear improvement. Among them, the proposed methods \texttt{MCPALR} and \texttt{MCPA} yield trajectories that best align with the ground truth. In contrast, \texttt{MultiCol} and \texttt{BACS} partially correct the initial errors and still show noticeable deviations in regions with complex motion. Quantitative results are reported in \cref{table:kitti_pose_0709}. The initial trajectories exhibit large errors, with average rotation errors exceeding $10^\circ$ and translation errors surpassing 50\%. After optimization, all methods significantly reduce these errors. Notably, \texttt{MCPA} achieves the best accuracy on two sequences. Compared to the baseline methods, \texttt{MCPA} reduces rotation error by 31.0\% and translation error by 20.6\%. These results strongly validate the effectiveness of multi-camera pose-only constraint and demonstrate the practical value of the proposed algorithm.}
\begin{figure*}[tbp]
	\centering
	\subfloat[MCPALR\label{fig:MCPALR0709}]{
		\centering
		\includegraphics[width=0.23\linewidth]{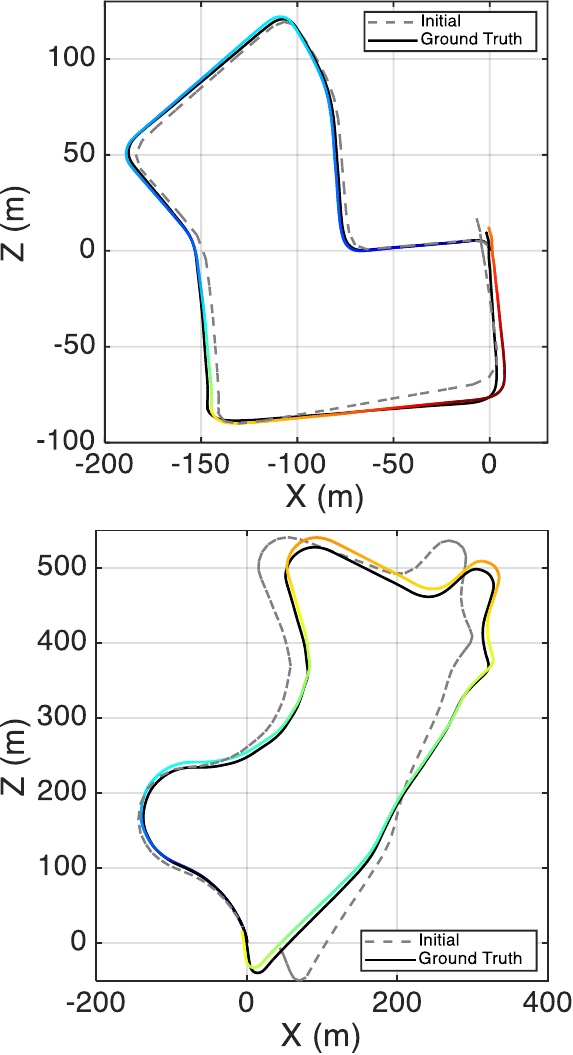}
	}
	\subfloat[MCPA\label{fig:MCPA0709}]{
		\centering
		\includegraphics[width=0.23\linewidth]{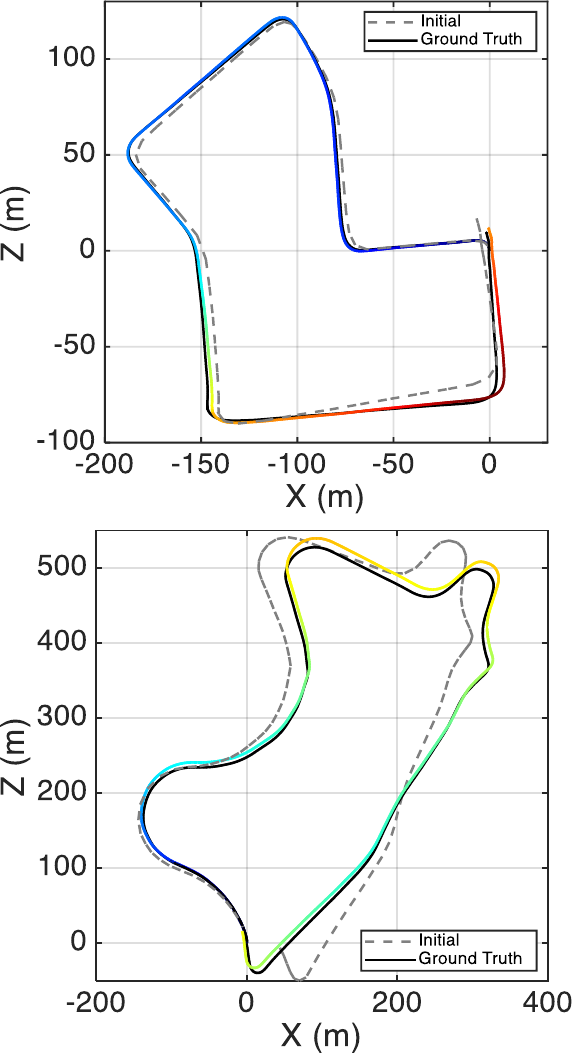}%
	}
	\subfloat[MultiCol\label{fig:MultiCol0709}]{
		\centering
		\includegraphics[width=0.23\linewidth]{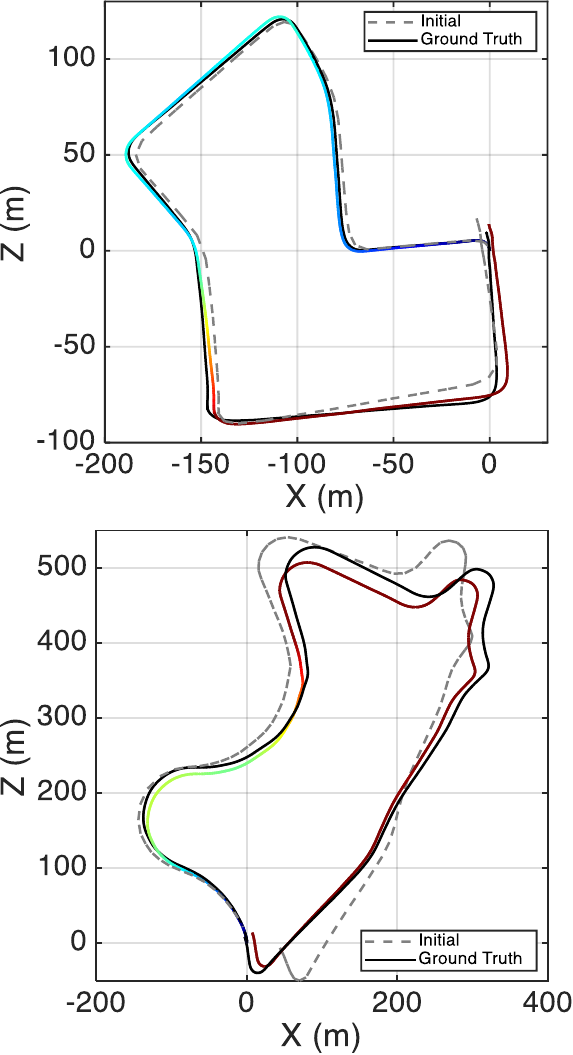}%
	}
	\subfloat[BACS\label{fig:BACS0709}]{
		\centering
		\includegraphics[width=0.273\linewidth]{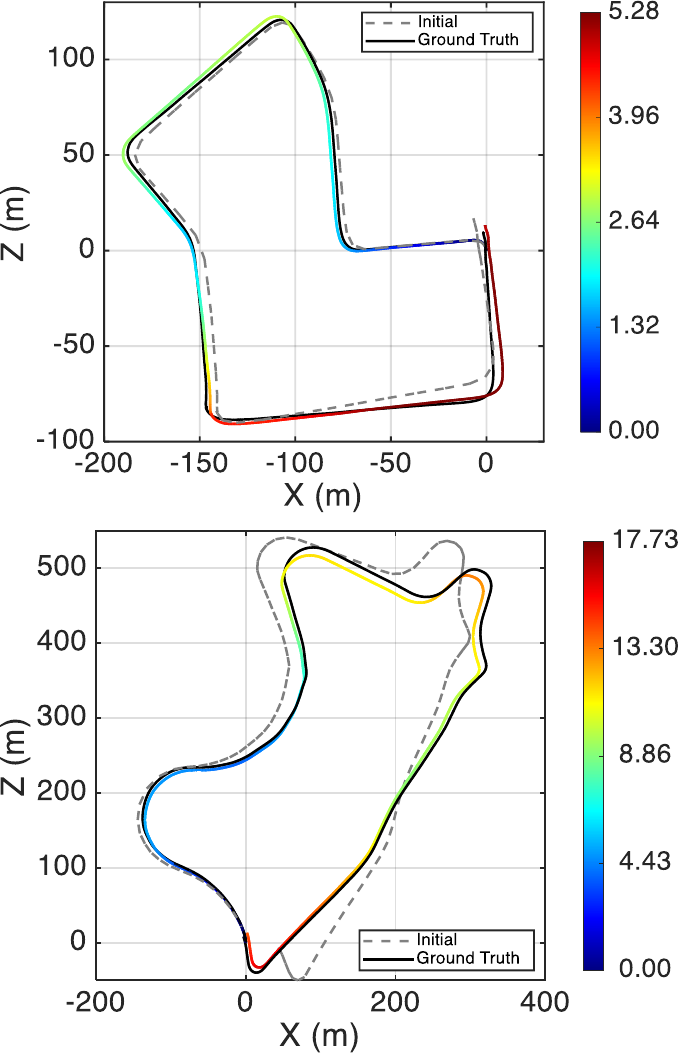}%
	}
	\caption{Comparison of trajectory optimization results on KITTI sequences Seq.07 (top row) and Seq.09 (bottom row), initialized using a realistic pose estimation pipeline. Gray dashed lines indicate the initial trajectories, black solid lines represent the ground truth, and colorcoded solid lines show the optimized trajectories obtained by different methods.}
	\label{fig:kitti_pose_0709_traj}
\end{figure*}
\begin{table}
	\centering
	\caption{Accuracy Comparison of Optimization Algorithms Initialized from the Pose Estimation Pipeline on KITTI Seq.07 and Seq.09.}
	\label{table:kitti_pose_0709}
	\begin{tabular}{clrrrrr}
		\toprule
		{Seq.} & {Metric} & {Initial} & {MCPALR} & {MCPA} & {MultiCol} & {BACS} \\
		\midrule
		\multirow{3}{*}{07} & $\bar{\varepsilon}_\mathbf{R}(^\circ)$ & 10.669 & \underline{1.007} & \textbf{0.947} & 1.550 & 1.373 \\
		{} & $\bar{\varepsilon}_\mathbf{t}(\%)$ & 56.291 & \underline{11.553} & \textbf{11.307} &17.188 & 14.245 \\
		{} & $\bar{\varepsilon}_{\mathbf{p}}(px)$ & 3.794 & \underline{0.699} & \textbf{0.676} & 0.745 & {0.705} \\
		\midrule
		\multirow{3}{*}{09} & $\bar{\varepsilon}_\mathbf{R}(^\circ)$ & 12.024 & \underline{0.699} & \textbf{0.681} & 2.312 & 1.385 \\
		{} & $\bar{\varepsilon}_\mathbf{t}(\%)$ & 52.405 & \underline{8.633} & \textbf{8.296} & 19.513 & {10.944} \\
		{} & $\bar{\varepsilon}_{\mathbf{p}}(px)$ & 2.521 & \underline{0.640} & \textbf{0.635} & 1.167 & 1.169 \\
		\bottomrule
	\end{tabular}
	\begin{tablenotes}
		\scriptsize
		\item \hspace{-1.5em} \textbf{Bold} values indicate the smallest error, and \underline{underline} indicate the second smallest.
	\end{tablenotes}
\end{table}
\section{Conclusion}
\label{sec:conclusion}

This paper proposes a novel pose adjustment method for multi-camera systems. Based on the generalized camera model, we introduce a multi-camera pose constraint, which implicitly represents 3D points as functions of the base observations and corresponding poses. Furthermore, the multi-camera pose adjustment algorithm is developed to achieve pose optimization without relying on 3D points. By removing 3D points from the optimization variables, the algorithm significantly enhances the computational efficiency when handling a large number of redundant observations. We further propose a base observations selection algorithm based on uncertainty ellipsoid roundness. Additionally, we establish a statistically optimal generalized camera reconstruction algorithm, which provides a closed-form solution for multi-observation intersection, thereby achieving efficient and accurate scene reconstruction. In experiments using synthetic data and real datasets, we validate the optimization accuracy and efficiency of our multi-camera pose adjustment algorithm. Compared to traditional bundle adjustment baseline methods, the proposed algorithm achieves a significant improvement in computational efficiency, while also demonstrating greater competitiveness in pose optimization accuracy.

However, our method assumes the multi-camera system has already been fully calibrated. Future research could further incorporate the calibration parameters of the multi-camera system into the optimization variables to accommodate more complex scenarios and application requirements.

\bibliographystyle{IEEEtran}
{
	\footnotesize
	\bibliography{main}

\begin{thebibliography}{10}
\providecommand{\url}[1]{#1}
\csname url@samestyle\endcsname
\providecommand{\newblock}{\relax}
\providecommand{\bibinfo}[2]{#2}
\providecommand{\BIBentrySTDinterwordspacing}{\spaceskip=0pt\relax}
\providecommand{\BIBentryALTinterwordstretchfactor}{4}
\providecommand{\BIBentryALTinterwordspacing}{\spaceskip=\fontdimen2\font plus
\BIBentryALTinterwordstretchfactor\fontdimen3\font minus
  \fontdimen4\font\relax}
\providecommand{\BIBforeignlanguage}[2]{{%
\expandafter\ifx\csname l@#1\endcsname\relax
\typeout{** WARNING: IEEEtran.bst: No hyphenation pattern has been}%
\typeout{** loaded for the language `#1'. Using the pattern for}%
\typeout{** the default language instead.}%
\else
\language=\csname l@#1\endcsname
\fi
#2}}
\providecommand{\BIBdecl}{\relax}
\BIBdecl

\bibitem{Liu2018Towards}
P.~Liu, M.~Geppert, L.~Heng, T.~Sattler, A.~Geiger, and M.~Pollefeys, ``Towards
  robust visual odometry with a multi-camera system,'' in \emph{IEEE
  International Conference on Intelligent Robots and Systems}, 2018, pp.
  1154--1161.

\bibitem{Forster2017SVO}
C.~Forster, Z.~Zhang, M.~Gassner, M.~Werlberger, and D.~Scaramuzza, ``Svo:
  Semidirect visual odometry for monocular and multicamera systems,''
  \emph{IEEE Transactions on Robotics}, vol.~33, no.~2, pp. 249--265, 2017.

\bibitem{Cui2023mcsfm}
H.~Cui, X.~Gao, and S.~Shen, ``Mcsfm: Multi-camera-based incremental
  structure-from-motion,'' \emph{IEEE Transactions on Image Processing},
  vol.~32, pp. 6441--6456, 2023.

\bibitem{Schmied2023r3d3}
A.~Schmied, T.~Fischer, M.~Danelljan, M.~Pollefeys, and F.~Yu, ``R3d3: Dense 3d
  reconstruction of dynamic scenes from multiple cameras,'' in \emph{IEEE
  International Conference on Computer Vision}, 2023, pp. 3193--3203.

\bibitem{Qin2020AVP}
T.~Qin, T.~Chen, Y.~Chen, and Q.~Su, ``Avp-slam: Semantic visual mapping and
  localization for autonomous vehicles in the parking lot,'' in \emph{IEEE
  International Conference on Intelligent Robots and Systems}, 2020, pp.
  5939--5945.

\bibitem{Heng2019AutoVision}
L.~Heng, B.~Choi, Z.~Cui, M.~Geppert, S.~Hu, B.~Kuan, P.~Liu, R.~Nguyen, Y.~C.
  Yeo, A.~Geiger, G.~H. Lee, M.~Pollefeys, and T.~Sattler, ``Project
  autovision: Localization and 3d scene perception for an autonomous vehicle
  with a multi-camera system,'' in \emph{International Conference on Robotics
  and Automation}, 2019, pp. 4695--4702.

\bibitem{Triggs2000bundle}
B.~Triggs, P.~F. McLauchlan, R.~I. Hartley, and A.~W. Fitzgibbon, ``Bundle
  adjustment—a modern synthesis,'' in \emph{Vision Algorithms: Theory and
  Practice: International Workshop on Vision Algorithms Corfu, Greece,
  September 21--22, 1999 Proceedings}.\hskip 1em plus 0.5em minus 0.4em\relax
  Springer, 2000, pp. 298--372.

\bibitem{Agarwal2010bundle}
S.~Agarwal, N.~Snavely, S.~M. Seitz, and R.~Szeliski, ``Bundle adjustment in
  the large,'' in \emph{European Conference on Computer Vision}.\hskip 1em plus
  0.5em minus 0.4em\relax Springer, 2010, pp. 29--42.

\bibitem{Agarwal_Ceres_Solver_2022}
\BIBentryALTinterwordspacing
S.~Agarwal, K.~Mierle, and T.~C.~S. Team, ``{Ceres Solver},'' 10 2023.
  [Online]. Available: \url{https://github.com/ceres-solver/ceres-solver}
\BIBentrySTDinterwordspacing

\bibitem{Ye2022coli}
Z.~Ye, G.~Li, H.~Liu, Z.~Cui, H.~Bao, and G.~Zhang, ``Coli-ba: Compact
  linearization based solver for bundle adjustment,'' \emph{IEEE Transactions
  on Visualization and Computer Graphics}, vol.~28, no.~11, pp. 3727--3736,
  2022.

\bibitem{Weber2023power}
S.~Weber, N.~Demmel, T.~C. Chan, and D.~Cremers, ``Power bundle adjustment for
  large-scale 3d reconstruction,'' in \emph{IEEE Conference on Computer Vision
  and Pattern Recognition}, 2023, pp. 281--289.

\bibitem{Schneider2012bundle}
J.~Schneider, F.~Schindler, T.~L{\"a}be, and W.~F{\"o}rstner, ``Bundle
  adjustment for multi-camera systems with points at infinity,'' \emph{ISPRS
  Annals of the Photogrammetry, Remote Sensing and Spatial Information
  Sciences}, vol.~1, pp. 75--80, 2012.

\bibitem{Urban2017multicol}
S.~Urban, S.~Wursthorn, J.~Leitloff, and S.~Hinz, ``Multicol bundle adjustment:
  a generic method for pose estimation, simultaneous self-calibration and
  reconstruction for arbitrary multi-camera systems,'' \emph{International
  Journal of Computer Vision}, vol. 121, pp. 234--252, 2017.

\bibitem{Cai2023pose}
Q.~Cai, L.~Zhang, Y.~Wu, W.~Yu, and D.~Hu, ``A pose-only solution to visual
  reconstruction and navigation,'' \emph{IEEE Transactions on Pattern Analysis
  and Machine Intelligence}, vol.~45, no.~1, pp. 73--86, 2023.

\bibitem{Cai2019equivalent}
Q.~Cai, Y.~Wu, L.~Zhang, and P.~Zhang, ``Equivalent constraints for two-view
  geometry: Pose solution/pure rotation identification and 3d reconstruction,''
  \emph{International Journal of Computer Vision}, vol. 127, pp. 163--180,
  2019.

\bibitem{Pless2003using}
R.~Pless, ``Using many cameras as one,'' in \emph{IEEE Computer Society
  Conference on Computer Vision and Pattern Recognition}, vol.~2.\hskip 1em
  plus 0.5em minus 0.4em\relax IEEE, 2003, pp. II--587.

\bibitem{Davison2007MonoSLAM}
A.~J. Davison, I.~D. Reid, N.~D. Molton, and O.~Stasse, ``Monoslam: Real-time
  single camera slam,'' \emph{IEEE Transactions on Pattern Analysis and Machine
  Intelligence}, vol.~29, no.~6, pp. 1052--1067, 2007.

\bibitem{Forster2014svo}
C.~Forster, M.~Pizzoli, and D.~Scaramuzza, ``Svo: Fast semi-direct monocular
  visual odometry,'' in \emph{IEEE international conference on robotics and
  automation}.\hskip 1em plus 0.5em minus 0.4em\relax IEEE, 2014, pp. 15--22.

\bibitem{Mur2017orbslam2}
R.~Mur-Artal and J.~D. Tardós, ``Orb-slam2: An open-source slam system for
  monocular, stereo, and rgb-d cameras,'' \emph{IEEE Transactions on Robotics},
  vol.~33, no.~5, pp. 1255--1262, 2017.

\bibitem{Wang2017stereo}
R.~Wang, M.~Schworer, and D.~Cremers, ``Stereo dso: Large-scale direct sparse
  visual odometry with stereo cameras,'' in \emph{IEEE international conference
  on computer vision}, 2017, pp. 3903--3911.

\bibitem{Qin2018vins}
T.~Qin, P.~Li, and S.~Shen, ``Vins-mono: A robust and versatile monocular
  visual-inertial state estimator,'' \emph{IEEE transactions on robotics},
  vol.~34, no.~4, pp. 1004--1020, 2018.

\bibitem{Leutenegger2015keyframe}
S.~Leutenegger, S.~Lynen, M.~Bosse, R.~Siegwart, and P.~Furgale,
  ``Keyframe-based visual--inertial odometry using nonlinear optimization,''
  \emph{The International Journal of Robotics Research}, vol.~34, no.~3, pp.
  314--334, 2015.

\bibitem{Tribou2015multi}
M.~J. Tribou, A.~Harmat, D.~W. Wang, I.~Sharf, and S.~L. Waslander,
  ``Multi-camera parallel tracking and mapping with non-overlapping fields of
  view,'' \emph{The International Journal of Robotics Research}, vol.~34,
  no.~12, pp. 1480--1500, 2015.

\bibitem{Klein2007parallel}
G.~Klein and D.~Murray, ``Parallel tracking and mapping for small ar
  workspaces,'' in \emph{IEEE and ACM international symposium on mixed and
  augmented reality}.\hskip 1em plus 0.5em minus 0.4em\relax IEEE, 2007, pp.
  225--234.

\bibitem{Urban2016multicol}
S.~Urban and S.~Hinz, ``Multicol-slam-a modular real-time multi-camera slam
  system,'' \emph{arXiv preprint arXiv:1610.07336}, 2016.

\bibitem{He2022towards}
Y.~He, H.~Yu, W.~Yang, and S.~Scherer, ``Towards robust visual-inertial
  odometry with multiple non-overlapping monocular cameras,'' in \emph{IEEE
  International Conference on Intelligent Robots and Systems}.\hskip 1em plus
  0.5em minus 0.4em\relax IEEE, 2022, pp. 9452--9458.

\bibitem{Zhang2023bamf}
W.~Zhang, S.~Wang, X.~Dong, R.~Guo, and N.~Haala, ``Bamf-slam: Bundle adjusted
  multi-fisheye visual-inertial slam using recurrent field transforms,'' in
  \emph{IEEE international conference on robotics and automation}.\hskip 1em
  plus 0.5em minus 0.4em\relax IEEE, 2023, pp. 6232--6238.

\bibitem{Wang2024mavis}
Y.~Wang, Y.~Ng, I.~Sa, A.~Parra, C.~Rodriguez-Opazo, T.~Lin, and H.~Li,
  ``Mavis: Multi-camera augmented visual-inertial slam using se 2 (3) based
  exact imu pre-integration,'' in \emph{IEEE International Conference on
  Robotics and Automation}.\hskip 1em plus 0.5em minus 0.4em\relax IEEE, 2024,
  pp. 1694--1700.

\bibitem{Zhang2021balancing}
L.~Zhang, D.~Wisth, M.~Camurri, and M.~Fallon, ``Balancing the budget: Feature
  selection and tracking for multi-camera visual-inertial odometry,''
  \emph{IEEE Robotics and Automation Letters}, vol.~7, no.~2, pp. 1182--1189,
  2021.

\bibitem{Kuo2020redesigning}
J.~Kuo, M.~Muglikar, Z.~Zhang, and D.~Scaramuzza, ``Redesigning slam for
  arbitrary multi-camera systems,'' in \emph{IEEE International Conference on
  Robotics and Automation}.\hskip 1em plus 0.5em minus 0.4em\relax IEEE, 2020,
  pp. 2116--2122.

\bibitem{Kaveti2023Design}
P.~Kaveti, S.~N. Vaidyanathan, A.~T. Chelvan, and H.~Singh, ``Design and
  evaluation of a generic visual slam framework for multi camera systems,''
  \emph{IEEE Robotics and Automation Letters}, vol.~8, no.~11, pp. 7368--7375,
  2023.

\bibitem{Hartley2003multiple}
R.~Hartley and A.~Zisserman, \emph{Multiple view geometry in computer
  vision}.\hskip 1em plus 0.5em minus 0.4em\relax Cambridge university press,
  2003.

\bibitem{Delaunoy2014photometric}
A.~Delaunoy and M.~Pollefeys, ``Photometric bundle adjustment for dense
  multi-view 3d modeling,'' in \emph{IEEE Conference on Computer Vision and
  Pattern Recognition}, 2014, pp. 1486--1493.

\bibitem{Engel2017direct}
J.~Engel, V.~Koltun, and D.~Cremers, ``Direct sparse odometry,'' \emph{IEEE
  transactions on pattern analysis and machine intelligence}, vol.~40, no.~3,
  pp. 611--625, 2017.

\bibitem{guan2025affine}
B.~Guan and J.~Zhao, ``Affine correspondences between multi-camera systems for
  relative pose estimation,'' \emph{IEEE Transactions on Pattern Analysis and
  Machine Intelligence}, 2025.

\bibitem{Johannes2016colmap}
J.~L. Schönberger and J.-M. Frahm, ``Structure-from-motion revisited,'' in
  \emph{IEEE Conference on Computer Vision and Pattern Recognition}, 2016, pp.
  4104--4113.

\bibitem{Liang2024camera}
S.~Liang, B.~Guan, Z.~Yu, P.~Sun, and Y.~Shang, ``Camera calibration using a
  collimator system,'' in \emph{European Conference on Computer Vision}.\hskip
  1em plus 0.5em minus 0.4em\relax Springer, 2024, pp. 374--390.

\bibitem{tan2026optimal}
D.~Tan, S.~Liang, B.~Li, B.~Guan, A.~Su, Y.~Lin, D.~Zhang, M.~Wan, Z.~Liu,
  C.~Wang \emph{et~al.}, ``Optimal pose guidance for stereo calibration in 3d
  deformation measurement,'' \emph{Experimental Mechanics}, pp. 1--14, 2026.

\bibitem{Demmel2021square}
N.~Demmel, C.~Sommer, D.~Cremers, and V.~Usenko, ``Square root bundle
  adjustment for large-scale reconstruction,'' in \emph{IEEE Conference on
  Computer Vision and Pattern Recognition}, 2021, pp. 11\,718--11\,727.

\bibitem{zhou2020stochastic}
L.~Zhou, Z.~Luo, M.~Zhen, T.~Shen, S.~Li, Z.~Huang, T.~Fang, and L.~Quan,
  ``Stochastic bundle adjustment for efficient and scalable 3d
  reconstruction,'' in \emph{European Conference on Computer Vision}.\hskip 1em
  plus 0.5em minus 0.4em\relax Springer, 2020, pp. 364--379.

\bibitem{wu2011multicore}
C.~Wu, S.~Agarwal, B.~Curless, and S.~M. Seitz, ``Multicore bundle
  adjustment,'' in \emph{IEEE Conference on Computer Vision and Pattern
  Recognition}.\hskip 1em plus 0.5em minus 0.4em\relax IEEE, 2011, pp.
  3057--3064.

\bibitem{Ren2022megba}
J.~Ren, W.~Liang, R.~Yan, L.~Mai, S.~Liu, and X.~Liu, ``Megba: A gpu-based
  distributed library for large-scale bundle adjustment,'' in \emph{European
  Conference on Computer Vision}.\hskip 1em plus 0.5em minus 0.4em\relax
  Springer, 2022, pp. 715--731.

\bibitem{Lu1997globally}
F.~Lu and E.~Milios, ``Globally consistent range scan alignment for environment
  mapping,'' \emph{Autonomous robots}, vol.~4, pp. 333--349, 1997.

\bibitem{Lee2021rotation}
S.~H. Lee and J.~Civera, ``Rotation-only bundle adjustment,'' in \emph{IEEE
  Conference on Computer Vision and Pattern Recognition}, 2021, pp. 424--433.

\bibitem{Kneip2013direct}
L.~Kneip and S.~Lynen, ``Direct optimization of frame-to-frame rotation,'' in
  \emph{IEEE International Conference on Computer Vision}, 2013, pp.
  2352--2359.

\bibitem{Ge2024pipo}
Y.~Ge, L.~Zhang, Y.~Wu, and D.~Hu, ``Pipo-slam: Lightweight visual-inertial
  slam with preintegration merging theory and pose-only descriptions of
  multiple view geometry,'' \emph{IEEE Transactions on Robotics}, vol.~40, pp.
  2046--2059, 2024.

\bibitem{Wang2025pokf}
L.~Wang, H.~Tang, T.~Zhang, Y.~Wang, Q.~Zhang, and X.~Niu, ``Po-kf: A pose-only
  representation-based kalman filter for visual inertial odometry,'' \emph{IEEE
  Internet of Things Journal}, 2025.

\bibitem{Scaramuzza2006Omnidirectional}
D.~Scaramuzza, A.~Martinelli, and R.~Siegwart, ``A flexible technique for
  accurate omnidirectional camera calibration and structure from motion,'' in
  \emph{IEEE International Conference on Computer Vision Systems}, 2006, pp.
  45--45.

\bibitem{Kannala2006generic}
J.~Kannala and S.~Brandt, ``A generic camera model and calibration method for
  conventional, wide-angle, and fish-eye lenses,'' \emph{IEEE Transactions on
  Pattern Analysis and Machine Intelligence}, vol.~28, no.~8, pp. 1335--1340,
  2006.

\bibitem{Zhao2015parallaxba}
L.~Zhao, S.~Huang, Y.~Sun, L.~Yan, and G.~Dissanayake, ``Parallaxba: bundle
  adjustment using parallax angle feature parametrization,'' \emph{The
  International Journal of Robotics Research}, vol.~34, no. 4-5, pp. 493--516,
  2015.

\bibitem{Beder2006determining}
C.~Beder and R.~Steffen, ``Determining an initial image pair for fixing the
  scale of a 3d reconstruction from an image sequence,'' in \emph{Joint Pattern
  Recognition Symposium}.\hskip 1em plus 0.5em minus 0.4em\relax Springer,
  2006, pp. 657--666.

\bibitem{helmert1872ausgleichungsrechnung}
F.~R. Helmert, \emph{Die Ausgleichungsrechnung nach der Methode der kleinsten
  Quadrate: mit Anwendungen auf die Geod{\"a}sie und die Theorie der
  Messinstrumente}.\hskip 1em plus 0.5em minus 0.4em\relax Leipzig: BG Teubner,
  1872.

\bibitem{Corrochano2005uncertainty}
E.~B. Corrochano and W.~F{\"o}rstner, ``Uncertainty and projective geometry,''
  \emph{Handbook of Geometric Computing: Applications in Pattern Recognition,
  Computer Vision, Neuralcomputing, and Robotics}, pp. 493--534, 2005.

\bibitem{forstner2016photogrammetric}
W.~F{\"o}rstner and B.~P. Wrobel, \emph{Photogrammetric computer vision}.\hskip
  1em plus 0.5em minus 0.4em\relax Springer, 2016, vol.~6.

\bibitem{forstner2010minimal}
W.~F{\"o}rstner, ``Minimal representations for uncertainty and estimation in
  projective spaces,'' in \emph{Asian Conference on Computer Vision}.\hskip 1em
  plus 0.5em minus 0.4em\relax Springer, 2010, pp. 619--632.

\bibitem{Ramalingam2006generic}
S.~Ramalingam, S.~K. Lodha, and P.~Sturm, ``A generic structure-from-motion
  framework,'' \emph{Computer Vision and Image Understanding}, vol. 103, no.~3,
  pp. 218--228, 2006.

\bibitem{Hartley1997triangulation}
R.~I. Hartley and P.~Sturm, ``Triangulation,'' \emph{Computer vision and image
  understanding}, vol.~68, no.~2, pp. 146--157, 1997.

\bibitem{yang2019iteratively}
K.~Yang, W.~Fang, Y.~Zhao, and N.~Deng, ``Iteratively reweighted midpoint
  method for fast multiple view triangulation,'' \emph{IEEE Robotics and
  Automation Letters}, vol.~4, no.~2, pp. 708--715, 2019.

\bibitem{Thomas2017ETH3D}
T.~Schöps, J.~L. Schönberger, S.~Galliani, T.~Sattler, K.~Schindler,
  M.~Pollefeys, and A.~Geiger, ``A multi-view stereo benchmark with
  high-resolution images and multi-camera videos,'' in \emph{IEEE Conference on
  Computer Vision and Pattern Recognition}, 2017, pp. 2538--2547.

\bibitem{Geiger2012KITTI}
A.~Geiger, P.~Lenz, and R.~Urtasun, ``Are we ready for autonomous driving? the
  kitti vision benchmark suite,'' in \emph{IEEE Conference on Computer Vision
  and Pattern Recognition}, 2012.

\bibitem{Zeisl2009estimation}
B.~Zeisl, P.~F. Georgel, F.~Schweiger, E.~G. Steinbach, N.~Navab, and
  G.~Munich, ``Estimation of location uncertainty for scale invariant features
  points.'' in \emph{British Machine Vision Conference}, 2009, pp. 1--12.

\bibitem{Muhle2023learning}
D.~Muhle, L.~Koestler, K.~M. Jatavallabhula, and D.~Cremers, ``Learning
  correspondence uncertainty via differentiable nonlinear least squares,'' in
  \emph{IEEE Conference on Computer Vision and Pattern Recognition}, 2023, pp.
  13\,102--13\,112.

\bibitem{li2008linear}
H.~Li, R.~Hartley, and J.-h. Kim, ``A linear approach to motion estimation
  using generalized camera models,'' in \emph{2008 IEEE Conference on Computer
  Vision and Pattern Recognition}.\hskip 1em plus 0.5em minus 0.4em\relax IEEE,
  2008, pp. 1--8.

\bibitem{kneip2014opengv}
L.~Kneip and P.~Furgale, ``Opengv: A unified and generalized approach to
  real-time calibrated geometric vision,'' in \emph{IEEE international
  conference on robotics and automation}.\hskip 1em plus 0.5em minus
  0.4em\relax IEEE, 2014, pp. 1--8.

\end{thebibliography}
}
\end{document}